%% file: journal-root.tex
\documentclass[sn-mathphys-num,iicol]{sn-jnl}%

\usepackage{times}
\usepackage{soul}
\usepackage{url}
\usepackage[utf8]{inputenc}
\usepackage{subcaption}
\usepackage{booktabs}

\usepackage{graphics} %
\usepackage{epsfig} %
\usepackage{mathptmx} %
\usepackage{amsmath} %
\usepackage{amssymb}  %
\usepackage{dutchcal}

\usepackage[export]{adjustbox}

\usepackage{siunitx}
\usepackage{multirow}
\usepackage{graphicx}
\usepackage{tabularx}
\usepackage{xcolor}
\usepackage{colortbl}
\usepackage{algorithm}
\usepackage[noend]{algorithmic}
\usepackage[switch]{lineno}
\usepackage{float}
\usepackage{array}

\newcommand{\tablefontsize}{\scriptsize}
\newcolumntype{C}{>{\centering\arraybackslash}p{0.4\textwidth}}

\newcommand\eq[1]{Eq.~\eqref{#1}}
\newcommand\fig[1]{Fig.~\ref{fig:#1}}
\newcommand\sect[1]{Section~\ref{sec:#1}}
\newcommand\tab[1]{Table~\ref{tab:#1}}
\newcommand\alg[1]{Algorithm~\ref{#1}}

\newcommand\comment[1]{\textcolor{black}{\,\,\,\,$\rhd$ #1}}
\DeclareMathOperator*{\argmax}{arg\,max}
\DeclareMathOperator*{\argmin}{arg\,min}
\DeclareMathOperator{\trace}{\mathbf{tr}}
\long\def\invis#1{}

\newcommand{\best}{\bf \color[HTML]{50ac61}}
\newcommand{\bestsecond}{\color[HTML]{ee834a}}
\newcommand{\inputtraj}{\color{orange}}
\newcommand{\outputtraj}{\color{blue}}

\newcommand\revisedtext[1]{\textcolor{black}{#1}}

\makeatletter
\let\old@float@makebox\float@makebox
\renewcommand{\float@makebox}[1]{%
  \color@vbox\normalcolor
    \old@float@makebox{#1}%
  \color@endbox}
\makeatother

\def\HiLi{\leavevmode\rlap{\hbox to \linewidth{\color{yellow!50}\leaders\hrule height .8\baselineskip depth .5ex\hfill}}}

\def\HiLiS{\leavevmode\rlap{\hbox to 0.96\linewidth{\color{yellow!50}\leaders\hrule height .8\baselineskip depth .5ex\hfill}}} %

\newcommand{\detected}{\mathcal{d}} 
\newcommand{\localized}{\mathcal{l}} 
\newcommand{\search}{\mathcal{s}}
\newcommand{\track}{\mathcal{t}}

\newcommand{\A}{A} %
\newcommand{\K}{O} %
\newcommand{\Kdetect}{\revisedtext{\K_\detected}} %
\newcommand{\M}{\mathbf{M}} %

\newcommand{\mi}[1]{\ensuremath{\mathit{#1}}}
\newcommand{\algorithmicbreak}{\textbf{break}}
\newcommand{\BREAK}{\STATE \algorithmicbreak}

\mathchardef\mhyphen="2D 
\newcommand{\ag}{a} %
\newcommand{\ob}{o} %
\newcommand{\ce}{c} %
\newcommand{\z}{z} %
\newcommand{\x}{x} %
\newcommand{\zdet}{\revisedtext{\z_{\ob\mhyphen\detected}}}%
\newcommand{\zlocag}{\revisedtext{\z_{\ag,\ob\mhyphen\localized}}}%
\newcommand{\zlocagt}{\revisedtext{\z^t_{\ag,\ob\mhyphen\localized}}}%
\newcommand{\zlocagzt}{\revisedtext{\z^{0:t}_{\ag,\ob\mhyphen\localized}}}%
\newcommand{\xlocagt}{\revisedtext{\x_{\ob\mhyphen\localized}}}%
\newcommand{\zloce}{\revisedtext{\z_{e,\ob\mhyphen\localized}}}%
\newcommand{\belsearch}{\revisedtext{\mi{bel}_\search}}%
\newcommand{\beltrack}{\revisedtext{\mi{bel}_\track}}%
\newcommand{\beliefsearch}[3]{\belsearch(#1,#2,#3)} %
\newcommand{\obeltrack}{\revisedtext{\overline{\mi{bel}}_\track}}%
\newcommand{\belieftrack}[3]{\beltrack(#1,#2,#3)} %
\newcommand{\obelieftrack}[3]{\obeltrack(#1,#2,#3)} %
\newcommand{\belsharedsearch}[2]{\belsearch(#1,#2)} %
\newcommand{\belsharedtrack}[1]{\beltrack(#1)} %
\newcommand{\ex}[3]{\Theta(#1,#2,#3)} %
\newcommand{\jexplore}{J_\mi{explore}} %
\newcommand{\jexploit}{J_\mi{exploit}} %

\title{Multi-Object Active Search and Tracking by Multiple Agents in Untrusted, Dynamically Changing Environments}

\author*[1]{\fnm{Mingi} \sur{Jeong}}\email{mingi.jeong.gr@dartmouth.edu}

\author[2]{\fnm{Cristian} \sur{Molinaro}}\email{cmolinaro@dimes.unical.it}

\author[3]{\fnm{Tonmoay} \sur{Deb}}\email{tonmoay.deb@northwestern.edu}

\author[4]{\fnm{Youzhi} \sur{Zhang}}\email{youzhi.zhang@cair-cas.org.hk}

\author[2]{\fnm{Andrea} \sur{Pugliese}}\email{andrea.pugliese@unical.it}

\author[5]{\fnm{Eugene} \sur{Santos Jr.}}\email{eugene.santos.jr@dartmouth.edu}

\author[3]{\fnm{VS} \sur{Subrahmanian}}\email{vss@northwestern.edu}

\author*[1]{\fnm{Alberto} \sur{Quattrini Li}}\email{alberto.quattrini.li@dartmouth.edu}

\affil*[1]{\orgdiv{Department of Computer Science}, \orgname{Dartmouth College}, \orgaddress{%
\city{Hanover}, \postcode{03755}, \state{NH}, \country{USA}}}

\affil[2]{\orgdiv{DIMES}, \orgname{University of
Calabria}, \orgaddress{%
\city{Arcavata}, \postcode{87036}, %
\country{Italy}}}

\affil[3]{\orgdiv{Department of Computer Science}, \orgname{Northwestern  University}, \orgaddress{%
\city{Evanston}, \postcode{60208}, \state{IL}, \country{USA}}}

\affil[4]{\orgdiv{Centre for Artificial Intelligence and Robotics
and Hong Kong Institute of Science and Innovation}, \orgname{Chinese Academy of
Science}, \orgaddress{\country{Hong Kong}}}

\affil[5]{\orgdiv{Thayer School of Engineering}, \orgname{Dartmouth College}, \orgaddress{%
\city{Hanover}, \postcode{03755}, \state{NH}, \country{USA}}}

\begin{document}

\abstract{
This paper addresses the problem of both actively \emph{searching} and \emph{tracking} multiple unknown dynamic objects in a known environment with multiple cooperative autonomous agents with partial observability. \revisedtext{The tracking of a target ends when the uncertainty is below a specified threshold}.  %
\revisedtext{Current methods typically assume homogeneous agents without access to external information and utilize short-horizon target predictive models. Such assumptions limit real-world applications. We propose a fully integrated pipeline where the main novel contributions are: (1) a time-varying weighted belief representation capable of handling knowledge that changes over time, which includes external reports of varying levels of trustworthiness in addition to the agents involved; (2) the integration of a Long Short Term Memory-based trajectory prediction within the optimization framework for long-horizon decision-making, which accounts for trajectory prediction in time-configuration space, thus increasing responsiveness; and (3) a comprehensive system that accounts for multiple agents and enables information-driven optimization during both the search and track tasks.} %
When communication is available, our proposed strategy consolidates exploration results collected asynchronously by agents and external sources into a headquarters, who can allocate each agent to maximize the overall team's utility, effectively using all available information.  We tested our approach
extensively in Monte Carlo simulations against baselines, \revisedtext{representative of classes of approaches from the literature,} and in robustness and ablation studies. \revisedtext{In addition, we performed experiments in a 3D physics based engine robot simulator to test the applicability in the real world,} as well as with real-world trajectories obtained from an oceanography computational fluid dynamics simulator. \revisedtext{Results show the effectiveness of our proposed method, which achieves mission completion times that are $1.3$ to $3.2$ times faster in finding all targets, even under the most challenging scenarios where the number of targets is \num{5} times greater than that of the agents.
}
}

\keywords{Multi-robot systems, Search and tracking}

\maketitle

\section{Introduction}
\begin{figure}[t!]
  \centering
  \includegraphics[width=0.7\columnwidth]{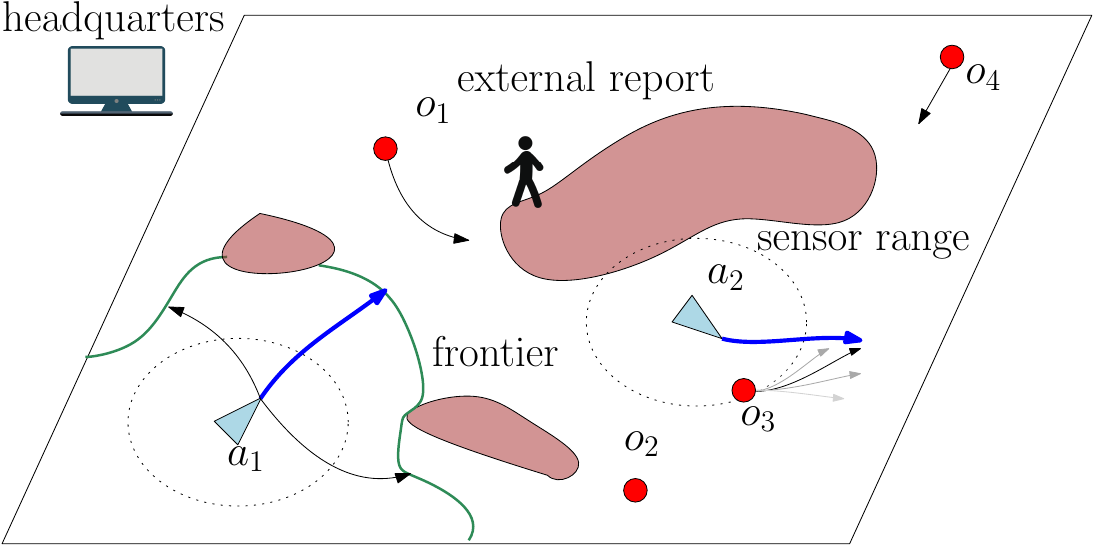}
  \caption{Example scenario: cooperative agents $\ag_1, \ag_2$ that search and track unknown target objects $\ob_1, \ob_2, \ob_3, \ob_4$ in a known environment. The headquarters collects information sources, including agents and independent external report information sources, and coordinates tasks to maximize information gain. $\ag_1$ is searching the environment following a trajectory with the best utility (blue) on the frontier, integrating the external report. $\ag_2$ detected $\ob_3$ and is tracking to reduce $\ob_3$ location's uncertainty, exploiting a long-term trajectory prediction. \revisedtext{Once the uncertainty is below a specified threshold, the target is cleared and does not need to be tracked anymore}.} 
  \label{fig:beautyfig}
  \vspace{-1.5em}
\end{figure}

This paper tackles the problem of both actively \textbf{searching} and \textbf{tracking} unknown dynamic objects in an environment with multiple cooperative autonomous agents with partial observability---see \fig{beautyfig}. Given a prior map of the environment (e.g., disaster zone), autonomous agents (e.g., robots) with a noisy sensor (e.g., camera), and external reports from other sources (e.g., other humans), the goal is to take actions to search the environment, detect unknown moving targets (e.g., stranded people), and track the detected targets to infer their location as quickly as possible, \revisedtext{reducing the uncertainty below a specified threshold}. 
The proposed method applies to a broad range of real-world application domains, such as animal tracking and search and rescue.

Despite several important studies on variants of search (sometimes called \emph{exploration}) and tracking (also called \emph{exploitation}), there are several open problems \cite{review-csat2018,taxonomy-review-2016}. 
Many studies analyzed the benefits of multi-robot coordination and the trade-off between exploration and exploitation, some looking at a centralized system \cite{upenn-balance-2015,target-assign-csat-2008,vijay-sat-2017}, others at a decentralized one \cite{increasing-autonomy-csat-2009,recursive-bayesian-csat-2006,optimal-swarm-2020}. Such approaches typically use a \emph{homogeneous source} of information from \emph{homogeneous agents}. This assumption limits  real-world applications of such systems, where external information sources, such as human reporting, are available, helping achieve the task more efficiently. In fact, heterogeneous teams have been shown to perform well in cooperative tasks~\cite{ov2020impact}. Homogeneous agents can also have different levels of uncertainty about a specific target's location. Thus, there is a need to fuse heterogeneous information. 
In addition, the lookahead for these approaches is typically single-step: %
the next-best action is based on a short-term view of where the detected target will go, using filtering methods \cite{increasing-autonomy-csat-2009,recursive-bayesian-csat-2006,target-assign-csat-2008,upenn-balance-2015,vijay-sat-2017,search-track-cyprus-2021-journal} or direct guidance to the target using variants of artificial vector fields or forces \cite{optimal-swarm-2020,harvard-source-seek-2021}. Instead, considering long-term intent could more effectively reduce uncertainty and increase the likelihood of detecting and tracking the target.

\begin{table*}[!t]
\tablefontsize
\centering
 \resizebox{\textwidth}{!}{%
\begin{tabular}
{c|c|ccc|c|c|c|ccc|}
\cmidrule{2-11}
& \multicolumn{6}{C|}{Scenario} & \multicolumn{4}{C|}{Approach} \\
\cmidrule{2-11}
\multirow{2}{*}{} &
  \multirow{2}{*}{\begin{tabular}[c]{@{}c@{}}Tested\\ environments\end{tabular}} &
  \multicolumn{3}{c|}{Target} &
  \multirow{2}{*}{\begin{tabular}[c]{@{}c@{}}Agent team \\ type\end{tabular}} &
  \multirow{2}{*}{\begin{tabular}[c]{@{}c@{}}External \\ info.\end{tabular}} &
  \multirow{2}{*}{\begin{tabular}[c]{@{}c@{}}Coordination\\ method\end{tabular}} &
  \multicolumn{3}{c|}{Planning} \\ \cmidrule{3-5} \cmidrule{9-11} 
 &
   &
  \multicolumn{1}{c|}{Motion} &
  Number  & \multicolumn{1}{|c|}{\begin{tabular}[c]{@{}c@{}}Known \\ \# objects\end{tabular}} &
   &
   &
   &
  \multicolumn{1}{c|}{Type} &
  \multicolumn{1}{c|}{\begin{tabular}[c]{@{}c@{}}Time-varying\\ for exploration\end{tabular}} &
  \begin{tabular}[c]{@{}c@{}}Target trajectory\\ estimation\end{tabular} \\ \hline \hline
\multicolumn{1}{|c|}{%
\cite{malika-search-2016}} &
  open &
  \multicolumn{1}{c|}{dynamic} &
  multi &
  \multicolumn{1}{|c|}{N} &
  \multicolumn{1}{c|}{single} &
  N &
  n/a &
  \multicolumn{1}{c|}{search only} &
  \multicolumn{1}{c|}{N} &
  n/a \\ \hline
\multicolumn{1}{|c|}{\cite{harvard-source-seek-2021}} &
  open &
  \multicolumn{1}{c|}{dynamic} &
  single &
  \multicolumn{1}{|c|}{n/a} &
  \multicolumn{1}{c|}{homogeneous} &
  N &
  centralized &
  \multicolumn{1}{c|}{track only} &
  \multicolumn{1}{c|}{N} &
  KF-based, $t+1$ \\ \hline
\multicolumn{1}{|c|}{\cite{optimal-swarm-2020}} &
  open &
  \multicolumn{1}{c|}{dynamic} &
  single &
  \multicolumn{1}{|c|}{n/a} &
  \multicolumn{1}{c|}{homogeneous} &
  N &
  decentralized &
  \multicolumn{1}{c|}{balanced} &
  \multicolumn{1}{c|}{N} &
  N \\ \hline
\multicolumn{1}{|c|}{%
\cite{increasing-autonomy-csat-2009}} &
  with obstacles &
  \multicolumn{1}{c|}{dynamic} &
  multi &
  \multicolumn{1}{|c|}{Y} &
  \multicolumn{1}{c|}{homogeneous} &
  N &
  decentralized &
  \multicolumn{1}{c|}{mode switch} &
  \multicolumn{1}{c|}{Y} &
  KF-based, $t+1$ \\ \hline
\multicolumn{1}{|c|}{%
\cite{recursive-bayesian-csat-2006}} &
  open &
  \multicolumn{1}{c|}{dynamic} &
  multi &
  \multicolumn{1}{|c|}{Y} &
  \multicolumn{1}{c|}{homogeneous} &
  N &
  decentralized &
  \multicolumn{1}{c|}{mode switch} &
  \multicolumn{1}{c|}{N} &
  Bayesian, $t+\kappa$ \\ \hline
\multicolumn{1}{|c|}{\revisedtext{\cite{gas-mapping-2019}}} &
  \revisedtext{open} &
  \multicolumn{1}{c|}{\revisedtext{static}} &
  \revisedtext{multi} &
  \multicolumn{1}{|c|}{\revisedtext{N}} &
  \multicolumn{1}{c|}{\revisedtext{homogeneous}} &
  \revisedtext{N} &
  \revisedtext{decentralized} &
  \multicolumn{1}{c|}{\revisedtext{balanced}} &
  \multicolumn{1}{c|}{\revisedtext{N}} &
  \revisedtext{n/a}  \\ \hline
\multicolumn{1}{|c|}{%
\cite{upenn-balance-2015}} &
  with obstacles &
  \multicolumn{1}{c|}{static} &
  multi &
  \multicolumn{1}{|c|}{N} &
  \multicolumn{1}{c|}{homogeneous} &
  N &
  centralized &
  \multicolumn{1}{c|}{balanced} &
  \multicolumn{1}{c|}{N} &
  n/a \\ \hline
\multicolumn{1}{|c|}{\revisedtext{\cite{cell-mb-2023}}} &
  \revisedtext{open} &
  \multicolumn{1}{c|}{\revisedtext{dynamic}} &
  \revisedtext{multi} &
  \multicolumn{1}{|c|}{\revisedtext{N}} &
  \multicolumn{1}{c|}{\revisedtext{single}} &
  \revisedtext{N} &
  \revisedtext{n/a} &
  \multicolumn{1}{c|}{\revisedtext{balanced}} &
  \multicolumn{1}{c|}{\revisedtext{N}} &
  \revisedtext{PHD-based, $t+1$}  \\ \hline
\multicolumn{1}{|c|}{\revisedtext{\cite{pan-tilt-camera-2019}}} &
\revisedtext{
  with obstacles} &
  \multicolumn{1}{c|}{\revisedtext{dynamic}} &
  \revisedtext{multi} &
  \multicolumn{1}{|c|}{\revisedtext{N}} &
  \multicolumn{1}{c|}{\revisedtext{single}} &
  \revisedtext{N} &
  \revisedtext{n/a} &
  \multicolumn{1}{c|}{\revisedtext{balanced}} &
  \multicolumn{1}{c|}{\revisedtext{N}} &
  \revisedtext{Velocity field} $t+1$  \\ \hline
\multicolumn{1}{|c|}{\revisedtext{\cite{distributed-control-silvia-2018}}} &
  \revisedtext{with obstacles} &
  \multicolumn{1}{c|}{\revisedtext{dynamic}} &
  \revisedtext{multi} &
  \multicolumn{1}{|c|}{\revisedtext{N}} &
  \multicolumn{1}{c|}{\revisedtext{homogeneous}} &
  \revisedtext{N} &
  \revisedtext{decentralized} &
  \multicolumn{1}{c|}{\revisedtext{balanced}} &
  \multicolumn{1}{c|}{\revisedtext{N}} &
  \revisedtext{Physics-based $t+1$}  \\ \hline
\multicolumn{1}{|c|}{%
\revisedtext{\cite{usv-pursuit-evasion-2023}}} &
  \revisedtext{with obstacles} &
  \multicolumn{1}{c|}{\revisedtext{dynamic}} &
  \revisedtext{multi} &
  \multicolumn{1}{|c|}{\revisedtext{N}} &
  \multicolumn{1}{c|}{\revisedtext{homogeneous}} &
  \revisedtext{N} &
  \revisedtext{decentralized} &
  \multicolumn{1}{c|}{\revisedtext{balanced}} &
  \multicolumn{1}{c|}{\revisedtext{N}} &
  \revisedtext{Physics-based $t+1$}  \\ \hline
\multicolumn{1}{|c|}{\cite{target-assign-csat-2008}} &
  open &
  \multicolumn{1}{c|}{dynamic} &
  multi &
  \multicolumn{1}{|c|}{N} &
  \multicolumn{1}{c|}{homogeneous} &
  N &
  centralized &
  \multicolumn{1}{c|}{mode switch} &
  \multicolumn{1}{c|}{N} &
  KF-based, $t+1$ \\ \hline
\multicolumn{1}{|c|}{\cite{vijay-sat-2017}} &
  open &
  \multicolumn{1}{c|}{dynamic} &
  multi &
  \multicolumn{1}{|c|}{N} &
  \multicolumn{1}{c|}{homogeneous} &
  N &
  centralized &
  \multicolumn{1}{c|}{balanced} &
  \multicolumn{1}{c|}{N} &
  PHD-based, $t+\kappa$ \\ \hline
\multicolumn{1}{|c|}{\cite{search-track-cyprus-2021-journal}} &
  open &
  \multicolumn{1}{c|}{dynamic} &
  multi &
  \multicolumn{1}{|c|}{N} &
  \multicolumn{1}{c|}{homogeneous} &
  N &
  decentralized &
  \multicolumn{1}{c|}{mode switch} &
  \multicolumn{1}{c|}{N} &
  PHD-based, $t+\kappa$ \\ \hline\hline
\multicolumn{1}{|c|}{\textbf{ours}} &
  with obstacles &
  \multicolumn{1}{c|}{dynamic} &
  multi &
  \multicolumn{1}{|c|}{N*} &
  \multicolumn{1}{c|}{\textbf{heterogeneous}} &
  \textbf{Y} & 
  \textbf{hybrid} &
  \multicolumn{1}{c|}{\begin{tabular}[c]{@{}c@{}}\textbf{mode swith}\\ \textbf{with balanced}\end{tabular}} &
  \multicolumn{1}{c|}{\textbf{Y}} &
  \textbf{LSTM-based}, $t+\kappa$ \\ \hline
\end{tabular}%
}
\caption{Comparison with some representative literature in search and tracking of dynamic targets. Our paper handles environments with obstacles where multiple targets  follow potentially non-linear trajectories. (*Note that while our approach can work without knowing the number of targets, in the experiments we assume we know it for identifying when to terminate them.) The agent team type, differently from literature, is heterogeneous and external information is available. Agents can coordinate through HQ but each agent makes its own decisions, thus the coordination method is hybrid. Planning follows mode switch but keeps multi-criteria optimization with different priorities in the two modes; it considers time-varying uncertainty; and it exploits target trajectory estimation based on LSTM. \revisedtext{We selected the baselines based on current related work and covering different classes of approaches, which are implemented for experiments and discussed in \sect{experiments}. %
}
}
\label{tab:literature-review}
\end{table*}

We address these challenges via a novel approach that introduces \textbf{active search and tracking of multiple dynamic targets} augmented by \textbf{learning-based intent-awareness and external information sources}. %
\revisedtext{While each agent can operate independently}, our approach consolidates search and track results collected asynchronously by agents and external sources into a central \revisedtext{headquarters (HQ)}, allowing central computation and decision-making to allocate each agent to maximize the team's overall utility consisting of exploration and exploitation. 
Specifically, we make the following main \revisedtext{technical contributions and advancements}:
\begin{itemize}
    \item \revisedtext{Time-varying weighted belief representation based on uncertainty level/trustworthiness, resulting in the ability to work with heterogeneous agents, independent third-party reporting, and target's dynamic nature;}
    \item \revisedtext{Integration of Long Short Term Memory (LSTM) based trajectory prediction that allows maximum probability of detection in time-configuration space and corresponding long-horizon decision-making to track a target;}
    \item \revisedtext{Full integrated pipeline for multiple agents, with information-theoretic optimization, mode switch decisions, and a hybrid coordination method, leading to a complete system for both search and track;}

    \item \revisedtext{Discussion with corresponding insights on effective search and tracking, from our implementation in the Robot Operating System (ROS) and extensive Monte Carlo simulations, compared to baselines covering classes of approaches in the literature, ablation studies, and real-world application with targets trajectories from realistic Computational Fluid Dynamic datasets and in 3D physic-based engine robotic simulator.}
\end{itemize}
\revisedtext{
Experiments showed the effectiveness of our proposed approach, demonstrating that it is $1.3$ to $3.2$ times faster at finding all targets, even under the most challenging scenarios where the ratio of agents to targets is $1:5$. Overall, the proposed novel contributions and complete pipeline integration will help autonomous agents to take effective decisions for important tasks, such as search and rescue. 
}

\section{Related Work}

\revisedtext{
The problem of finding target objects has been addressed in the literature with a variety of  assumptions, falling in three categories, \emph{static targets}, \emph{pursuit evasion}, and \emph{search, acquisition, and tracking}. Here, we highlight how each category relate or differ with the problem and the approach we propose.}

\begin{figure*}[t!]
    \centering
    \includegraphics[width=\textwidth]{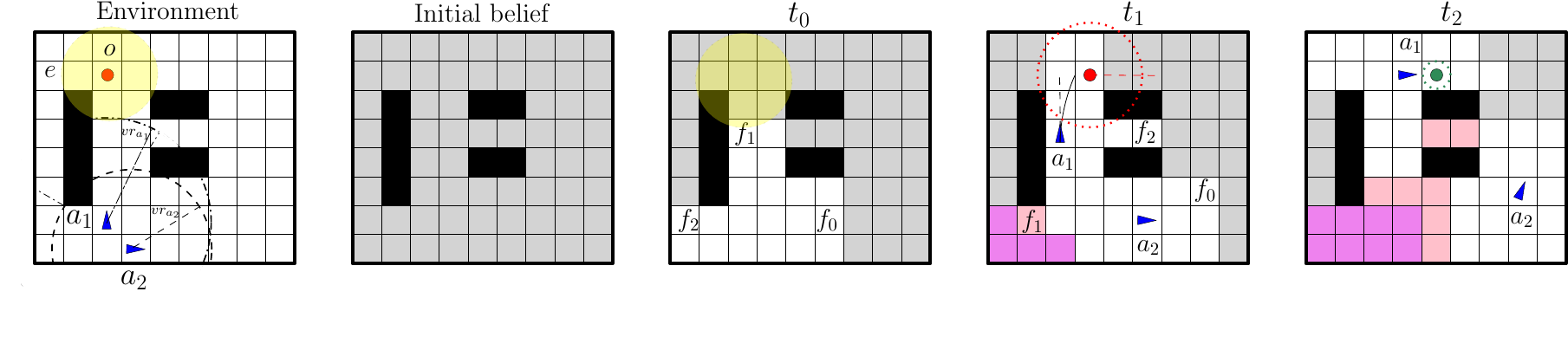}
    \vspace{-3em}
    \caption{Simplified example over a small grid that shows: (1) \emph{Environment}, with two agents \revisedtext{($\ag_1$, $\ag_2$ in blue)} with their sensor ranges \revisedtext{(dashed circles)}, one object \revisedtext{($\ob$ in red)}, and one event of third-party reporting \revisedtext{($e$ in yellow circle)}; (2) \emph{Initial belief} without any prior, thus all cells unknown; (3) at time $t_0$ the shared belief after the initial sensor measurements by both agents with the third-party reporting and the frontiers \revisedtext{($f_0, f_1, f_2)$}; (4) at $t_1$ the agents move to the optimal frontier locations and get corresponding sensor measurements, where $\ag_1$ also detects an object $o$ with uncertainty \revisedtext{(red dotted circle)} and estimates its trajectory \revisedtext{(red dashed line)}; and (5) at $t_2$, the agents select their next action, resulting in $\ag_1$ achieving the tracking of the object\revisedtext{, i.e., reducing the uncertainty (green dotted circle) and clearing the target (green)}, and $\ag_2$ continuing independently on the search. \revisedtext{Note that the cells (in pink) represent the time-varying object occupancy proposed in this study where the lighter color, the more recently explored.}}
    \label{fig:example}
\end{figure*}

\textbf{Static targets.} Some work in the space of coverage~\cite{galceran2013survey}, exploration~\cite{quattrini2020exploration}, search and rescue~\cite{drew2021multi}, information gathering~\cite{bai2021information}, and source seeking~\cite{hajieghrary2017information,marjovi2010multi} seeks to discover information about features in the environment such as the location of or targets~\cite{rouvcek2020darpa}. In classical frontier-based exploration~\cite{yamauchi1998frontier}, the robot selects the closest location at the boundary between known and unknown parts of the environment---the \emph{frontier}---to proceed with exploration. Information-theoretic approaches include information gain together with the cost of visiting locations~\cite{burgard2005coordinated,basilico2011exploration}. A variation of the problem includes having competitive teams searching for a target~\cite{otte2018competitive}. Unlike our paper, these methods do not consider moving targets --- once an area is marked as explored, it is not re-visited. \revisedtext{We get inspiration from frontier-based exploration to identify areas to explore, including a time-varying component, which is not present in its vanilla version.}

\textbf{Pursuit/evasion.} Research that typically considers moving targets addresses the pursuit-evasion problem~\cite{chung2011search}.
The first body of work in pursuit-evasion studies the problem from a theoretical perspective, modeling the environment either as a graph~\cite{isler2006randomized,borie2011algorithms,kehagias2009graph} or as polygons~\cite{isler2005randomized,guibas1999visibility,quattrini2018search,stiffler2017complete}. These approaches try to find methods with worst-case theoretical guarantees  such as the number of pursuers needed to capture an evader, or the traveled distance before the evader is captured. Instead of studying the worst case, another body of work solves the problem of localizing and tracking other agents within a probabilistic optimization framework~\cite{li2021bayesian,chung2011analysis,lau2006probabilistic,matzliach2020cooperative}. Past work typically assumes that the searcher agent can rely on its own sensors, there is  a priori knowledge of where the targets might be, or there is a trail left behind, in a source-seeking problem formulation. Patrolling can be considered a dual problem to pursuit-evasion, in that an area is initially considered secure and should be protected by attacks~\cite{basilico2022recent}, where an adversarial model is assumed to be known. \revisedtext{Generally, methods from the pursuit/evasion literature might take longer trajectories to account for the adversarial behaviors. Instead,} in our paper, we assume that targets are not adversarial and that no prior knowledge is available. 

\textbf{Search, Acquisition, and Tracking.} Unlike  pursuit-evasion, other work studied non-adversarial but non-cooperative scenario---i.e., objects are not trying to actively evade, but are also not helping others to locate themselves. This problem is also called the  \emph{cooperative search, acquisition, and tracking problem} \cite{review-csat2018,taxonomy-review-2016,sat-swarm-review-2016} and is the focus of this paper.  
Some past efforts focus on a single task, e.g. search \cite{malika-search-2016} or tracking \cite{harvard-source-seek-2021}. 
Other efforts include both. 
Some resolve conflicting behaviors between search and tracking with \textit{mode switch} \cite{target-assign-csat-2008,increasing-autonomy-csat-2009,recursive-bayesian-csat-2006,search-track-cyprus-2021-journal,goldhoorn2018searching}. If agents detect a target, they switch to tracking mode to ensure the detected target's location is within an error bound. Others propose \textit{balanced} approaches, with a single objective function such as the \revisedtext{expected number of detections  \cite{vijay-sat-2017, gas-mapping-2019}}, \revisedtext{a multi-criteria optimization \cite{uav-route-csat-balance-2012,upenn-balance-2015, cell-mb-2023}}, \revisedtext{a multi-objective optimization \cite{pan-tilt-camera-2019, distributed-control-silvia-2018},} or swarm intelligence \revisedtext{\cite{blum2015swarm, optimal-swarm-2020, usv-pursuit-evasion-2023}}.  
These approaches consider \emph{homogeneous agents}, no available external information, and no time-varying uncertainty. Also, many approaches have been only tested in open-space environments, thereby overlooking realistic cluttered environments\revisedtext{, or considering cluttered environments but with a single agent (sensor platform) \cite{pan-tilt-camera-2019}}. \revisedtext{A time-varying 3-D coverage cone was introduced to consider a spatio-temporal space during tracking dynamic targets after detection. For the search phase, a} diffusion model was introduced to account for dynamic targets \cite{increasing-autonomy-csat-2009}. This model favors revisiting previously explored areas and maintains a map for each obstacle and updates the map using phantom obstacles. \revisedtext{To improve scalability with respect to the number of targets, we store and update only one time-varying map for exploration, without losing any information.} %
The trajectory estimation is based primarily on Kalman filter or iterative filtering approaches, which generally might fail in long-term predictions \revisedtext{\cite{intention-vehicle-prediction-2022}}, or PHD filter which might not capture complex trajectories\revisedtext{, due to its limit in distinguishing between objects \cite{phd-filter-limit-2022,vijay-sat-2017}}.

\revisedtext{
Real-world scenarios for search and tracking include heterogeneous agents and third-party external information. Therefore, this paper investigates how to model and use effectively such information for improved search and tracking task (see \tab{literature-review} for an overview on the difference between our work and the literature).
}

\section{Problem Formulation} \label{sec:prob-state}
We formulate a problem called \emph{active target state estimation}. In a known environment, autonomous agents search for unknown targets (\textit{search}) and, when detected, exploit the  target location for improving the localization accuracy of the target until a certain threshold (\textit{track}), \revisedtext{resulting in the clearing of the target}. A simplified example that represents the problem described in this section is shown in \fig{example}-Environment.

\textbf{Environment.} We consider a known 2D bounded environment $\M \subset \mathbb{R}^2$ with static obstacles: points can be either freespace $\M_\mi{free}$ or obstacle $\M_\mi{obs}$.

\textbf{Target objects.} $\M_\mi{free}$ contains non-adversarial non-cooperative moving target objects $\K$ whose locations are not initially known. 
Target objects are considered to be points, without loss of generality, by growing the obstacles to account for their real size, as done commonly in path planning. Note that no location can be occupied by more than one object or agent. 
\revisedtext{The number of target objects is not necessarily known---in our experiments we use that number just as a termination condition for the experiments.}
Each target object $\ob$ can move with a maximum velocity $s_\ob$, following its own motion model, which is not known to the agents.

\textbf{Agents.} A team of heterogeneous autonomous agents $\A = \{ \ag_1, \ldots, \ag_n \}$ are deployed in the freespace $\M_\mi{free}$. Their initial pose is   known. As in the case of targets, agents are also considered to be points, without loss of generality, and no location can be occupied by more than one agent or object. \revisedtext{There is no constraint on the number of agents relative to the number of targets, ensuring that the overall search and track task remains generalizable, even when there are fewer agents than targets.}

\textit{Motion model.} Each agent $\ag \in \A$ can move at a maximum speed $s_\ag$. We assume that $s_\ob < s_\ag$, given that the scenario is non-adversarial. This assumption ensures that \revisedtext{the problem can be solved with any arbitrary number of agents and targets}: an object can be tracked once it is detected by any $\ag \in \A$. It would be interesting for future work to relax this assumption.
Each agent can be controlled with velocity commands or waypoints in $\M_\mi{free}$, given that the agent has a navigation system, consisting of a global path planner (e.g., A*~\cite{hart1968formal}) and a local controller (e.g., time-elastic band  \cite{rosmann2017integrated}).

\begin{figure*}[t!]
    \centering
    \includegraphics[width=.9\textwidth]{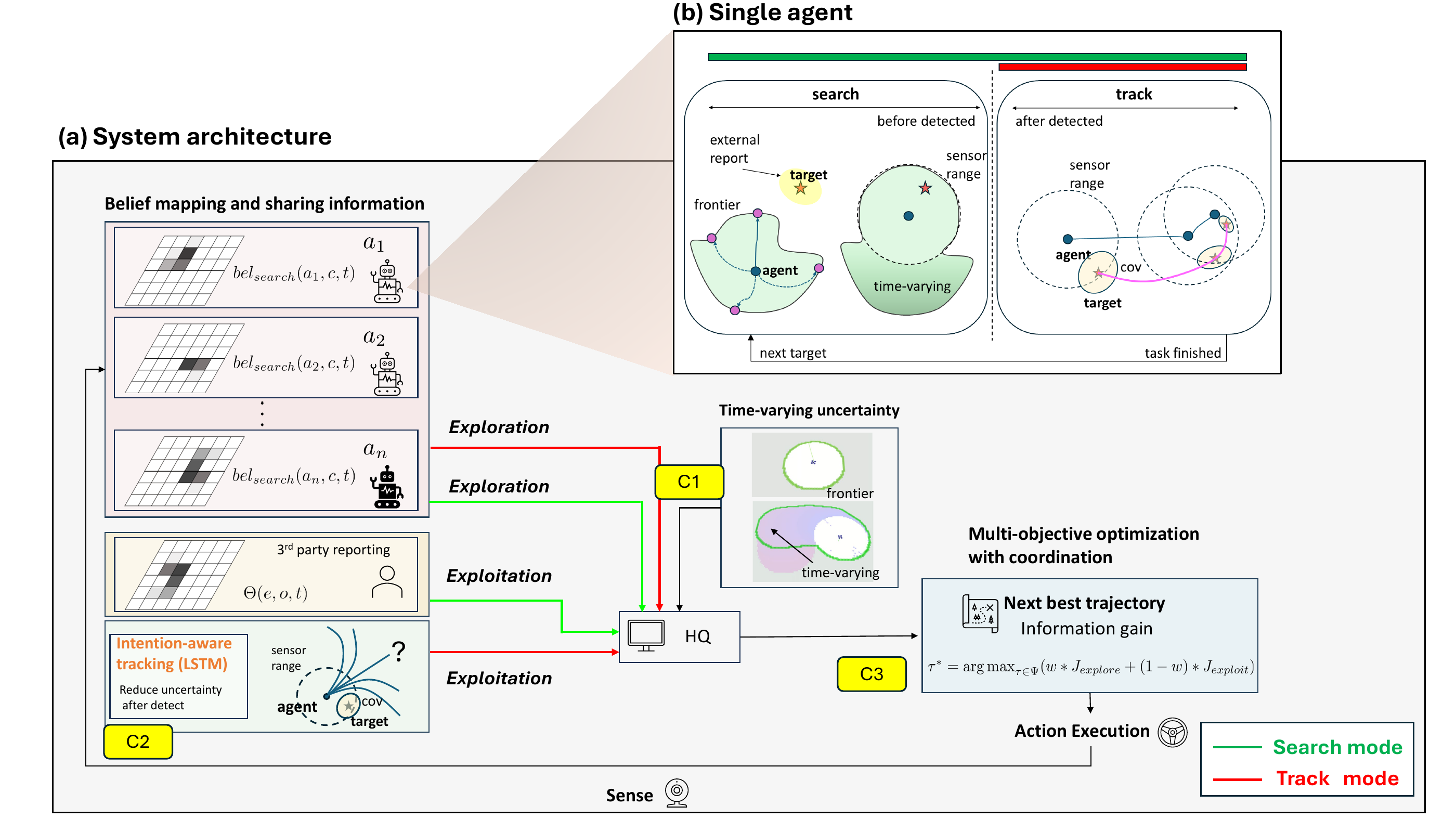}
    \caption{\revisedtext{Our proposed multi-object active search and tracking system. \textbf{(a) Overall system architecture} for search and track process, with multiple agents that can move and sense, forming a belief with time-varying uncertainty and including third-party reporting, as well as intention-aware tracking, to then select the best trajectory based on both exploration and exploitation values during both modes. \textbf{(b) A single agent's perspective}: during the search phase, to quickly detect the target, the agent follows an optimal trajectory (blue line) towards a frontier, guided by the $\jexplore$ and $\jexploit$ utilities that include information from third-party reporting. Once the target is within the agent's visual range, $\mi{vr}_\ag$, the tracking phase begins. To rapidly reduce the uncertainty of the target’s location, the agent follows a long-horizon optimal trajectory (blue line), assisted by $\jexplore$ and $\jexploit$ utilities that include intent information from learning-based trajectory prediction (pink line). After the task on the current target is completed, the agent searches for the next target. Note that the technical contributions in our paper are highlighted in $C1$ to $C3$.}}
    \label{fig:system-architecture}
\end{figure*}

\textit{Sensor model.} Each agent $a$ has a localization system (e.g., adaptive Monte Carlo localizer), allowing it to localize in the environment, potentially with some noise. Agents are equipped with a sensor (e.g., omnidirectional RGB camera, thermal camera, RADAR),   which has a \ang{360} field of view (FoV) and a maximum range $\mi{vr}_\ag$, able to: \revisedtext{(i)} detect and identify target objects $\ob$ and \revisedtext{(ii)} return the locations of the detected targets. 
\revisedtext{We consider a realistic sensor that} follows a noisy observation/measurement model:

\noindent (i) \revisedtext{\emph{detection and identification}}: the actual detection $\zdet$ of the target $\ob$ given the true state $x_\ob$ follows the probability $p(\zdet|x_\ob)$ \revisedtext{ that the sensor returns a measurement $\zdet$ given $x_\ob$, the actual existence of the obstacle in the sensor range}:
\begin{small}
\begin{align*}
p(\zdet=1 ~|~ x_\ob=1) &= 1-\alpha & p(\zdet=0 ~|~ x_\ob=1) &= \alpha \\
p(\zdet=1 ~|~ x_\ob=0) &= \beta & p(\zdet=0 ~|~ x_\ob=0) &= 1- \beta
\end{align*}
\end{small}
\!\!where $\alpha$ and $\beta$ are between $0$ and $1$. %
The first row of the sensor model represents true positive and false negative, whereas the second row represents false positive and true negative, respectively. Note that if $\alpha$ and $\beta$ are equal to $0$ the detection has no noise.

\noindent (ii) \revisedtext{\emph{measurement of the location}: if detected, the returned location $\zlocag$ of a target $\ob$, at its true location $\xlocagt$ (unknown to the agent) is affected by white Gaussian noise $w_\ob \sim \mathcal{N}(0, \Sigma_\ob(d(\ag,\ob)))$ and the measurement function $h(\cdot)$:
\begin{equation} \label{eq:track-loc-measure}
    \zlocag = h(\xlocagt) + w_\ob
\end{equation}}
\noindent %
The covariance $\Sigma_\ob$ is a 2x2 matrix, where the diagonal elements represent noise over the 2D location coordinates and the off-diagonal elements are $0$. Note that the covariance depends on the distance $d$ between the agent and the object: the closer the agent is to the object, the smaller the uncertainty.

\textit{HQ/communication.} We assume the agents in $\A$ can communicate with reliable communication link (e.g., Starlink satellite communication) with each other and a headquarters (HQ), allowing them to share information and potentially coordinate.

\textbf{External report.} There can also be an additional source of information $\zloce$ on the location of a target object $\ob$ at time $t$. 
Such information can be provided to the agents in $\A$ and/or HQ by third-party entities $e \in E$, where $E$ is a set of external agent-independent sources. Similarly to the observation model of the agents in $\A$, that information has varying degrees of uncertainty.

\textbf{Agents' belief.}
For each agent $\ag\in \A$, %
location $\ce\in \M_{\mi{free}}$, and time $t$, each $\ag$ has a belief about $\ce$ being occupied or free at time $t$, denoted  $\beliefsearch{\ag}{\ce}{t}$ for the \textit{search} process. Once $\ag$ detects the target object $\ob$,  $\ag$ has a belief on the target location $\belieftrack{\ag}{\ob}{t}$, with an uncertainty expressed in covariance $\Sigma_{\ob}$ for the \textit{tracking} process.

Because the target object can move, $\beliefsearch{\ag}{\ce}{t}$ can vary as time elapses, even if a cell was observed before, as the target object can reenter a previously explored area.

\textbf{Objective.} Given an environment $\M$, third-party reporters $E$, a team of agents $\A$ starting from an initial position $q_\ag^0$ for each $\ag \in \A$, the \revisedtext{overall} goal is to find a sequence of positions $Q_\ag=\langle q_\ag^0, \ldots, q_\ag^n  \rangle$ for each agent $\ag \in \A$, so that all target objects $\K$  are \emph{searched} and \emph{tracked} by $\A$ in the minimum amount of time. 
More formally, for every object $\ob \in \K$, $\ob$ is (1) \revisedtext{\textit{detected}} by some $\ag \in \A$, i.e., $\ob$ is located within the sensor range of $\ag$, at a distance less than or equal to $\mi{vr}_\ag$; and (2) \revisedtext{\textit{tracked}} by agent $\ag$ such that the uncertainty of the target object $\ob$ location is lower than or equal to a threshold $\theta$. Given the relationship between sensor uncertainty and distance, $\theta$ can also be set to achieve a desired distance to the target. \revisedtext{The task for an individual object is completed when the object is tracked once, thus will be no longer considered, i.e., $\ag$ is certain about $\ob$'s location---similar to the rescue of a person overboard in a practical scenario.} An efficient strategy \revisedtext{for the overall goal} is one that \emph{minimizes} the mission time, where all objects are tracked, $Q^*_\A=\argmin_{Q_\A} T ~ \textrm{s.t.} ~ \forall \ob \in \K ~ \mi{tracked}(\ob)$, or \emph{maximizes} the number of objects tracked, if a time limit elapsed. %

\section{Multi-agent Active Search and Tracking with Heterogeneous Information} \label{sec:main-approach}
Our proposed approach is online and follows a hybrid multi-agent architecture, where agents can operate autonomously, but share information with a central HQ to effectively coordinate. As agents get sensor measurements, their beliefs about cells occupied by targets and target locations are updated. Accordingly, we propose a multi-criteria optimization problem solved during search or tracking, finding the best trajectory for each agent---see \fig{system-architecture} for an overview of the architecture. \revisedtext{%
Our approach can search for and track all objects by reducing their uncertainty below the threshold, even with a small number of agents. This capability is due to the combination of proposed elements: time-varying uncertainty, shared belief, and third-party reporting, which will be explained in the following sections.}

\subsection{Belief Representation and Update}\label{sec:belief}
Two main belief representations are needed to achieve the search and track mission---note that there is also the pose estimate of each agent, which is handled by the probabilistic localization system running on-board the agent.

\textbf{Time-varying object occupancy map during search.}
We use a grid map covering the environment, where each cell is a square of fixed size, representing the belief $\beliefsearch{\ag}{\ce}{t}$ of the agents about whether a location $\ce$ is potentially occupied by a target object \revisedtext{at the current time $t$}. Discretizing the continuous belief distribution allows agents to reason in real time. Each cell can have a value between $0.0$ (no object) and $1.0$ (occupied by an object). $0.5$ means unknown with the highest uncertainty. Thus, %
$\beliefsearch{\ag}{\ce}{t}$ is a Bernoulli random variable that can be used for exploration.
Initially, for each agent $\ag\in \A$ and free location $\ce\in \M_{\mi{free}}$, $\beliefsearch{\ag}{\ce}{0}$ is a uniform distribution, before $\ob$ is detected by some $\ag$, as we assume we do not have any prior knowledge (\fig{example}-Initial belief). If prior knowledge is available, that initial distribution can be used.

As the search progresses, if a cell is within the FoV (explored), the uncertainty of the cell decreases, based on the sensor model for the actual detection $\zdet$ (\fig{example}-$t_0$). 

\begin{figure}[t!]
    \centering
    \begin{minipage}[t]{0.45\columnwidth}
        \begin{subfigure}[b]{\textwidth}            
            \includegraphics[width=\textwidth]{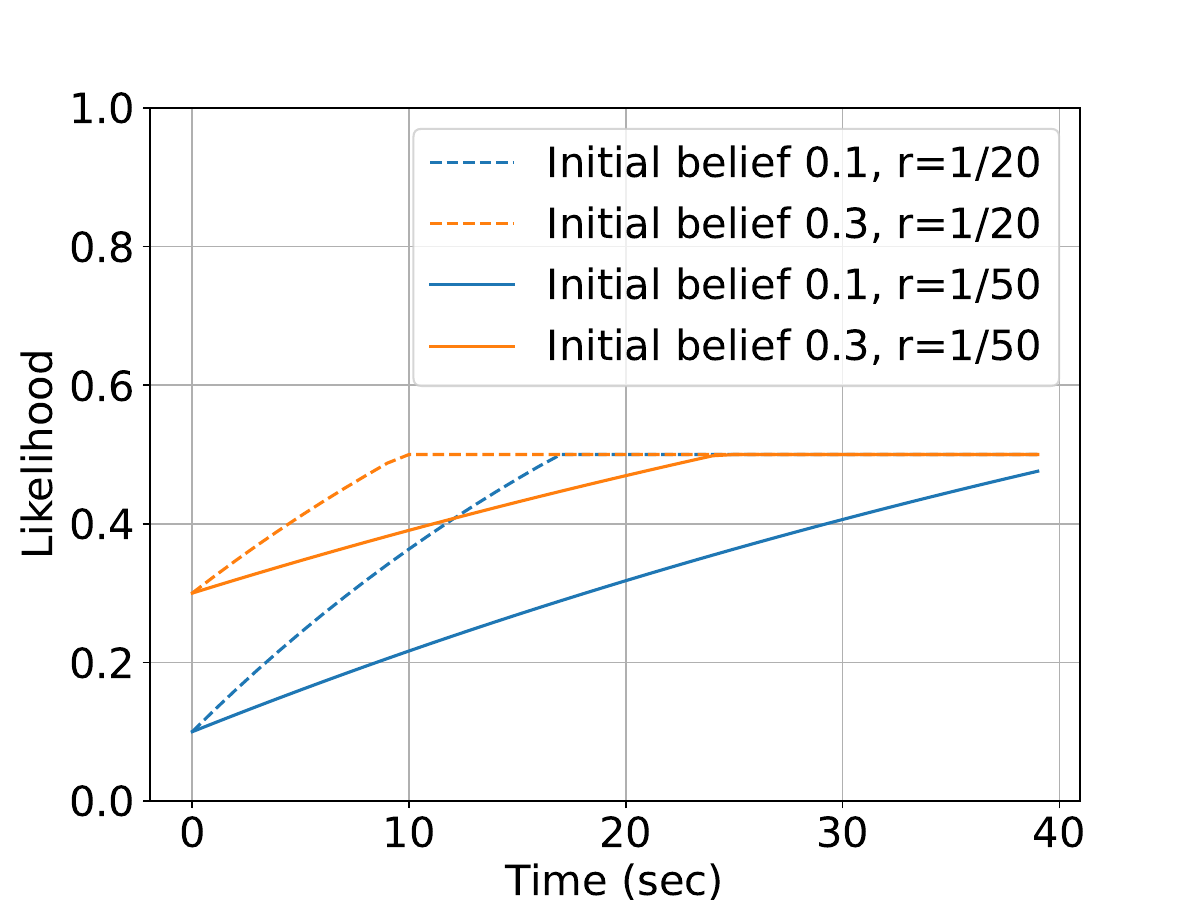}
        \end{subfigure}
    \end{minipage}
    \begin{minipage}[t]{0.45\columnwidth}
        \begin{subfigure}[b]{\textwidth}
            \includegraphics[width=\textwidth]{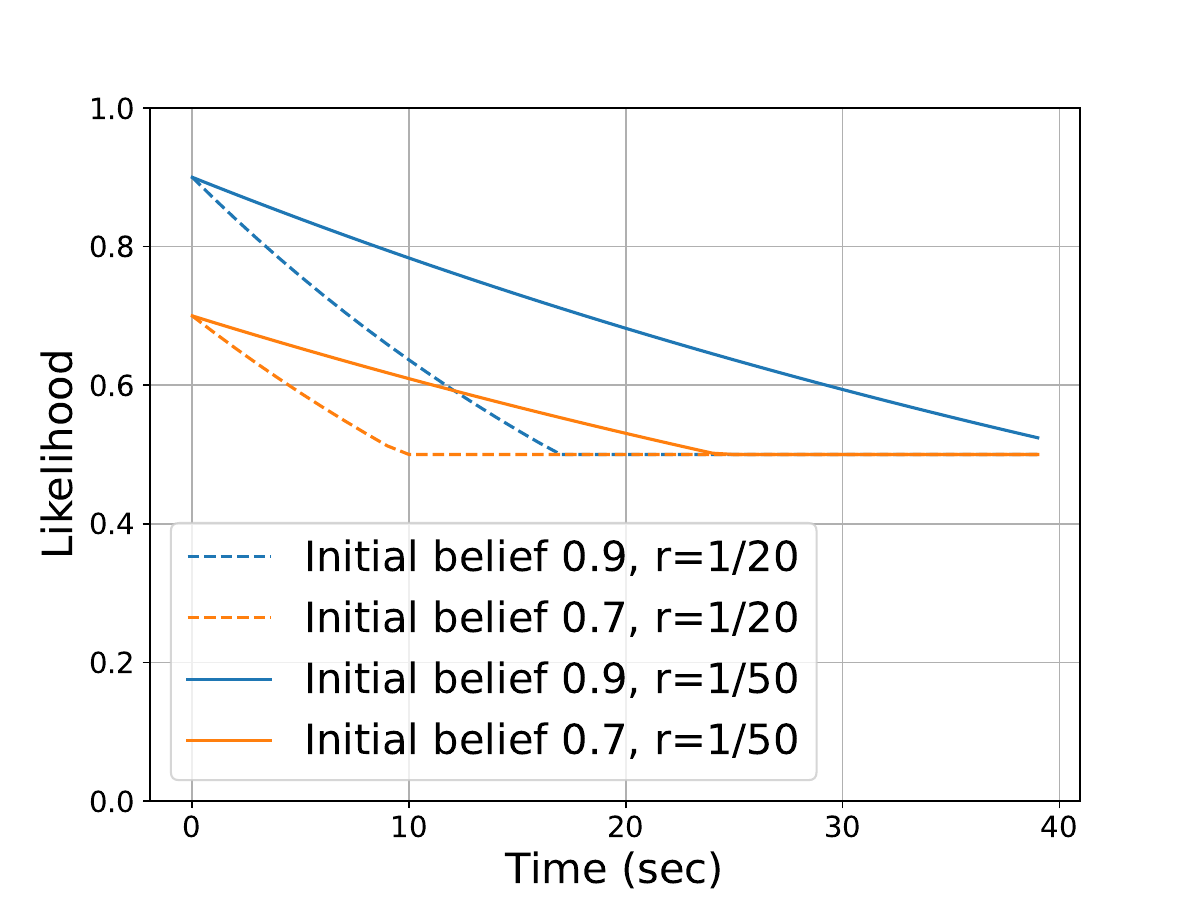}
        \end{subfigure}
    \end{minipage}
    \caption{Example of cell time-varying uncertainty: (left) case (1) without any target detected; (right) case (2) target detected but lost. For both cases, large $r$ value (dashed) represents a faster convergence than smaller $r$ (solid).}
    \label{fig:time-varying}
\end{figure}

As objects can move, we modeled temporal variations of uncertainty (of the cells that were explored previously) and are not currently in the FoV (\fig{example}-$t_1$,$t_2$). 
There are two conditions in which the uncertainty changes: 

\noindent \textbf{(1)} when the number of detected target objects, denoted $|\Kdetect|$, is less than the total number of objects (if known), denoted as $|\K|$, the uncertainty in cells outside FoV will increase as time elapses.  
As shown in \fig{time-varying}(left), %
we use a monotonically increasing exponential function in a visited cell $c$ where $\beliefsearch{\ag}{\ce}{t} < 0.5$, i.e., target was not found:
\begin{equation}
    \beliefsearch{\ag}{\ce}{t} = \min(0.5, \beliefsearch{\ag}{\ce}{t^{\mi{last}}} * e^{-r\Delta t})
\end{equation}
where $t^{\mi{last}}$ is the last time step when the cell $c$ was explored; $\beliefsearch{\ag}{\ce}{t^{\mi{last}}}$ is the belief at $t^{\mi{last}}$; $r\in \mathbb{R}_{>0}$ is a decay coefficient; $e$ is Euler's number but can be any constant greater than $1$; \revisedtext{and $\Delta t = t - {t^{\mi{last}}}$}. %
Intuitively, a large $r$ represents a fast convergence into the initial probability $0.5$, i.e., it becomes more rapidly uncertain than a smaller $r$. If $|\Kdetect| = |\K|$, i.e., once search and tracking have been completed, the uncertainty outside the field of view remains unchanged. \revisedtext{Note that this condition applies if $|\K|$ is known; otherwise, the agents will perform persistent search.}

\noindent \textbf{(2)} when a target object has already been detected but goes outside FoV, the uncertainty of the cells outside FoV will increase. For example, if agent $\ag$ detects a target object $\ob$ and the agent can be sure that the object remains stationary, the entropy of the corresponding cell does not increase. However, \revisedtext{assuming no knowledge on the target motion model}, if the target object is in motion, we use a time-varying function that follows a radial shape. In this case, we update the belief with a monotonically decreasing exponential function in a visited cell $\ce$ where $\beliefsearch{\ag}{\ce}{t} > 0.5$ (\fig{time-varying}(right)):%
\begin{equation}
    \beliefsearch{\ag}{\ce}{t} = \max(0.5, 1 - (1-\beliefsearch{\ag}{\ce}{t^{\mi{last}}})) * e^{-r\Delta t})
\end{equation}

\textbf{Probabilistic location estimate of an object during track.}
For each object $\ob$, we have the belief on their positions, which is conditioned on the locations \revisedtext{$q_\ag^{0:t}$ the agent $\ag$ visited  and corresponding sensor measurements $\zlocagzt$, from the beginning of the mission, time $0$, up to the current time $t$, $\belieftrack{\ag}{\ob}{t} = p(\ob^t|\zlocagzt, q_\ag^{0:t})$}.  %
This belief can be updated via Bayesian filtering\revisedtext{, i.e., \textit{prediction and update} steps}. \emph{Propagating} the belief after executing the action \revisedtext{(\textit{prediction step})} that will get the agent to $q_\ag^t$ means: 
\revisedtext{
$\obelieftrack{\ag}{\ob}{t} = \sum_{q_\ag^{t}} \, p(\ob^t|\ob^{t-1}, q_\ag^{t}) \cdot \belieftrack{\ag}{\ob}{t-1}$}
(Note that to simplify notation, we use $\ob$ to refer also to the object's location.). 
The agent then gets noisy sensor measurements and the belief distribution  $\obelieftrack{\ag}{\ob}{t}$ \revisedtext{ from the prediction step} can be \emph{updated} into the new belief $\belieftrack{\ag}{\ob}{t}$ \revisedtext{(\textit{update step})}. More formally, \revisedtext{$\belieftrack{\ag}{\ob}{t} = \eta \,\, p(\zlocagt) \cdot \obelieftrack{\ag}{\ob}{t}$, where $\eta$ is a normalizer and $\zlocagt$ is a sensor measurement of the target $\ob$ at time $t$ as per \eq{eq:track-loc-measure}. We include the full math derivation for the prediction and update steps in the Appendix.}

\paragraph{Third-party Reporting}
The location estimate provided by third-party reporters  (see \sect{prob-state}) based on their measurement $\zloce$ is represented as a separate belief $\ex{e}{\ob}{t}$ independent of the belief map $\belieftrack{\ag}{\ob}{t}$ by the agents (\fig{example}-$t_0$). This separation allows us to satisfy the condition that a target must be seen by some $\ag \in \A$ (it is not enough for a third-party to report seeing the object) and for increased flexibility on how to handle that information. 
This representation encodes different uncertainty from different external third-party reporters, thus the level of trustworthiness. This information can reduce the overall search time of the team. 

\paragraph{LSTM-based prediction} 
\revisedtext{To track the object effectively once it is searched for and detected, a robust model that predicts the object's future trajectory is required. Previous research on motion models of a target includes: a physics-based parametric model \cite{model-based-prediction-2010,model-based-prediction2-2006}, Bayesian filtering approaches such as Kalman Filter (KF) \cite{kalman-filter}, and data-driven approaches like Gaussian Process (GP) regression \cite{GPR-2006}.} A key distinction of our proposed approach from traditional Bayesian filtering lies in long-horizon prediction to provide a better understanding of the target object's intentions (\fig{example}-$t_1$). For instance, in scenarios where there is significant noise in the measurement data regarding $\ob$, or when the target object $\ob$ suddenly executes a turning maneuver, short-term approaches might fail to accurately predict the actual trajectory. Hence, this could cause agent $\ag$ to resume searching, leading to a higher likelihood of losing sight of $\ob$ compared to what would be expected with a long-term approach. \revisedtext{A physics-based model or GP regression can be used for long-horizon prediction. However, the accuracy of physics-based models is low when predicting non-smooth trajectories. For the GP model, there are several limitations \cite{gp-trajectory-nguyen-2024, gp-trajectory-IV-2018, gp-trajectory-iros-park-2020}: (1) the choice of kernel; (2) high computational complexity and difficulty in retraining with streaming data in real time; (3) the assumption of independence  between $x$ (latitude) and $y$ (longitude) coordinates.}

\revisedtext{Therefore, for reliable real-time long-horizon motion planning during object tracking}, we \revisedtext{choose} a \revisedtext{neural network} model, consisting of a Long Short-Term Memory (LSTM) layer---to capture the temporal dependency of the trajectory---and a multilayer perceptron (MLP) with three layers---to then output the corresponding prediction, based on the successful application in the trajectory prediction domain \cite{social-lstm-2016}. \revisedtext{(Note that developing a completely new model for trajectory prediction can be a research direction by itself and is out of scope of this paper; here we focus on how to integrate such predictions effectively in the decision.)}

The neural network takes as input, the location coordinates of an object over a time window, and provides as output, the next location coordinates over a subsequent time window. To have a general model that can work in different environments, we pre-processed and transformed the location coordinates so that the last coordinate of each observed (input) trajectory is centered at the origin $(0,0)$ \cite{LSTM-normalization-pedestrain-2022} and the model does not consider interaction pooling  layers \cite{social-lstm-2016,multi-agent-tensor-fusion-2019} given that the targets act following independent non-cooperative motions. 

 Based on the proposed normalization scheme, the model predicts in real time future trajectories according to the observed trajectories of the target objects, which are transformed back to the original global reference frame. 

\revisedtext{Each agent runs the LSTM module so that the agents can operate independently from each other and have a faster reaction when a target is detected.} %

\textbf{Shared Belief.}
Both types of beliefs, $\belsearch$ and $\beltrack$ which the agents build over time, can be shared among the agents and HQ, so that the search and track can proceed efficiently (\fig{example}). Inspired by \textit{A-opt} from \cite{Carrillo-shanon-renyi-entropy-2015}, we build the team's central belief about the occupancy map by shared information as follows: 
\begin{equation} \label{eq:shared-bel}
    \belsearch = \belsharedsearch{c}{t} = \sum_{\ag \in \A} w_\ag * \beliefsearch{\ag}{c}{t}
\end{equation}
where  \revisedtext{$w_\ag = \eta_{\mathcal{s}} \cdot \frac{\trace(\mathbf{cov^{-1}_\ag})}{\trace(\mathbf{cov^{-1}_{\max}})} \geq 0$ is the corresponding weight that represents the reliability of the shared information by $\ag$, $\trace(\cdot)$ is a trace (sum of the square of variances) of information matrix $\mathbf{cov^{-1}(\cdot)}$, $\trace(\mathbf{cov^{-1}_{\max}})$ is a maximum trace among all traces of agents $\A$ interpreted as the most reliable, and $\eta_{\mathcal{s}}$ is a normalization factor such that $\sum_{\ag \in \A} w_{\ag}=1$}. 
Intuitively, an agent with less uncertainty has a higher weight for the search belief than an agent with greater uncertainty. 

Similarly, the belief $\beltrack$ on the potential location of the detected objects can be fused, using the weight calculated with the trace of information matrices.
Such weights will quantify the trustworthiness of the agents' and third-party reports.

\subsection{Multi-Criteria Decision Making} \label{sec:best-action}

With the belief representations just described,  we formulate the following general multi-criteria optimization function to find the best trajectory that an agent should take:%
\begin{equation} \label{eq:main-opt}
    \begin{split}
    \tau^* &=  \argmax_{\tau \in \Psi} (w * \jexplore + (1-w) * \jexploit) \\ & s.t. \,\, D(\tau) \geq D_{\mi{thre}}
    \end{split}
\end{equation} 

\noindent where $\tau$ is a candidate trajectory that an agent can follow, belonging to a set $\Psi$ of feasible trajectories; $\jexplore$ and $\jexploit$ (formally introduced in the following)
represent the information gain from exploration and exploitation component, respectively, \revisedtext{during search or track task mode}, and both are normalized;
  $w \in [0,1]$ determines the balance between the exploration and exploitation terms; $D(\cdot)$ is an expected distance with other agents in the team, and $D_{\mi{thre}}$ is a distance threshold to maintain between agents to favor spread. We choose to have different weight values for search and tracking phases, so that during search exploration is favored to find objects as soon as possible, while during tracking, exploitation is prioritized to clear detected objects as soon as possible. Their values have been empirically set and it would be interesting as future research to adaptively find the best weights.  

\revisedtext{This constrained optimization indirectly solves the problem formulated in \sect{prob-state}:} as the locations of the objects are not known and the method is online, we cannot directly minimize the time to find the objects; but $\jexplore$ and $\jexploit$ are related proxies: maximizing the exploration information gain allows us to reduce the uncertainty of larger portions of the environment; likewise, maximizing the exploitation information gain allows us to further reduce the uncertainty on the location of the detected objects. \revisedtext{Because of the assumptions on the speed of the agents and targets, the time-varying uncertainty, and the tracking completion of targets once the uncertainty is below a specified threshold, the problem is feasible to solve with the proposed constrained optimization. To help with intuitive understanding of the proposed approach, a single-agent perspective task process is shown in \fig{system-architecture}b.} 

The set of candidate trajectories for each agent is generated based on the current location of an agent and whether the agent is in search or tracking mode.
Finding all feasible trajectories for exploration and exploitation is computationally intractable \cite{burgard2005coordinated,sampling-long-horizon-hollinger-2015}. Thus, we use a subset of feasible trajectories at a specific timestamp, as described in the following subsections, to reduce computational complexity and achieve the task reasonably well in real time. \fig{example}-$t_0$--$t_2$ shows a simplified example of the evolution of the search and track.

\subsubsection{Exploration} \label{sec:exploration}
We define $\jexplore$ so that trajectories directed toward uncertain areas, based on the current $\belsearch$, are prioritized:
\begin{flalign}
    \label{eq:info-gain-explore}
    \jexplore &= I(\belsearch|\tau) \\ 
    & = H(\belsearch|\mathbf{u,z}) - H(\belsearch|\mathbf{u,z, u_\tau, \hat{z}}) \nonumber 
\end{flalign}
where $I(\belsearch|\tau)$ is the information gain based on following trajectory $\tau$; $H(\belsearch|\mathbf{u,z})$ represents the current entropy of the belief representation used for search; $H(\belsearch|\mathbf{u,z, u_\tau, \hat{z}})$ is the expected entropy; $\mathbf{u,z}$ represents history of data consisting of control inputs and measurements; $\mathbf{u}_\tau=u_{t_c:t_c + T_L}$ is the action that makes the agent follow $\tau$ from current time $t_c$, look-ahead time $T_L$; and $\mathbf{\hat{z}}=z_{t_c:t_c + T_L}$ represents expected sensor measurements along  $\tau$.

The entropy for cell $c$ at time $t$ is: 
\begin{equation}
    H(\ce_t) = -p(\ce_t)\log\,p(\ce_t) - (1-p(\ce_t))\log\,(1-p(\ce_t))
\end{equation} where $H(\cdot)$ is the explored environment's entropy and $p(\ce_t)$ is the probability that  cell $\ce$ a\revisedtext{t} time $t$ is occupied by an object, denoted by $\belsharedsearch{\ce}{t}$. 
 Note that from the measurement $z_o$ at time $t$, we use $\ce_t = C(z_o)$ where $C(\cdot)$ is an operator to return cell $\ce$ to which $z_o$ belongs to. We define the total entropy of the map environment at time $t$  as $H(M_{t}) = \sum_{c\in \mathbf{M}_{\mi{free}}} H(\ce_t)$ where $M_{t}$ represents environment $\textbf{M}$ that has cells potentially occupied by a target object at time $t$, used for $H(\cdot)$ of the entire environment. 
 Because of the time-varying scenario, as described in \sect{belief}, the entropy can change over time even in areas already explored.

\subsubsection{Exploitation} \label{sec:exploitation}
We design $\jexploit$ to prioritize going towards detected targets to lower the uncertainty until the required threshold, based on $\beltrack$:
\begin{flalign}
\label{eq:info-gain-exploit}
\jexploit &= I(\Kdetect|\tau) \\
& = H(\Kdetect|\mathbf{u,z}) - H(\Kdetect|\mathbf{u,z, u_\tau, \hat{z}}) \nonumber \\
& = \sum_{\ob \in \Kdetect} H(\belsharedtrack{\ob}|\mathbf{u,z}) - H(\belsharedtrack{\ob}|\mathbf{u,z, u_\tau, \hat{z}}) \nonumber
\end{flalign}
\noindent where $I(\Kdetect|\tau)$ is the information gain for the detected target objects in $\Kdetect$ when following trajectory $\tau$ and is calculated using the entropy by Gaussian posterior \cite{Hausman-et-al-2015}, similarly to $\jexplore$. 

The agents complete the \revisedtext{\emph{tracking}} of an object $\ob$ after the detected target's localization error (trace of covariance) is within a threshold, i.e. $\trace(\Sigma_\ob) \leq \trace(\Sigma_{\mi{thre}})$, where $\trace(\Sigma_{\mi{thre}})$ is a desired threshold. \revisedtext{Once the condition is true, the object does not need to be tracked anymore.}

Note that we can obtain the posterior $p(\ob|z_{\ob})$ as a Gaussian distribution, given that both the likelihood and prior are Gaussian, with a sensor model as introduced previously. Hence, if $\ag$ follows $\tau$ with a positive information gain, the agent will experience a reduction in expected entropy, which is equivalent to a reduction in \textit{covariance}. \revisedtext{It is worth noting that if a target object moves out of the agent’s visibility range (denoted as $\mi{vr}_\ag$) while the tracking phase is active, the uncertainty about the target's position naturally worsens due to the missing update step (\sect{belief}). However, our method uses an exploration-biased action, i.e., \emph{search mode}, which enables the agents to relocate the target and reduce the uncertainty.}

During \revisedtext{\emph{search}}, $\jexploit$ is also considered, using the third-party report: trajectories that reduce the uncertainty of reported obstacles will be preferred.

\subsubsection{Trajectory generations and modes}
Agents operate in two modes: \emph{search}, if no object is assigned to an agent, or \emph{track}, if an object that has been detected is assigned to an agent.

\revisedtext{In \emph{search mode}, the trajectories \revisedtext{$\Psi$}} are found by identifying  \textit{frontiers} \revisedtext{$F$ (locations between the known and unknown areas of exploration on the time-varying map)}, i.e., $F = \{ \ce \in \M_\mi{free} ~ | ~  \belsharedsearch{c}{t} < \mi{unknownThreshold} ~ \& ~ \mi{hasUnknownCellNeighbor}(\ce) \}$---where $\mi{hasUnknownCellNeighbor}(\ce)$ returns true if there is a neighbor cell to $\ce$, which is considered unknown, i.e., $0.5\pm \mi{unknownThreshold}$. \revisedtext{More formally, $\Psi = \{\tau ~|~ \mi{Reachable}(\tau, \ag, c)\}$ where $\mi{Reachable}$ returns true if $c \in F$ is reachable from $\ag \in A$ by $\tau$.} To reduce the number of cells considered as frontiers, we cluster frontiers that form connected neighbors and pick the centroid of the cluster. \revisedtext{The trajectories are evaluated considering both  $\jexplore$ and $\jexploit$.}

In \emph{track mode}, \revisedtext{for a long-horizon planning that adopts LSTM-prediction introduced in \sect{belief}, we utilize a sampling-based approach in the velocity space starting from the current location of the agent.} %
\revisedtext{Like \emph{search mode},} these trajectories are \revisedtext{also} evaluated considering both $\jexploit$ and $\jexplore$, and discarding samples that would result in collisions. Sampling directly in the space of velocities allows agents to have better control in tracking an object.

In either mode, the multi-criteria optimization problem is solved, following \eq{eq:main-opt}. Different weights are used in either mode as mentioned above.

\subsection{Task Assignment and Execution of Action} \label{sec:task-assign-action-execute}

The overall task allocation is coordinated with the HQ, which has collected knowledge, including external information sources. \revisedtext{Not only relying on centralized decisions from HQ, our system is hybrid in that agents can operate in a decentralized manner for robustness as a fallback.} Our proposed method also operates asynchronously: the HQ consolidates exploration results collected by the agents and third parties, identifies a desired action at that time point, allows some agents to compute their utility, and makes a decision to allocate the task. This \revisedtext{hybrid and} asynchronous design fits well into real-world scenarios: there can still be agents actively performing previously assigned tasks \revisedtext{and they can continue to operate even if communication between HQ and agents breaks.} 
\begin{algorithm}[b!]
\caption{Tracking Task Assignment (HQ).}
\label{alg:track-task-assignment}
\begin{algorithmic}[1]
\renewcommand{\algorithmicrequire}{\textbf{Input:}}
\renewcommand{\algorithmicensure}{\textbf{Output:}}
    \scriptsize
    \REQUIRE \,
    \vspace{-1em}
    \begin{itemize}
        \itemsep-0.2em
        \item Agents set $\A$
        \item Target objects set $\K$
    \end{itemize} 
    \vspace{-0.5em}
    \ENSURE hash table $Q$ for optimal trajectory $\tau^*$ and corresponding $\ag_{\mi{opt}}$ assigned to $\tau^*$
    \STATE $Q$ $\gets$ \textrm{empty} 
    \STATE \HiLi $\revisedtext{\A_{\detected}}, \Kdetect$ $\gets$ $\texttt{getDetectReport}(\K, \A)$  \comment{agent set which detected $\revisedtext{\A_{\detected}}$, detected object set $\Kdetect$}
    \STATE \HiLi $\K_{\track}$ $\gets$ $\texttt{getObjectForTrack}(\K, \Kdetect, \revisedtext{\A_{\detected}})$ \comment{priority queue}
        \WHILE{$\K_{\track} \neq \emptyset $}
            \STATE $\ob_{\track}$ $\gets$ $\K_{\track}$\textrm{.pop()}
            \STATE $Q_{\track}$ $\gets$ \textrm{empty} \comment{hash table for tracking agent}
            \FOR{$\ag_{\detected} \in \revisedtext{\A_{\detected}}$}
                \IF{ \HiLiS \texttt{checkAssignAvailable}$(\ag_{\detected})$}
                    \STATE $\tau_{\ag_{\detected}}$ $\gets$ \texttt{requestBid}($\ag_{\detected}$) \comment{$\ag_{\detected}$ submits a bid}
                    \STATE $Q_{\track} \gets \textrm{update}(Q_{\track}, \tau_{\ag_{\detected}}, \ag_{\detected})$
                \ENDIF
            \ENDFOR
            \STATE \HiLi $\tau^*_{\ob_{\track}}, \ag_{\mi{opt}}$ $\gets$ $\texttt{getBestBid}(Q_{\track})$
            \STATE $Q$ $\gets$ $\textrm{update}(\tau^*_{\ob_{\track}}, \ag_{\mi{opt}})$
            \STATE $\A_{\mi{act}}$ $\gets$ $\A_{\mi{act}} \cup \{\ag_{\mi{opt}}\}$
        \ENDWHILE
\RETURN $Q$ 
\end{algorithmic} 
\end{algorithm}

\textbf{Search}: the HQ places bids for search to agents that are not assigned to any tracking task (described below). The choice of first assigning agents to tracking tasks reflects the priority on clearing objects as soon as they are detected. The unassigned agents submit their bids, and the HQ selects an agent with the best utility. \revisedtext{While the design of the approach is similar to the state-of-the-art auction method \cite{market-approach-survey-2006}, our proposed algorithm explicitly includes not only the typical $\jexplore$, but also $\jexploit$ of the third party reporting, which can bias to a target location, not within the field of view by the agents. The technical detail of the algorithm is included in the Appendix.}

\textbf{Tracking} (\alg{alg:track-task-assignment}): the HQ identifies the detection status of existing target objects and assigns agents to track the detected targets. \revisedtext{In \alg{alg:track-task-assignment}, we highlighted the novel parts, used differently from the conventional market-based approach in search and tracking problem.} More specifically, if any target is within an agent $\ag$'s sensor range $\mi{vr}_\ag$ and requires tracking, the HQ collects reported information (trace of covariance, distance, monitoring time) by the detected agent(s) \revisedtext{(\texttt{getDetectReport} in line 2). Unlike the literature, we use $\beltrack$ based on \eq{eq:shared-bel} to address the information coming from heterogeneous sources. Then, HQ identifies and prioritizes the target for tracking (\texttt{getObjectForTrack} in line 3)}, and select $\ag_{\mi{opt}}$ with the best utility calculated as in \sect{best-action} \revisedtext{(lines 4--11)}.   \revisedtext{Our proposed approach is advantageous because the HQ and agent(s) robustly handle the following multi-target, multi-agent scenarios:} \revisedtext{Case 1 -- \textit{multiple targets}:} if there are multiple targets within $\mi{vr}_\ag$, a target with the best expected utility (line 3), i.e., minimum expected covariance after an action, will be chosen for tracking \revisedtext{to first conduct the soon-to-be-completed tracking (line 5). Note that our algorithm explicitly includes $\jexploit$ with the heterogeneous sources, as well as $\jexplore$, even during tracking. Intuitively, our algorithm evaluates `\textit{in case}' scenarios where a tracked target goes missing, enabling the tracking agent to quickly adjust and resume the search, thereby increasing the probability of re-detecting the target. After the agent is assigned to a specific target, we include hysteresis \texttt{checkAssignAvailable} in line 8) to prevent oscillation occurring from switching the tracking assignment}; \revisedtext{Case 2 -- \textit{multiple agents}:} on the other hand, if there is one target seen \revisedtext{and reported} by multiple agents, the proposed algorithm does not constrain it to a single agent (lines 2--7, line 11) so that multiple agents can naturally cooperate to complete the tracking task  and enhance the team's task performance; \revisedtext{and Case 3 -- \textit{multiple targets and agents}: if there are multiple targets seen and reported by multiple agents, the algorithm applies the same principle as Case 1, 2. In other words, a target with the best utility is assigned to an agent with the best bid, and the same procedure is followed for the next target with the second-best utility. This greedy assignment can naturally handle the distribution of agents in a practical scenario and has a time complexity $O(n^2)$, which is better than the Hungarian algorithm~\cite{Kuhn1955Hungarian}--$O(n^3)$--used for task assignment where $n = \text{max}(|\Kdetect|, |\A_{\detected}|)$, with $\A_{\detected}$ agents that are in search mode. We left for future work a case where agents explicitly coordinate to encircle a target, which would be useful in an adversarial scenario. %
}

\textbf{Action execution.} Once an agent $\ag$ is assigned to an optimal trajectory $\tau^*$, the agent follows it in a closed-loop system. \revisedtext{Aside from trivial centralized approach, our hybrid system integrates agent's independent behavior for robustness (\alg{alg:agent-independent-assignment}) as a back-up and effective task completion. More specifically, during \emph{search}, if an agent does not receive frontier assignment by HQ (e.g., due to a communication failure, line 2), the agent evaluates its own frontier and follows an optimal trajectory (lines 3--5). If a target is detected by an agent in the \emph{search} mode, for fast reaction, the agent will immediately switch mode to \emph{tracking}, acting as a decentralized intelligence (lines 7--8), and inform the HQ (line 9), which could reassign the tracking of the detected object if there are better assignments as per \alg{alg:track-task-assignment}.}

\begin{algorithm}[t!]
\caption{Independent Action (Agent).}
\label{alg:agent-independent-assignment}
\begin{algorithmic}[1]
\renewcommand{\algorithmicrequire}{\textbf{Input:}}
\renewcommand{\algorithmicensure}{\textbf{Output:}}
    \scriptsize
    \REQUIRE \,
    \vspace{-1em}
    \begin{itemize}
        \itemsep-0.2em
        \item Agent task mode $TM$
        \item Agent's own frontier $F_a$
        \item Target objects set $\K$
    \end{itemize} 
    \vspace{-0.5em}
    \ENSURE optimal trajectory $\tau^*$
    \revisedtext{
    \IF{$TM == search$}
        \IF{$\neg \, \texttt{receiveAssignment()}$}
            \STATE $F_{\search a}$ $\gets$ $\texttt{getBestFrontiers}(F_a)$ \comment{agent's frontier for search}
            \STATE $fr_{\search a}$ $\gets$ $F_{\search a}$\textrm{.pop()} \comment{priority queue}
            \STATE $\tau^* \gets$ \texttt{getTrajectory($fr_{\search a}$)}
        \ELSIF{\texttt{detectObject()}}
            \STATE $\ob_{\track a}$ $\gets$ \texttt{getObjectForTrack}($\K, \Kdetect_a$)
            \STATE \texttt{switchToTrack()} \STATE \texttt{reportToHQ($\ob_{\track a}$)}
        \ENDIF
    \ELSIF{$TM == track$}
        \STATE $\tau^* \gets$ \texttt{getTrajectory($\ob_{\track a}$)}
    \ENDIF 
\RETURN $\tau^*$
}
\end{algorithmic} 
\end{algorithm}

\textbf{Parallelization.} 
We designed the search and tracking task assignment algorithm to run in parallel to achieve real-time decisions from the HQ and asynchronously process incoming data of detected target data and frontiers from agents. Furthermore, identifying a desired task and placing a bid can reduce the waste of resources, because not all agents need to compute all candidate tasks and desired trajectories to achieve the tasks.

\section{Experimental Results} \label{sec:experiments}
We performed extensive simulated experiments---\revisedtext{6870}, for a total of around \revisedtext{570} hours---within \revisedtext{robot simulators} under different scenarios. \revisedtext{To make the experiments realistic, our setting includes the following elements, which will be described in this section: 
\begin{itemize}
    \item Multiple heterogeneous agents with heterogeneous sensor configuration (e.g., sensor range, uncertainties);
    \item Real-world environment map embedded in the simulation (i.e., areas of Shanghai, Boston, and AWS office);
    \item Various combinations of multiple agents and targets;
    \item Asynchronous and real-time data streaming from heterogeneous sources;  
    \item Target movements based on real-world data collected by computational fluid dynamics particles; and
    \item 3D physics-enabled simulation including real-world like environments.
\end{itemize}
}
\noindent Our experiments compared with several baselines, included ablation studies, and evaluated with real-world data for target trajectories. Overall, our proposed approach showed consistently superior 
performance across the different experiments.

\subsection{Experimental setup}
Our extensive Monte Carlo simulations were conducted using Stage~\cite{vaughan2008massively}, a 2D robot simulator that includes sensor localization noise. 

We implemented our approach in Python and PyTorch, with the Robot Operating System (ROS) for portability. \emph{We will release our experimental data and code upon publication of this paper.} %
We implemented an LSTM-MLP architecture that takes as input the $10$ past coordinates and predicts a future trajectory with $15$ coordinates (same as our tracking trajectory evaluation length). The time frequency $dt$ for each coordinate is set to $\SI{0.2}{\second}$ (same as the sensor frequency for detecting a target). These values can be adjusted based on the expected speed of the targets. We used MSE loss function for training and ran $300$ epochs. We trained the model on a mix of real pedestrian dataset \cite{eth-dataset-iccv-2009} and a synthetic dataset representing potential arbitrary motions of objects. The synthetic dataset is composed of randomly generated trajectory in an environment not used in the experiments. 

For each agent to navigate in the environment without colliding with obstacles and each other, we use the opensource \texttt{navigation} stack, which includes a localization system, A* as global path planner, and 
time-elastic band local planner \cite{rosmann2017integrated}. 
We used a computer equipped with an 
AMD Ryzen 9 6900HX 8-core \SI{4.9}{GHz} processor, \SI{32}{GB} RAM, and NVIDIA GPU RTX 3080 Ti with \SI{16}{GB} memory.

\begin{figure}[!b]
\centering
\includegraphics[height=0.9in]{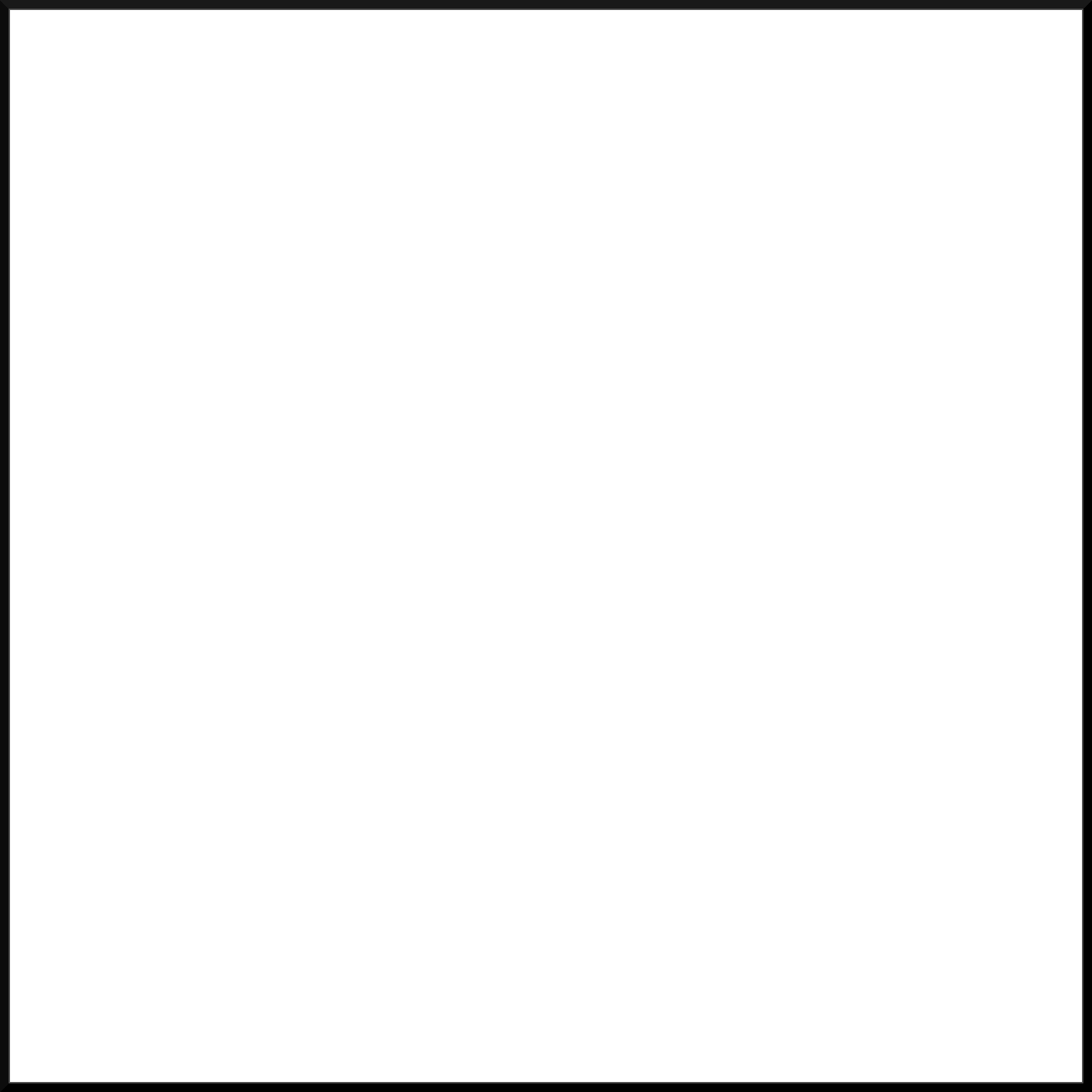}~~~
\includegraphics[height=0.9in,frame]{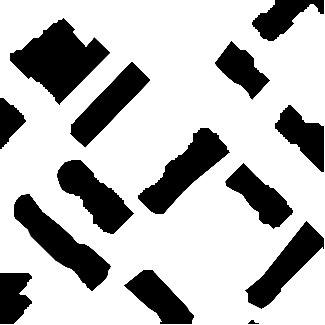}~~~
\includegraphics[height=0.9in,frame]{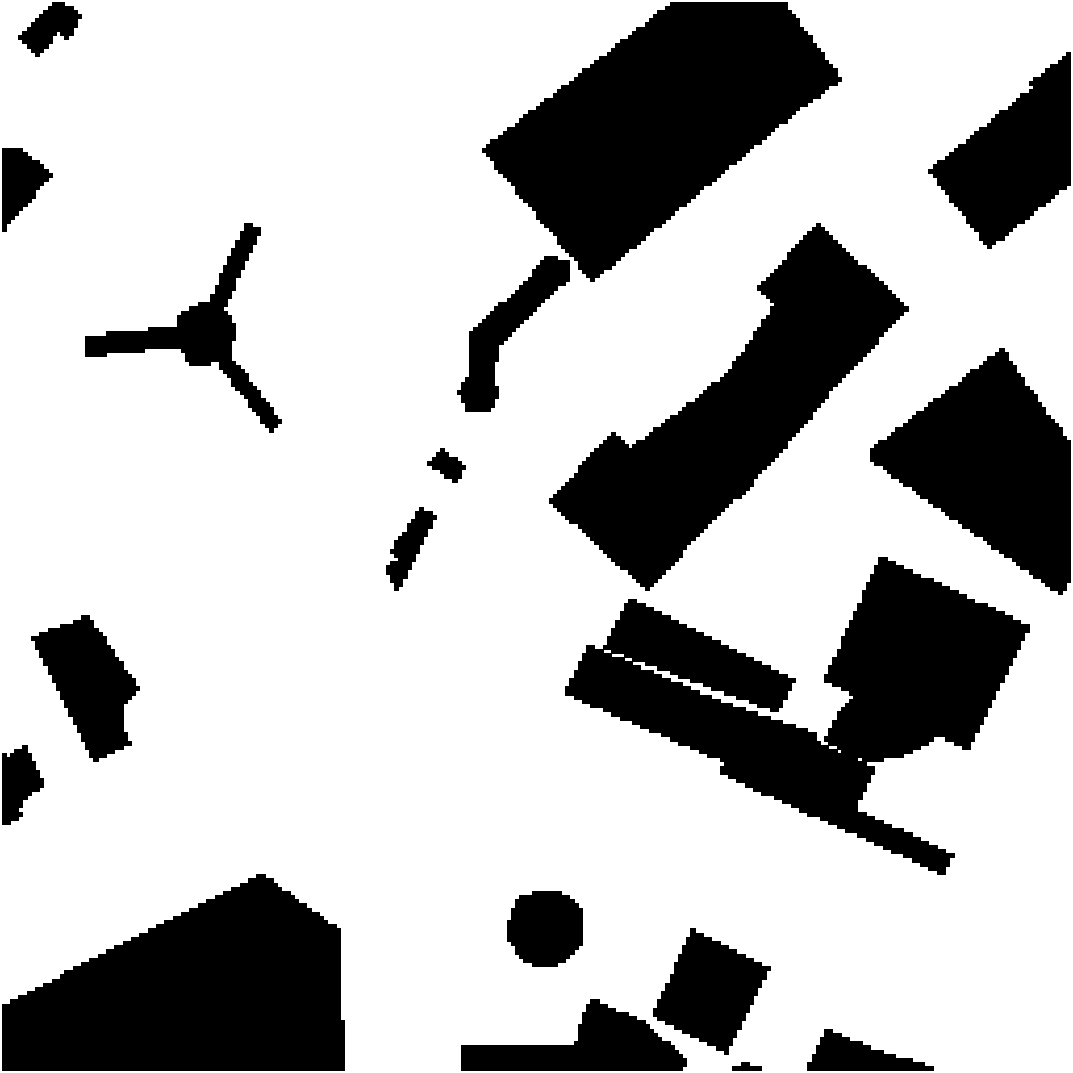}
\caption{Environments used for simulation: (left) \emph{open} $\sim$\qtyproduct{50 x 50}{m}, (mid, right) \emph{city1}---Boston, \emph{city2}---Shanghai $\sim$\qtyproduct{100 x 100}{m}.}\label{fig:environments}
\end{figure}

Experiments were conducted in three representative realistic environments within \qtyproduct{50 x 50}{m} and \qtyproduct{100 x 100}{m} (\fig{environments}) for quantitative analysis, an open space and two city environments (parts of Boston \revisedtext{in $42.33669$ N, $71.08012$ W} and Shanghai \revisedtext{in $31.24103$ N, $121.49693$ E}), from a pathfinding benchmark~\cite{pathfinding-benchmark}.

For agents in the set $\A$, we configured different numbers of agents $\{2,4,8\}$ with \textbf{(1)} homogeneous capabilities such as $s_\ag$ defined in \sect{prob-state}; and \textbf{(2)} heterogeneous capabilities such as $\mi{vr}_\ag$, detection probability and noise. Moreover, we configured different number of targets $\{2,5,10\}$ that follow 
non-cooperative motion. Non-cooperative motion is modeled as a random selection of waypoints in the freespace for the comparative analysis, except for the real-world scenario where the motion comes from a computational fluid dynamics model representing, e.g., floating objects in ocean. 

Three different initial sets of initial positions for agents and targets are randomly found.

Accordingly, an experimental configuration is defined by environment, number of agents, number of targets, initial set of positions, and method. Each experimental configuration is run $5$ or $10$ times---generally we didn't observe large variations of the obtained results across runs---and we performed \textit{t-test} \cite{pestman1998mathematical} to verify if there is a statistically significant difference in the results. 

We recorded the following quantitative metrics:
\begin{itemize}
    \item \emph{Complete mission time} (\SI{}{\second}), to fulfill the search and track requirement  across all the targets existing in the environment (covariance threshold met by at least one agent) as a core metric.
    \item \emph{Tracked objects ratio} (\%), if the agents didn't find and track all targets within  \SI{300}{\second};
    \item \emph{Tracking time} (\SI{}{\second}), averaged over all detected targets and agents (measured from detection until  threshold is met); 
    \item \emph{Traveled distance} (\SI{}{m}), averaged over all agents.
\end{itemize}
The proposed approach operated in real-time, thus we do not report the computation time. \revisedtext{Note that we limit the mission time cap to \SI{300}{\second} for practicability in real-world scenarios. However, with an unlimited time cap, our proposed method can achieve task completion as shown in other experiments below.}

\begin{table*}[!t]
\tablefontsize
\begin{subtable}[t]{\linewidth}
\centering
\resizebox{.89\linewidth}{!}{%
\begin{tabular}{|cc|cc|cc|cc|cc|cc|rr|}
\hline
\multicolumn{1}{|c|}{{\color[HTML]{000000} }} &
  {\color[HTML]{000000} } &
  \multicolumn{2}{c|}{{\color[HTML]{000000} \textbf{Ours}}} &
  \multicolumn{2}{c|}{{\color[HTML]{000000} \textbf{Central KF}}} &
  \multicolumn{2}{c|}{{\color[HTML]{000000} \textbf{Independent}}} &
  \multicolumn{2}{c|}{{\color[HTML]{000000} \textbf{Random}}} &
  \multicolumn{2}{c|}{\revisedtext{\textbf{Exhaustive}}} &
  \multicolumn{2}{c|}{\revisedtext{\textbf{Swarm}}} \\ \cmidrule{3-14} 
\multicolumn{1}{|c|}{\multirow{-2}{*}{{\color[HTML]{000000} \textbf{\begin{tabular}[c]{@{}c@{}}Agent \\ num\end{tabular}}}}} &
  \multirow{-2}{*}{{\color[HTML]{000000} \textbf{\begin{tabular}[c]{@{}c@{}}Target \\ num\end{tabular}}}} &
  \multicolumn{1}{c|}{{\color[HTML]{000000} mean}} &
  {\color[HTML]{000000} std.} &
  \multicolumn{1}{c|}{{\color[HTML]{000000} mean}} &
  {\color[HTML]{000000} std.} &
  \multicolumn{1}{c|}{{\color[HTML]{000000} mean}} &
  {\color[HTML]{000000} std.} &
  \multicolumn{1}{c|}{{\color[HTML]{000000} mean}} &
  {\color[HTML]{000000} std.} &
  \multicolumn{1}{c|}{mean} &
  std. &
  \multicolumn{1}{c|}{mean} &
  \multicolumn{1}{c|}{std.} \\ \hline
\multicolumn{1}{|c|}{{\color[HTML]{000000} }} &
  {\color[HTML]{000000} 2} &
  \multicolumn{1}{c|}{{\color[HTML]{009901} \textbf{134.00}}} &
  {\color[HTML]{000000} 38.13} &
  \multicolumn{1}{c|}{{\color[HTML]{F8A102} 148.32}} &
  {\color[HTML]{000000} 74.72} &
  \multicolumn{1}{c|}{{\color[HTML]{000000} 258.68}} &
  {\color[HTML]{000000} 63.26} &
  \multicolumn{1}{c|}{{\color[HTML]{000000} 217.92}} &
  {\color[HTML]{000000} 75.27} &
  \multicolumn{1}{c|}{247.60} &
  57.50 &
  \multicolumn{1}{r|}{275.62} &
  43.87 \\ \cmidrule{2-14} 
\multicolumn{1}{|c|}{{\color[HTML]{000000} }} &
  {\color[HTML]{000000} 5} &
  \multicolumn{1}{c|}{{\color[HTML]{009901} \textbf{260.68}}} &
  {\color[HTML]{000000} 43.62} &
  \multicolumn{1}{c|}{{\color[HTML]{000000} 289.40}} &
  {\color[HTML]{000000} 32.75} &
  \multicolumn{1}{c|}{{\color[HTML]{000000} 300.00}} &
  {\color[HTML]{000000} 0.00} &
  \multicolumn{1}{c|}{{\color[HTML]{F8A102} 282.96}} &
  {\color[HTML]{000000} 27.69} &
  \multicolumn{1}{c|}{300.00} &
  0.00 &
  \multicolumn{1}{r|}{300.00} &
  0.00 \\ \cmidrule{2-14} 
\multicolumn{1}{|c|}{\multirow{-3}{*}{{\color[HTML]{000000} 2}}} &
  {\color[HTML]{000000} 10} &
  \multicolumn{1}{c|}{{\color[HTML]{009901} \textbf{294.96}}} &
  {\color[HTML]{000000} 15.94} &
  \multicolumn{1}{c|}{{\color[HTML]{000000} 300.00}} &
  {\color[HTML]{000000} 0.00} &
  \multicolumn{1}{c|}{{\color[HTML]{000000} 300.00}} &
  {\color[HTML]{000000} 0.00} &
  \multicolumn{1}{c|}{{\color[HTML]{000000} 300.00}} &
  {\color[HTML]{000000} 0.00} &
  \multicolumn{1}{c|}{300.00} &
  0.00 &
  \multicolumn{1}{r|}{300.00} &
  0.00 \\ \hline
\multicolumn{1}{|c|}{{\color[HTML]{000000} }} &
  {\color[HTML]{000000} 2} &
  \multicolumn{1}{c|}{{\color[HTML]{009901} \textbf{84.52}}} &
  {\color[HTML]{000000} 12.45} &
  \multicolumn{1}{c|}{{\color[HTML]{000000} 232.04}} &
  {\color[HTML]{000000} 48.66} &
  \multicolumn{1}{c|}{{\color[HTML]{000000} 173.68}} &
  {\color[HTML]{000000} 115.70} &
  \multicolumn{1}{c|}{{\color[HTML]{000000} 160.54}} &
  {\color[HTML]{000000} 72.46} &
  \multicolumn{1}{c|}{{\color[HTML]{F8A102} 146.06}} &
  102.93 &
  \multicolumn{1}{r|}{282.22} &
  31.81 \\ \cmidrule{2-14} 
\multicolumn{1}{|c|}{{\color[HTML]{000000} }} &
  {\color[HTML]{000000} 5} &
  \multicolumn{1}{c|}{{\color[HTML]{009901} \textbf{127.98}}} &
  {\color[HTML]{000000} 36.55} &
  \multicolumn{1}{c|}{{\color[HTML]{000000} 222.38}} &
  {\color[HTML]{000000} 50.83} &
  \multicolumn{1}{c|}{{\color[HTML]{000000} 197.10}} &
  {\color[HTML]{000000} 80.85} &
  \multicolumn{1}{c|}{{\color[HTML]{000000} 261.00}} &
  {\color[HTML]{000000} 51.99} &
  \multicolumn{1}{c|}{{\color[HTML]{F8A102} 141.40}} &
  59.28 &
  \multicolumn{1}{r|}{289.66} &
  31.73 \\ \cmidrule{2-14} 
\multicolumn{1}{|c|}{\multirow{-3}{*}{{\color[HTML]{000000} 4}}} &
  {\color[HTML]{000000} 10} &
  \multicolumn{1}{c|}{{\color[HTML]{009901} \textbf{232.84}}} &
  {\color[HTML]{000000} 51.10} &
  \multicolumn{1}{c|}{{\color[HTML]{000000} 280.34}} &
  {\color[HTML]{000000} 41.60} &
  \multicolumn{1}{c|}{{\color[HTML]{000000} 273.04}} &
  {\color[HTML]{000000} 35.14} &
  \multicolumn{1}{c|}{{\color[HTML]{F8A102} 251.24}} &
  {\color[HTML]{000000} 73.50} &
  \multicolumn{1}{c|}{297.92} &
  6.58 &
  \multicolumn{1}{r|}{300.00} &
  0.00 \\ \hline
\multicolumn{1}{|c|}{{\color[HTML]{000000} }} &
  {\color[HTML]{000000} 2} &
  \multicolumn{1}{c|}{{\color[HTML]{009901} \textbf{59.96}}} &
  {\color[HTML]{000000} 9.86} &
  \multicolumn{1}{c|}{{\color[HTML]{000000} 119.68}} &
  {\color[HTML]{000000} 35.56} &
  \multicolumn{1}{c|}{{\color[HTML]{000000} 95.70}} &
  {\color[HTML]{000000} 37.62} &
  \multicolumn{1}{c|}{{\color[HTML]{000000} 210.00}} &
  {\color[HTML]{000000} 71.71} &
  \multicolumn{1}{c|}{{\color[HTML]{F8A102} 69.56}} &
  8.06 &
  \multicolumn{1}{r|}{243.02} &
  51.80 \\ \cmidrule{2-14} 
\multicolumn{1}{|c|}{{\color[HTML]{000000} }} &
  {\color[HTML]{000000} 5} &
  \multicolumn{1}{c|}{{\color[HTML]{009901} \textbf{76.90}}} &
  {\color[HTML]{000000} 8.68} &
  \multicolumn{1}{c|}{{\color[HTML]{F8A102} 107.74}} &
  {\color[HTML]{000000} 17.42} &
  \multicolumn{1}{c|}{{\color[HTML]{000000} 145.84}} &
  {\color[HTML]{000000} 60.03} &
  \multicolumn{1}{c|}{{\color[HTML]{000000} 211.48}} &
  {\color[HTML]{000000} 54.79} &
  \multicolumn{1}{c|}{139.46} &
  32.33 &
  \multicolumn{1}{r|}{279.36} &
  33.67 \\ \cmidrule{2-14} 
\multicolumn{1}{|c|}{\multirow{-3}{*}{{\color[HTML]{000000} 8}}} &
  {\color[HTML]{000000} 10} &
  \multicolumn{1}{c|}{{\color[HTML]{009901} \textbf{113.58}}} &
  {\color[HTML]{000000} 31.62} &
  \multicolumn{1}{c|}{{\color[HTML]{000000} 183.96}} &
  {\color[HTML]{000000} 83.01} &
  \multicolumn{1}{c|}{{\color[HTML]{000000} 231.50}} &
  {\color[HTML]{000000} 46.12} &
  \multicolumn{1}{c|}{{\color[HTML]{000000} 197.84}} &
  {\color[HTML]{000000} 55.55} &
  \multicolumn{1}{c|}{{\color[HTML]{F8A102} 151.14}} &
  49.90 &
  \multicolumn{1}{r|}{277.10} &
  43.08 \\ \hline
\multicolumn{2}{|c|}{average} &
  \multicolumn{1}{c|}{{\color[HTML]{000000} \textbf{153.94}}} &
  27.55 &
  \multicolumn{1}{c|}{209.32} &
  42.73 &
  \multicolumn{1}{c|}{219.50} &
  48.75 &
  \multicolumn{1}{c|}{232.55} &
  53.66 &
  \multicolumn{1}{c|}{199.24} &
  35.18 &
  \multicolumn{1}{r|}{283.00} &
  26.22 \\ \hline
\end{tabular}%
}
\vspace{-0.1cm}
\caption{Open}
\end{subtable}

\begin{subtable}[t]{\linewidth}
\centering
\resizebox{.75\linewidth}{!}{%
\begin{tabular}{|cc|rr|rr|rr|rr|cc|}
\hline
\multicolumn{1}{|c|}{{\color[HTML]{000000} }} &
  {\color[HTML]{000000} } &
  \multicolumn{2}{c|}{{\color[HTML]{000000} \textbf{Ours}}} &
  \multicolumn{2}{c|}{{\color[HTML]{000000} \textbf{Central KF}}} &
  \multicolumn{2}{c|}{{\color[HTML]{000000} \textbf{Independent}}} &
  \multicolumn{2}{c|}{{\color[HTML]{000000} \textbf{Random}}} &
  \multicolumn{2}{c|}{\revisedtext{\textbf{Exhaustive}}} \\ \cmidrule{3-12} 
\multicolumn{1}{|c|}{\multirow{-2}{*}{{\color[HTML]{000000} \textbf{\begin{tabular}[c]{@{}c@{}}Agent \\ num\end{tabular}}}}} &
  \multirow{-2}{*}{{\color[HTML]{000000} \textbf{\begin{tabular}[c]{@{}c@{}}Target \\ num\end{tabular}}}} &
  \multicolumn{1}{c|}{{\color[HTML]{000000} mean}} &
  \multicolumn{1}{c|}{{\color[HTML]{000000} std.}} &
  \multicolumn{1}{c|}{{\color[HTML]{000000} mean}} &
  \multicolumn{1}{c|}{{\color[HTML]{000000} std.}} &
  \multicolumn{1}{c|}{{\color[HTML]{000000} mean}} &
  \multicolumn{1}{c|}{{\color[HTML]{000000} std.}} &
  \multicolumn{1}{c|}{{\color[HTML]{000000} mean}} &
  \multicolumn{1}{c|}{{\color[HTML]{000000} std.}} &
  \multicolumn{1}{c|}{mean} &
  std. \\ \hline
\multicolumn{1}{|c|}{{\color[HTML]{000000} }} &
  {\color[HTML]{000000} 2} &
  \multicolumn{1}{r|}{{\color[HTML]{009901} \textbf{208.62}}} &
  {\color[HTML]{000000} 81.88} &
  \multicolumn{1}{r|}{{\color[HTML]{000000} 243.66}} &
  {\color[HTML]{000000} 40.56} &
  \multicolumn{1}{r|}{{\color[HTML]{000000} 279.12}} &
  {\color[HTML]{000000} 36.59} &
  \multicolumn{1}{r|}{{\color[HTML]{F8A102} 242.42}} &
  {\color[HTML]{000000} 62.77} &
  \multicolumn{1}{c|}{300.00} &
  0.00 \\ \cmidrule{2-12} 
\multicolumn{1}{|c|}{{\color[HTML]{000000} }} &
  {\color[HTML]{000000} 5} &
  \multicolumn{1}{r|}{{\color[HTML]{009901} \textbf{286.26}}} &
  {\color[HTML]{000000} 26.99} &
  \multicolumn{1}{r|}{{\color[HTML]{000000} 294.34}} &
  {\color[HTML]{000000} 17.90} &
  \multicolumn{1}{r|}{{\color[HTML]{000000} 300.00}} &
  {\color[HTML]{000000} 0.00} &
  \multicolumn{1}{r|}{{\color[HTML]{F8A102} 293.60}} &
  {\color[HTML]{000000} 18.04} &
  \multicolumn{1}{c|}{300.00} &
  0.00 \\ \cmidrule{2-12} 
\multicolumn{1}{|c|}{\multirow{-3}{*}{{\color[HTML]{000000} 2}}} &
  {\color[HTML]{000000} 10} &
  \multicolumn{1}{r|}{{\color[HTML]{000000} 300.00}} &
  {\color[HTML]{000000} 0.00} &
  \multicolumn{1}{r|}{{\color[HTML]{000000} 300.00}} &
  {\color[HTML]{000000} 0.00} &
  \multicolumn{1}{r|}{{\color[HTML]{000000} 300.00}} &
  {\color[HTML]{000000} 0.00} &
  \multicolumn{1}{r|}{{\color[HTML]{000000} 300.00}} &
  {\color[HTML]{000000} 0.00} &
  \multicolumn{1}{c|}{300.00} &
  0.00 \\ \hline
\multicolumn{1}{|c|}{{\color[HTML]{000000} }} &
  {\color[HTML]{000000} 2} &
  \multicolumn{1}{r|}{{\color[HTML]{009901} \textbf{118.74}}} &
  {\color[HTML]{000000} 19.79} &
  \multicolumn{1}{r|}{{\color[HTML]{F8A102} 132.80}} &
  {\color[HTML]{000000} 60.90} &
  \multicolumn{1}{r|}{{\color[HTML]{000000} 250.38}} &
  {\color[HTML]{000000} 104.65} &
  \multicolumn{1}{r|}{{\color[HTML]{000000} 243.50}} &
  {\color[HTML]{000000} 53.50} &
  \multicolumn{1}{c|}{290.88} &
  14.80 \\ \cmidrule{2-12} 
\multicolumn{1}{|c|}{{\color[HTML]{000000} }} &
  {\color[HTML]{000000} 5} &
  \multicolumn{1}{r|}{{\color[HTML]{009901} \textbf{137.20}}} &
  {\color[HTML]{000000} 16.98} &
  \multicolumn{1}{r|}{{\color[HTML]{000000} 288.66}} &
  {\color[HTML]{000000} 33.28} &
  \multicolumn{1}{r|}{{\color[HTML]{000000} 299.12}} &
  {\color[HTML]{000000} 2.78} &
  \multicolumn{1}{r|}{{\color[HTML]{F8A102} 230.16}} &
  {\color[HTML]{000000} 60.18} &
  \multicolumn{1}{c|}{300.00} &
  0.00 \\ \cmidrule{2-12} 
\multicolumn{1}{|c|}{\multirow{-3}{*}{{\color[HTML]{000000} 4}}} &
  {\color[HTML]{000000} 10} &
  \multicolumn{1}{r|}{{\color[HTML]{000000} 300.00}} &
  {\color[HTML]{000000} 0.00} &
  \multicolumn{1}{r|}{{\color[HTML]{000000} 300.00}} &
  {\color[HTML]{000000} 0.00} &
  \multicolumn{1}{r|}{{\color[HTML]{000000} 300.00}} &
  {\color[HTML]{000000} 0.00} &
  \multicolumn{1}{r|}{{\color[HTML]{000000} 300.00}} &
  {\color[HTML]{000000} 0.00} &
  \multicolumn{1}{c|}{300.00} &
  0.00 \\ \hline
\multicolumn{1}{|c|}{{\color[HTML]{000000} }} &
  {\color[HTML]{000000} 2} &
  \multicolumn{1}{r|}{{\color[HTML]{009901} \textbf{102.06}}} &
  {\color[HTML]{000000} 19.13} &
  \multicolumn{1}{r|}{{\color[HTML]{000000} 129.02}} &
  {\color[HTML]{000000} 73.65} &
  \multicolumn{1}{r|}{{\color[HTML]{000000} 221.68}} &
  {\color[HTML]{000000} 102.16} &
  \multicolumn{1}{r|}{{\color[HTML]{000000} 129.70}} &
  {\color[HTML]{000000} 39.01} &
  \multicolumn{1}{c|}{{\color[HTML]{F8A102} 118.54}} &
  9.30 \\ \cmidrule{2-12} 
\multicolumn{1}{|c|}{{\color[HTML]{000000} }} &
  {\color[HTML]{000000} 5} &
  \multicolumn{1}{r|}{{\color[HTML]{009901} \textbf{136.29}}} &
  {\color[HTML]{000000} 25.51} &
  \multicolumn{1}{r|}{{\color[HTML]{000000} 215.62}} &
  {\color[HTML]{000000} 67.37} &
  \multicolumn{1}{r|}{{\color[HTML]{000000} 283.82}} &
  {\color[HTML]{000000} 33.10} &
  \multicolumn{1}{r|}{{\color[HTML]{F8A102} 200.32}} &
  {\color[HTML]{000000} 46.81} &
  \multicolumn{1}{c|}{278.18} &
  29.12 \\ \cmidrule{2-12} 
\multicolumn{1}{|c|}{\multirow{-3}{*}{{\color[HTML]{000000} 8}}} &
  {\color[HTML]{000000} 10} &
  \multicolumn{1}{r|}{{\color[HTML]{009901} \textbf{299.27}}} &
  {\color[HTML]{000000} 2.41} &
  \multicolumn{1}{r|}{{\color[HTML]{000000} 300.00}} &
  {\color[HTML]{000000} 0.00} &
  \multicolumn{1}{r|}{{\color[HTML]{000000} 300.00}} &
  {\color[HTML]{000000} 0.00} &
  \multicolumn{1}{r|}{{\color[HTML]{000000} 300.00}} &
  {\color[HTML]{000000} 0.00} &
  \multicolumn{1}{c|}{300.00} &
  0.00 \\ \hline
\multicolumn{2}{|c|}{{\color[HTML]{000000} average}} &
  \multicolumn{1}{r|}{{\color[HTML]{000000} \textbf{209.83}}} &
  {\color[HTML]{000000} 21.41} &
  \multicolumn{1}{r|}{{\color[HTML]{000000} 244.90}} &
  {\color[HTML]{000000} 32.63} &
  \multicolumn{1}{r|}{{\color[HTML]{000000} 281.57}} &
  {\color[HTML]{000000} 31.03} &
  \multicolumn{1}{r|}{{\color[HTML]{000000} 248.86}} &
  {\color[HTML]{000000} 31.15} &
  \multicolumn{1}{c|}{276.40} &
  5.91 \\ \hline
\end{tabular}%
}
\vspace{-0.1cm}
\caption{City1}
\end{subtable}

\begin{subtable}[t]{\linewidth}
\centering
\resizebox{.75\linewidth}{!}{%
\begin{tabular}{|cc|rr|rr|rr|rr|cc|}
\hline
\multicolumn{1}{|c|}{{\color[HTML]{000000} }} &
  {\color[HTML]{000000} } &
  \multicolumn{2}{c|}{{\color[HTML]{000000} \textbf{Ours}}} &
  \multicolumn{2}{c|}{{\color[HTML]{000000} \textbf{Central KF}}} &
  \multicolumn{2}{c|}{{\color[HTML]{000000} \textbf{Independent}}} &
  \multicolumn{2}{c|}{{\color[HTML]{000000} \textbf{Random}}} &
  \multicolumn{2}{c|}{\revisedtext{\textbf{Exhaustive}}} \\ \cmidrule{3-12} 
\multicolumn{1}{|c|}{\multirow{-2}{*}{{\color[HTML]{000000} \textbf{\begin{tabular}[c]{@{}c@{}}Agent \\ num\end{tabular}}}}} &
  \multirow{-2}{*}{{\color[HTML]{000000} \textbf{\begin{tabular}[c]{@{}c@{}}Target\\ num\end{tabular}}}} &
  \multicolumn{1}{c|}{{\color[HTML]{000000} mean}} &
  \multicolumn{1}{c|}{{\color[HTML]{000000} std.}} &
  \multicolumn{1}{c|}{{\color[HTML]{000000} mean}} &
  \multicolumn{1}{c|}{{\color[HTML]{000000} std.}} &
  \multicolumn{1}{c|}{{\color[HTML]{000000} mean}} &
  \multicolumn{1}{c|}{{\color[HTML]{000000} std.}} &
  \multicolumn{1}{c|}{{\color[HTML]{000000} mean}} &
  \multicolumn{1}{c|}{{\color[HTML]{000000} std.}} &
  \multicolumn{1}{c|}{mean} &
  std. \\ \hline
\multicolumn{1}{|c|}{{\color[HTML]{000000} }} &
  {\color[HTML]{000000} 2} &
  \multicolumn{1}{r|}{{\color[HTML]{009901} \textbf{42.89}}} &
  {\color[HTML]{000000} 15.98} &
  \multicolumn{1}{r|}{{\color[HTML]{000000} 154.96}} &
  {\color[HTML]{000000} 33.04} &
  \multicolumn{1}{r|}{{\color[HTML]{000000} 300.00}} &
  {\color[HTML]{000000} 0.00} &
  \multicolumn{1}{r|}{{\color[HTML]{F8A102} 121.93}} &
  {\color[HTML]{000000} 23.55} &
  \multicolumn{1}{c|}{300.00} &
  0.00 \\ \cmidrule{2-12} 
\multicolumn{1}{|c|}{{\color[HTML]{000000} }} &
  {\color[HTML]{000000} 5} &
  \multicolumn{1}{r|}{{\color[HTML]{009901} \textbf{269.32}}} &
  {\color[HTML]{000000} 27.28} &
  \multicolumn{1}{r|}{{\color[HTML]{000000} 300.00}} &
  {\color[HTML]{000000} 0.00} &
  \multicolumn{1}{r|}{{\color[HTML]{000000} 300.00}} &
  {\color[HTML]{000000} 0.00} &
  \multicolumn{1}{r|}{{\color[HTML]{000000} 300.00}} &
  {\color[HTML]{000000} 0.00} &
  \multicolumn{1}{c|}{300.00} &
  0.00 \\ \cmidrule{2-12} 
\multicolumn{1}{|c|}{\multirow{-3}{*}{{\color[HTML]{000000} 2}}} &
  {\color[HTML]{000000} 10} &
  \multicolumn{1}{r|}{{\color[HTML]{000000} 300.00}} &
  {\color[HTML]{000000} 0.00} &
  \multicolumn{1}{r|}{{\color[HTML]{000000} 300.00}} &
  {\color[HTML]{000000} 0.00} &
  \multicolumn{1}{r|}{{\color[HTML]{000000} 300.00}} &
  {\color[HTML]{000000} 0.00} &
  \multicolumn{1}{r|}{{\color[HTML]{000000} 300.00}} &
  {\color[HTML]{000000} 0.00} &
  \multicolumn{1}{c|}{300.00} &
  0.00 \\ \hline
\multicolumn{1}{|c|}{{\color[HTML]{000000} }} &
  {\color[HTML]{000000} 2} &
  \multicolumn{1}{r|}{{\color[HTML]{009901} \textbf{57.61}}} &
  {\color[HTML]{000000} 25.73} &
  \multicolumn{1}{r|}{{\color[HTML]{000000} 204.68}} &
  {\color[HTML]{000000} 40.67} &
  \multicolumn{1}{r|}{{\color[HTML]{000000} 190.24}} &
  {\color[HTML]{000000} 144.25} &
  \multicolumn{1}{r|}{{\color[HTML]{F8A102} 127.41}} &
  {\color[HTML]{000000} 26.80} &
  \multicolumn{1}{c|}{300.00} &
  0.00 \\ \cmidrule{2-12} 
\multicolumn{1}{|c|}{{\color[HTML]{000000} }} &
  {\color[HTML]{000000} 5} &
  \multicolumn{1}{r|}{{\color[HTML]{009901} \textbf{171.60}}} &
  {\color[HTML]{000000} 73.81} &
  \multicolumn{1}{r|}{{\color[HTML]{F8A102} 289.36}} &
  {\color[HTML]{000000} 19.63} &
  \multicolumn{1}{r|}{{\color[HTML]{000000} 295.52}} &
  {\color[HTML]{000000} 10.01} &
  \multicolumn{1}{r|}{{\color[HTML]{000000} 300.00}} &
  {\color[HTML]{000000} 0.00} &
  \multicolumn{1}{c|}{300.00} &
  0.00 \\ \cmidrule{2-12} 
\multicolumn{1}{|c|}{\multirow{-3}{*}{{\color[HTML]{000000} 4}}} &
  {\color[HTML]{000000} 10} &
  \multicolumn{1}{r|}{{\color[HTML]{009901} \textbf{257.84}}} &
  {\color[HTML]{000000} 35.88} &
  \multicolumn{1}{r|}{{\color[HTML]{000000} 300.00}} &
  {\color[HTML]{000000} 0.00} &
  \multicolumn{1}{r|}{{\color[HTML]{000000} 300.00}} &
  {\color[HTML]{000000} 0.00} &
  \multicolumn{1}{r|}{{\color[HTML]{000000} 300.00}} &
  {\color[HTML]{000000} 0.00} &
  \multicolumn{1}{c|}{300.00} &
  0.00 \\ \hline
\multicolumn{1}{|c|}{{\color[HTML]{000000} }} &
  {\color[HTML]{000000} 2} &
  \multicolumn{1}{r|}{{\color[HTML]{009901} \textbf{55.99}}} &
  {\color[HTML]{000000} 29.53} &
  \multicolumn{1}{r|}{{\color[HTML]{000000} 134.53}} &
  {\color[HTML]{000000} 54.56} &
  \multicolumn{1}{r|}{{\color[HTML]{F8A102} 64.17}} &
  {\color[HTML]{000000} 40.07} &
  \multicolumn{1}{r|}{{\color[HTML]{000000} 114.45}} &
  {\color[HTML]{000000} 27.64} &
  \multicolumn{1}{c|}{80.16} &
  10.67 \\ \cmidrule{2-12} 
\multicolumn{1}{|c|}{{\color[HTML]{000000} }} &
  {\color[HTML]{000000} 5} &
  \multicolumn{1}{r|}{{\color[HTML]{009901} \textbf{177.66}}} &
  {\color[HTML]{000000} 28.30} &
  \multicolumn{1}{r|}{{\color[HTML]{000000} 295.40}} &
  {\color[HTML]{000000} 6.50} &
  \multicolumn{1}{r|}{{\color[HTML]{F8A102} 261.81}} &
  {\color[HTML]{000000} 54.22} &
  \multicolumn{1}{r|}{{\color[HTML]{000000} 280.70}} &
  {\color[HTML]{000000} 21.80} &
  \multicolumn{1}{c|}{300.00} &
  0.00 \\ \cmidrule{2-12} 
\multicolumn{1}{|c|}{\multirow{-3}{*}{{\color[HTML]{000000} 8}}} &
  {\color[HTML]{000000} 10} &
  \multicolumn{1}{r|}{{\color[HTML]{009901} \textbf{253.13}}} &
  {\color[HTML]{000000} 43.96} &
  \multicolumn{1}{r|}{{\color[HTML]{000000} 300.00}} &
  {\color[HTML]{000000} 0.00} &
  \multicolumn{1}{r|}{{\color[HTML]{000000} 300.00}} &
  {\color[HTML]{000000} 0.00} &
  \multicolumn{1}{r|}{{\color[HTML]{000000} 300.00}} &
  {\color[HTML]{000000} 0.00} &
  \multicolumn{1}{c|}{300.00} &
  0.00 \\ \hline
\multicolumn{2}{|c|}{{\color[HTML]{000000} average}} &
  \multicolumn{1}{r|}{{\color[HTML]{000000} \textbf{176.23}}} &
  {\color[HTML]{000000} 31.16} &
  \multicolumn{1}{r|}{{\color[HTML]{000000} 253.22}} &
  {\color[HTML]{000000} 17.16} &
  \multicolumn{1}{r|}{{\color[HTML]{000000} 256.86}} &
  {\color[HTML]{000000} 27.62} &
  \multicolumn{1}{r|}{{\color[HTML]{000000} 238.28}} &
  {\color[HTML]{000000} 11.09} &
  \multicolumn{1}{c|}{275.57} &
  1.19 \\ \hline
\end{tabular}%
}
\vspace{-0.1cm}
\caption{City2}
\end{subtable}
\caption{Mission completion time (sec), with a cut-off time at $\SI{300}{\second}$---the lower the better \revisedtext{across each row}. If a method has a reported mean of $\SI{300}{\second}$, that means that some targets were not found. Best performance in {\best green}, second best in {\bestsecond orange}. Our method has better completion time than other baseline methods. %
}

\label{tab:mission}
\end{table*}

\begin{figure}[t!]
    \centering
    \includegraphics[width=\columnwidth]{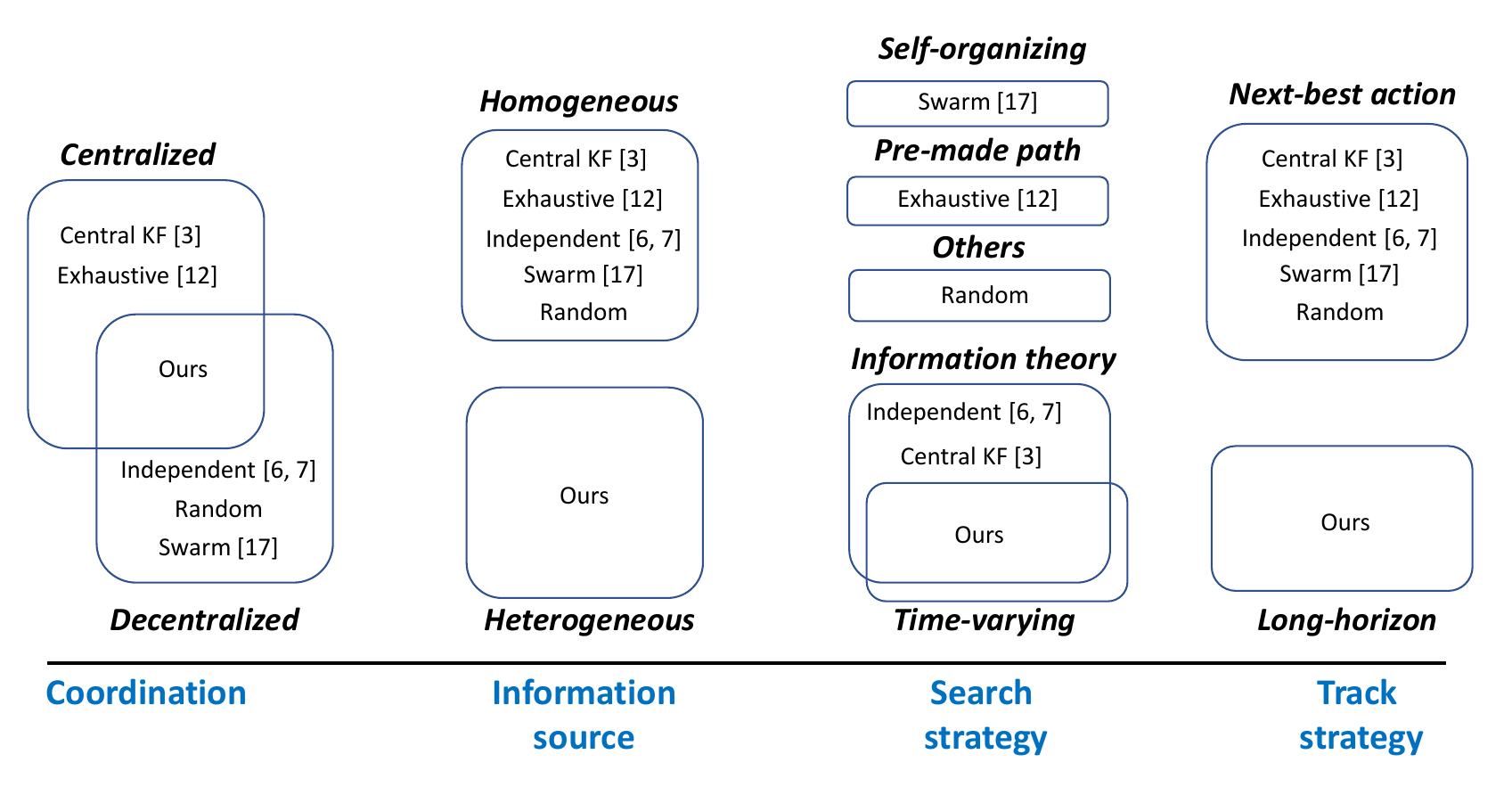}
    \caption{\revisedtext{Our method and tested baselines as representative of classes of approaches, which are marked with the corresponding references. Note that we chose \emph{Random} baseline as the commonly used one in the literature, e.g., \cite{cell-mb-2023, gas-mapping-2019}}.}
    \label{fig:method-dim}
\end{figure}

\begin{figure}[t!]
    \centering
    \begin{subfigure}{\columnwidth}
        \centering
        \includegraphics[width=\columnwidth]{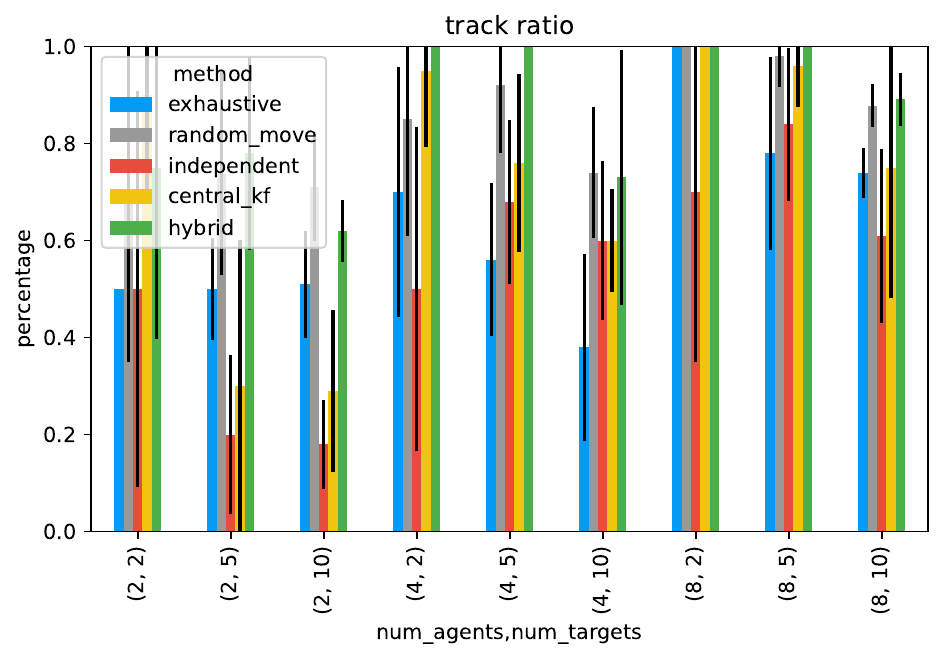} %
        \label{fig:sub1}
    \end{subfigure}
    \begin{subfigure}{\columnwidth}
        \centering
        \includegraphics[width=\columnwidth]{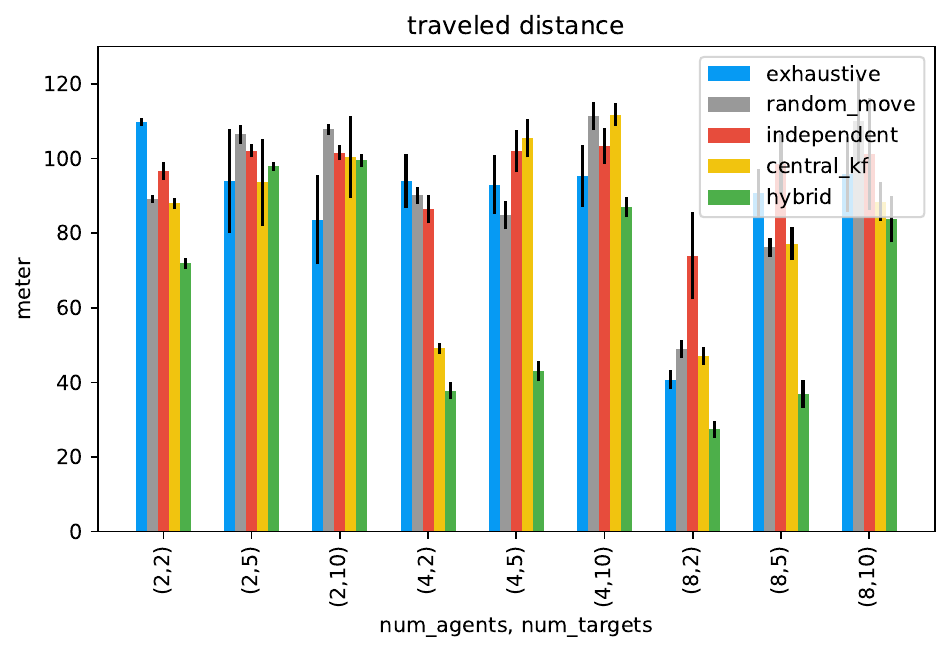} %
        \label{fig:sub3}
\end{subfigure}
\caption{\revisedtext{\emph{City1} example's tracking ratio (\textit{top}---higher is better) and traveled distance (\textit{bottom}---lower is better) performance per (agent, target) pair experiments. Our proposed method (hybrid) is shown in green. Overall, our proposed method shows higher track ratio, lower time taken for tracking, and lower traveled distance, compared to the baselines.}}
\label{fig:track-time-distance}
\end{figure}

\subsection{Comparative analysis}
To the best of our knowledge, there is no open-source benchmark for search and track  \cite{sat-swarm-review-2016}. Hence, we implemented the following baselines from the literature and conducted comparative analysis. \revisedtext{From the literature review \tab{literature-review}, we chose the methods to cover diverse aspects per category (see \fig{method-dim}). These baselines are commonly used from other recent search and track-related prior works such as \cite{cell-mb-2023, gas-mapping-2019, distributed-control-silvia-2018, upenn-balance-2015}}. The implementation of the baselines as well as our proposed method will be released opensource and is included in the supplementary material: (1) \emph{Random walk} where a random location on the current frontier is selected and followed during exploration, and during tracking the agent gets to the last known position of the detected object; 
(2) \emph{Independent}, has each agent behaving independently while maximizing its own information gain without any coordination between HQ and an agent during exploration and during tracking, the agents follows the predicted location by Kalman Filter \revisedtext{(KF)} \cite{increasing-autonomy-csat-2009, recursive-bayesian-csat-2006}. 
(3) \emph{Central KF}, uses an active searching strategy based on team-level information gain \cite{upenn-balance-2015} \revisedtext{-- without time-varying component and balancing $\jexplore$ and $\jexploit$, i.e., relying on only $\jexplore$ during search --} while predicting targets based on myopic view by Kalman filter;
\revisedtext{(4) \emph{Exhaustive}, decomposes the known environment into Voronoi partitioning \cite{vornoi-1991, vornoi-mcpp-2018}. Employing the strategy from \cite{malika-search-2016}, the agents conduct search based on the multi-robot coverage paths built by the Boustrophedon algorithm \cite{Mier_Fields2Cover_An_open-source_2023} and follow the predicted location by KF}; and
\revisedtext{(5) \emph{Swarm intelligence},  utilizes a self-organizing cooperative pursuit strategy by controlling the formation of swarms and determining the next-best action with a vector force during the search and tracking of dynamic objects, employing strategies from \cite{usv-pursuit-evasion-2023}. Note that we tested all the baselines with our proposed approach, except for \emph{Swarm intelligence}, which was only tested in an \emph{open} environment due to the topology and confined spaces between obstacles in real environments.}

\begin{figure}
    \centering
    \begin{minipage}[t]{0.45\columnwidth}
        \begin{subfigure}[b]{\textwidth}
            \includegraphics[width=\textwidth]{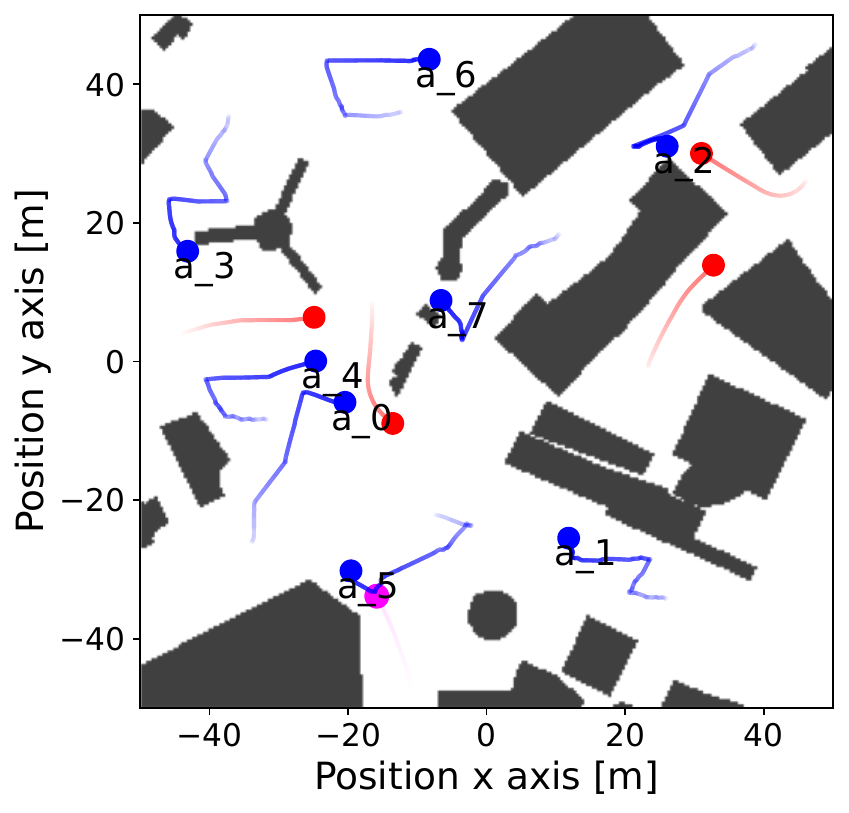}
            \vspace{-0.6cm}
            \caption{\emph{\best Ours (\emph{hybrid})} -- \SI{100}{\second}}
            \label{fig:hybrid-100}
        \end{subfigure}
    \end{minipage}
    \begin{minipage}[t]{0.45\columnwidth}
        \begin{subfigure}[b]{\textwidth}
            \includegraphics[width=\textwidth]{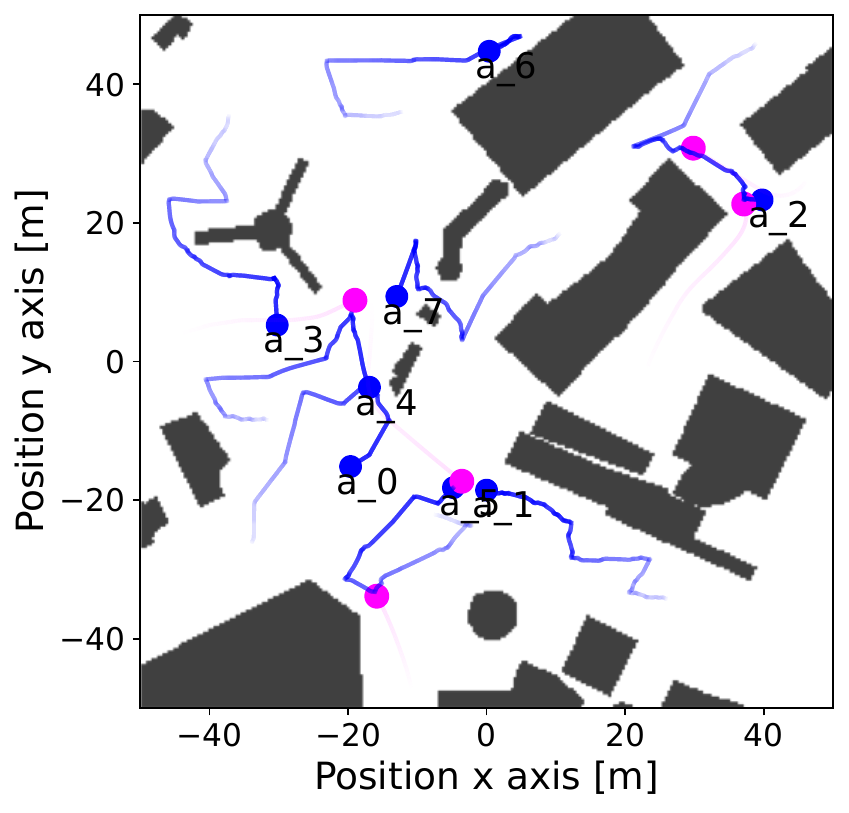}
            \vspace{-0.6cm}
            \caption{\emph{\best Ours (\emph{hybrid})} -- \SI{200}{\second}}
            \label{fig:hybrid-200}
        \end{subfigure}
    \end{minipage}
    \begin{minipage}[t]{0.45\columnwidth}
        \begin{subfigure}[b]{\textwidth}            
            \includegraphics[width=\textwidth]{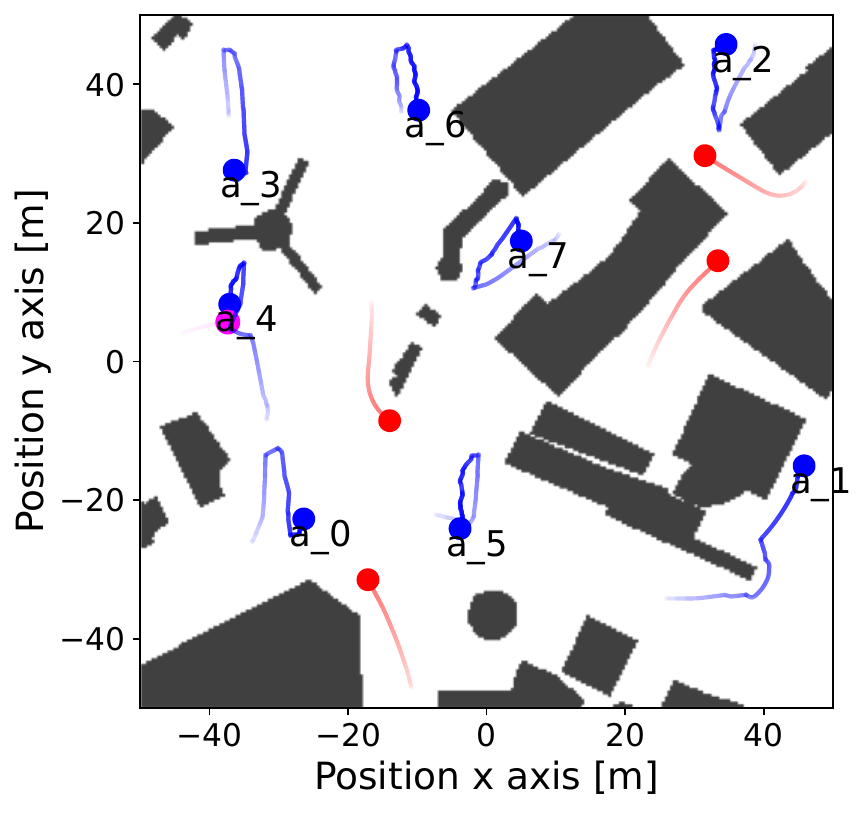}
            \vspace{-0.6cm}
            \caption{\emph{exhaustive} -- \SI{100}{\second}}
            \label{fig:exhaustive-100}
        \end{subfigure}
    \end{minipage}
    \begin{minipage}[t]{0.45\columnwidth}
        \begin{subfigure}[b]{\textwidth}
            \includegraphics[width=\textwidth]{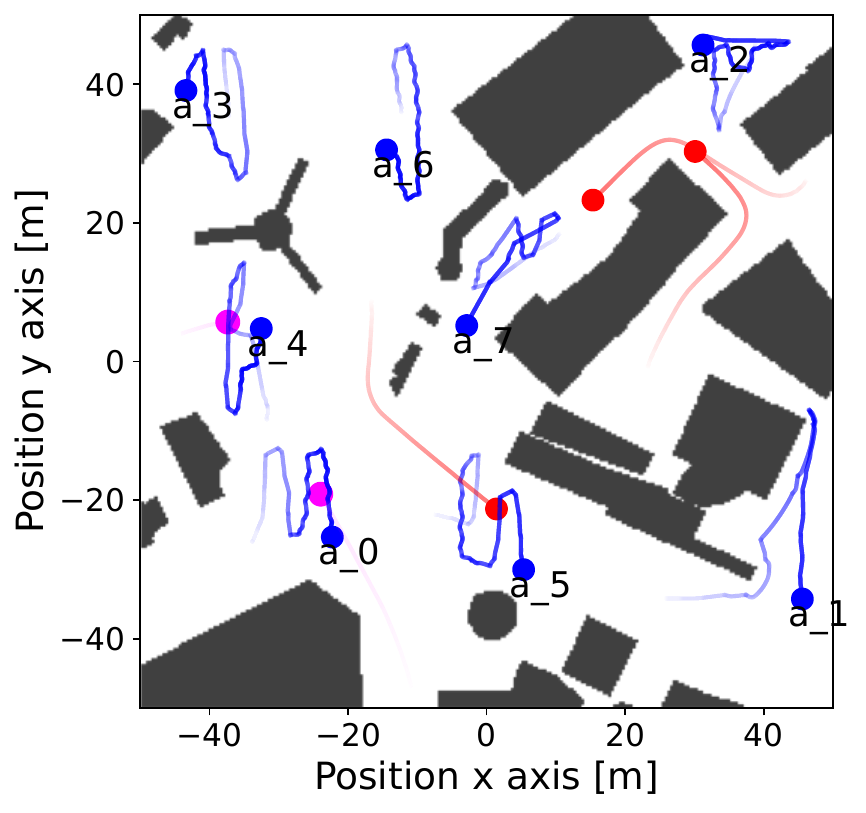}
            \vspace{-0.6cm}
            \caption{\emph{exhaustive} -- \SI{200}{\second}}
            \label{fig:exhaustive-200}
        \end{subfigure}
    \end{minipage}
    \newline
    \begin{minipage}[t]{0.45\columnwidth}
        \begin{subfigure}[b]{\textwidth}            
            \includegraphics[width=\textwidth]{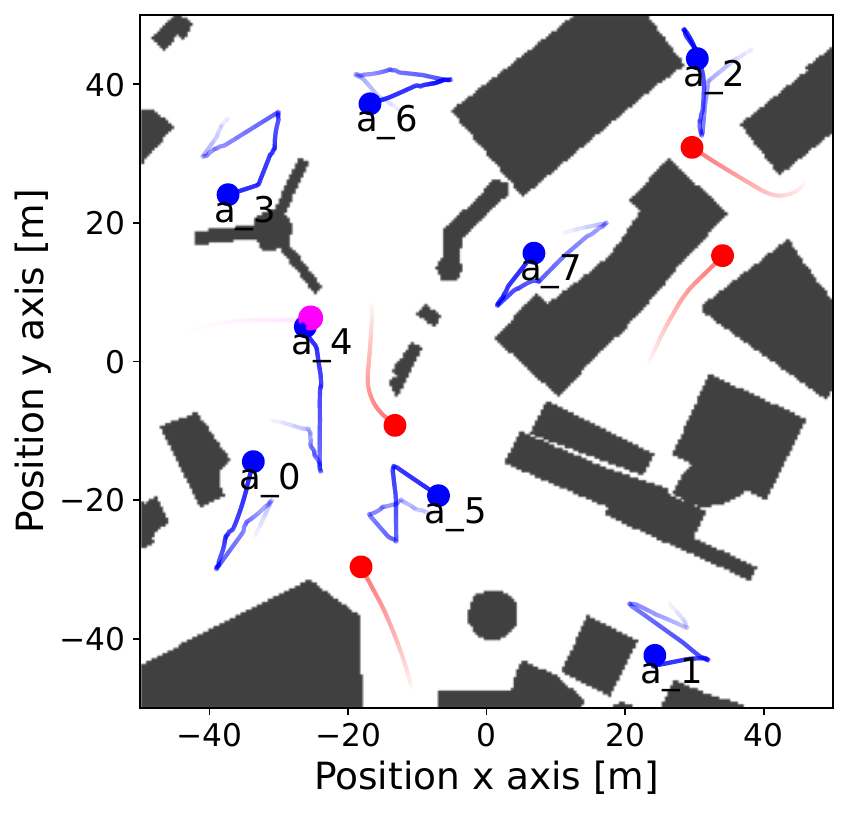}
            \vspace{-0.6cm}
            \caption{\emph{random walk} -- \SI{100}{\second}}
            \label{fig:random-100}
        \end{subfigure}
    \end{minipage}
    \begin{minipage}[t]{0.45\columnwidth}
        \begin{subfigure}[b]{\textwidth}
            \includegraphics[width=\textwidth]{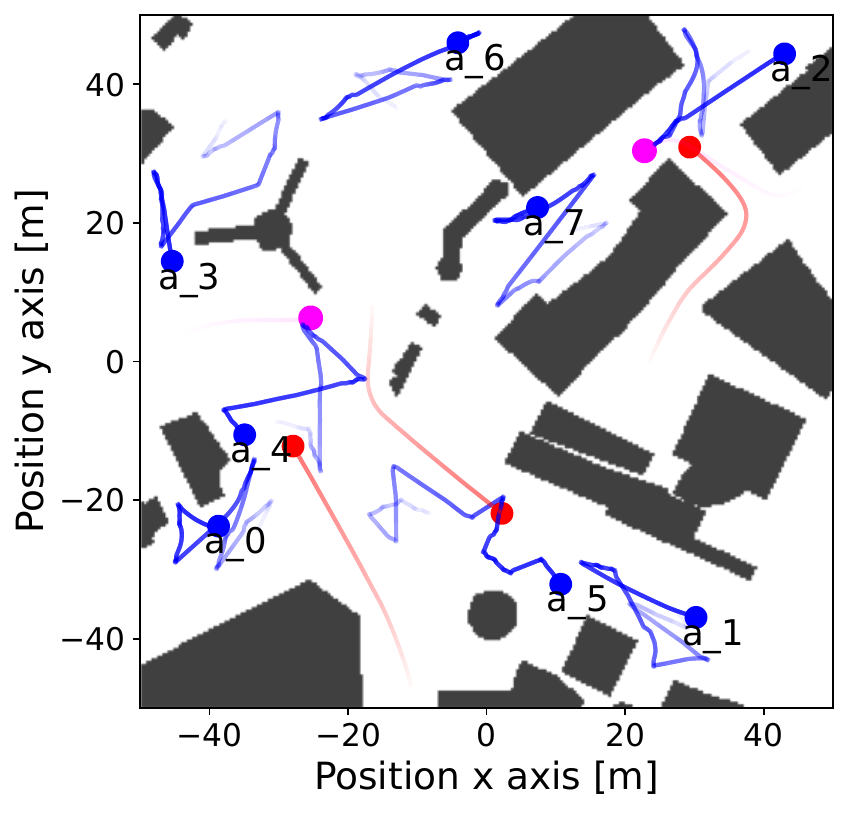}
            \vspace{-0.6cm}
            \caption{\emph{random walk} -- \SI{200}{\second}}
            \label{fig:random-200}
        \end{subfigure}
    \end{minipage}
    \newline
    \begin{minipage}[t]{0.45\columnwidth}
        \begin{subfigure}[b]{\textwidth}            
            \includegraphics[width=\textwidth]{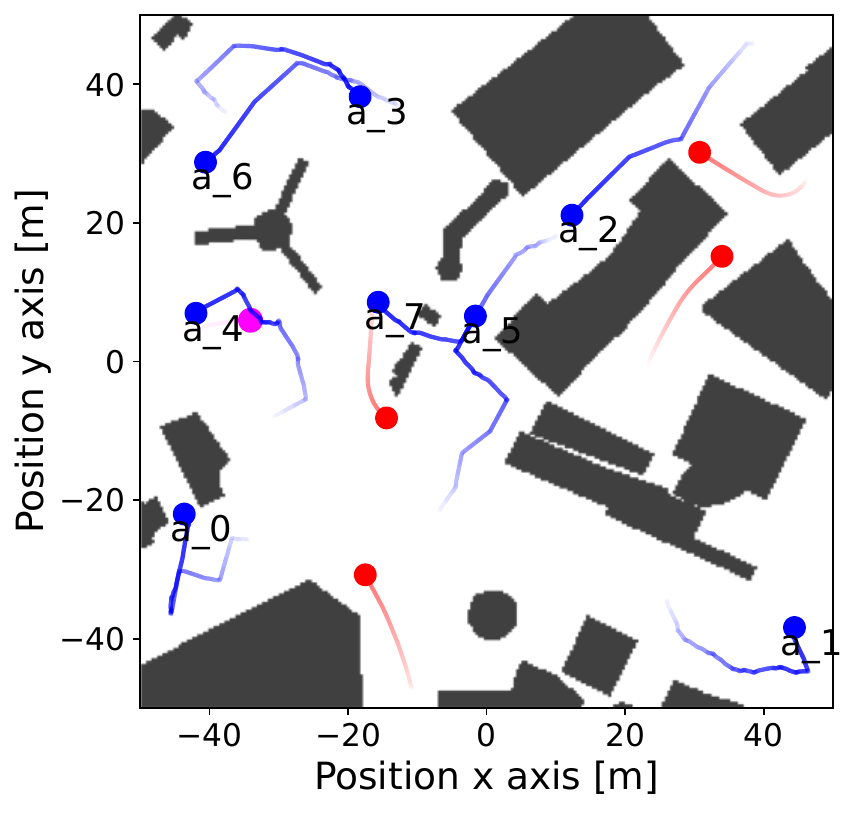}
            \vspace{-0.6cm}
            \caption{\emph{independent} -- \SI{100}{\second}}
            \label{fig:independent-100}
        \end{subfigure}
    \end{minipage}
    \begin{minipage}[t]{0.45\columnwidth}
        \begin{subfigure}[b]{\textwidth}
            \includegraphics[width=\textwidth]{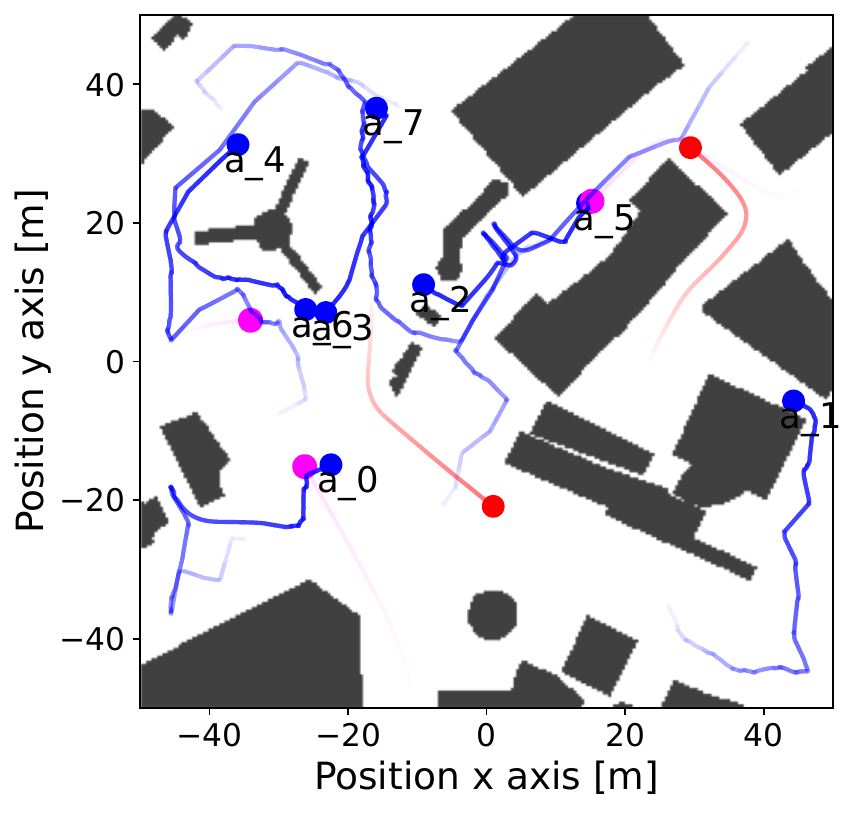}
            \vspace{-0.6cm}
            \caption{\emph{independent} -- \SI{200}{\second}}
            \label{fig:independent-200}
        \end{subfigure}
    \end{minipage}
    \newline
    \begin{minipage}[t]{0.45\columnwidth}
        \begin{subfigure}[b]{\textwidth}            
            \includegraphics[width=\textwidth]{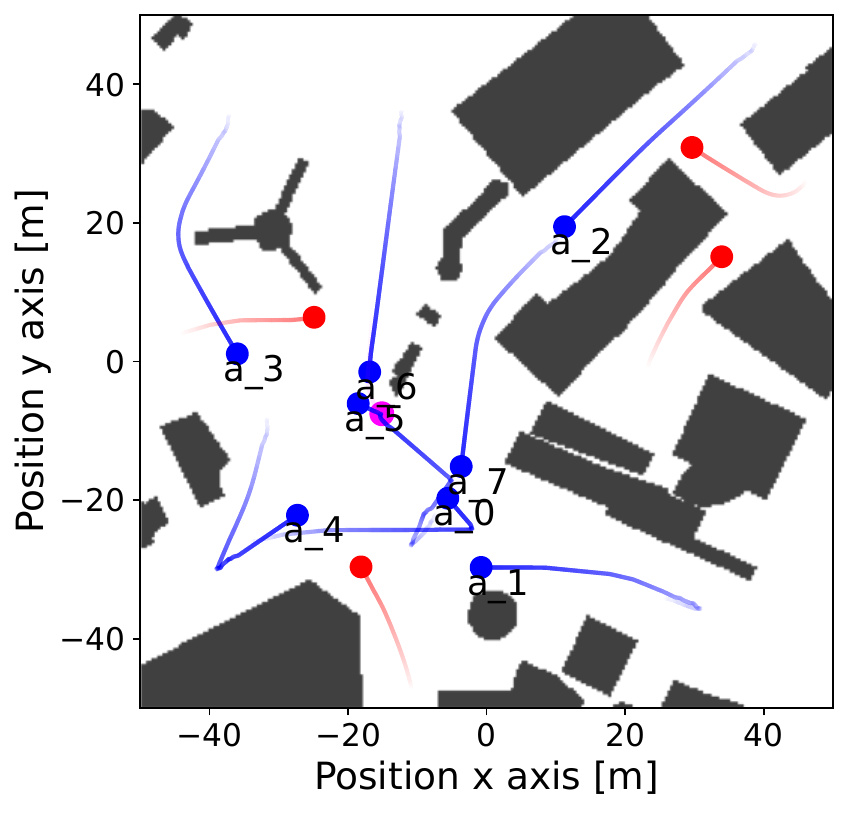}
            \vspace{-0.6cm}
            \caption{\emph{central KF} -- \SI{100}{\second}}
            \label{fig:central-100}
        \end{subfigure}
    \end{minipage}
    \begin{minipage}[t]{0.45\columnwidth}
        \begin{subfigure}[b]{\textwidth}
            \includegraphics[width=\textwidth]{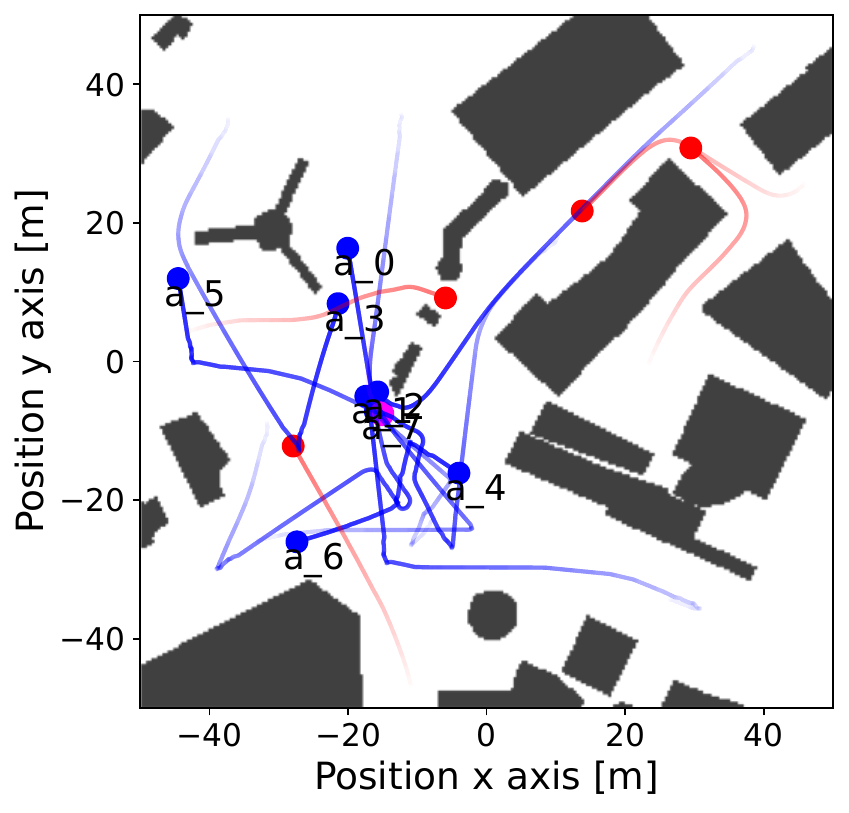}
            \vspace{-0.6cm}
            \caption{\emph{central KF} -- \SI{200}{\second}}
            \label{fig:central-200}
        \end{subfigure}
    \end{minipage}
    \newline

    \caption{\revisedtext{Qualitative results from trajectory comparisons in \emph{City2} example with agent, target (\num{8},\num{5}) configuration at $\SI{100}{\second}$ and $\SI{200}{\second}$. Trajectories of agents and targets are represented in blue and red, respectively. Targets after the completion of the search and track (which no longer move) are colored in magenta. Our proposed hybrid method finds the targets and reduces the location uncertainty with the best performance---no targets are left at $\SI{200}{\second}$ compared to the baselines--by utilizing both exploration and exploitation during the search and track process.}}
    \label{fig:qualitative-compare}
\end{figure}

Running all experimental configurations with repetitions resulted in \num{810} runs per each method \revisedtext{(\num{270} runs for \emph{swarm intelligence}}), totaling \revisedtext{\num{4320}} runs  for Monte Carlo simulations. We used the completion threshold of tracking $\trace(\Sigma_{\mi{thre}}) = 2.0$ given the size of the robots. Considering a \SI{300}{\second} mission time limit, the simulations ran approximately \revisedtext{\num{360}} hours.  

\tab{mission} shows that our proposed method achieves better mission completion time than all the baselines. 
The environments \emph{city1, city2} are more complex than \emph{open}, resulting in some cases where the agents couldn't track all objects within the 5-minute time limit, especially when there are not sufficiently many agents for the task. There were a few cases there was no statistically significant difference between the performance of our proposed method and the others, e.g., agent, target (\num{2}, \num{10}) in \emph{open} with $p\textrm{-value}$ \num{0.33} given the challenging scenario caused by the small number of agents compared to the large number of target objects. Most of the  results showed  statistically significant difference, $p\textrm{-value} < 0.01$ using $t$-test, across all the baselines, e.g., agent, target (\num{4}, \num{5}) in \emph{city1}, $p\textrm{-value} = 0.0002$. 
Looking at the track ratio and traveled distance (\fig{track-time-distance}), our proposed method was able to track objects efficiently, without as many lost detection as other methods and with the agent exploring the environment.  
The entire comparative analysis results are included in the supplementary material.

\revisedtext{
\fig{qualitative-compare} presents the qualitative results from trajectory comparisons across each method in the \emph{City2} scenario with an agent-target configuration of (\num{8},\num{5}). Our proposed method (\fig{hybrid-100}, \fig{hybrid-200}) effectively finds targets by leveraging information gained from  exploration and third-party reporting. Notably, at \SI{100}{\second}, agents $\ag_0$ and $\ag_2$ are biased towards the area where the targets are located and successfully capture them. At \SI{200}{\second}, agents $\ag_1$ and $\ag_5$ cooperatively track the target (position near $[0,-20]$) and complete the mission, demonstrating the robustness of our allocation method outlined in \alg{alg:track-task-assignment}, different from conventional simple auction-based methods. In contrast, as shown in \fig{exhaustive-100} to \fig{random-200}, both the \emph{Exhaustive} and \emph{Random Walk} methods exhibit unnecessarily repeated patterns and fail to effectively find and track targets due to: (1) a focus on multi-robot coverage of the environment (\emph{Exhaustive}), and (2) random choices from the frontier that overlook biased decisions towards target-related information (\emph{Random Walk}). The \emph{Independent} method (\fig{independent-100}, \fig{independent-200}), which lacks coordination with headquarters, results in unnecessary overlaps of trajectories (e.g., agents $\ag_4$ and $\ag_6$). Meanwhile, the \emph{Central KF} method (\fig{central-100}, \fig{central-200}), lacking an exploitation component and a time-varying map during the search despite having coordination with HQ, leads agents to concentrate in areas ideal for exploration from the shared frontier (central part of the map with many openings at the beginning). However, these agents fail to find and track many targets, with \num{4} still unlocated at \SI{200}{\second}.
}

\subsection{Ablation Studies}
\begin{table*}
\tablefontsize
\begin{subtable}{\linewidth}
\centering
\resizebox{.75\linewidth}{!}{%

\begin{tabular}{|cc|cc|rr|rr|rr|rr|}
\hline
\multicolumn{1}{|c|}{} &
   &
  \multicolumn{2}{c|}{\textbf{Ours}} &
  \multicolumn{2}{c|}{\textbf{TR, LSTM}} &
  \multicolumn{2}{c|}{\textbf{LSTM}} &
  \multicolumn{2}{c|}{\textbf{TV, LSTM}} &
  \multicolumn{2}{c|}{\textbf{TR,TV}} \\ \cmidrule{3-12} 
\multicolumn{1}{|c|}{\multirow{-2}{*}{\textbf{Agent num}}} &
  \multirow{-2}{*}{\textbf{Target num}} &
  \multicolumn{1}{c|}{mean} &
  std &
  \multicolumn{1}{c|}{mean} &
  \multicolumn{1}{c|}{std} &
  \multicolumn{1}{c|}{mean} &
  \multicolumn{1}{c|}{std} &
  \multicolumn{1}{c|}{mean} &
  \multicolumn{1}{c|}{std} &
  \multicolumn{1}{c|}{mean} &
  \multicolumn{1}{c|}{std} \\ \hline
\multicolumn{1}{|c|}{} &
  2 &
  \multicolumn{1}{c|}{{\color[HTML]{009901} \textbf{134.00}}} &
  38.13 &
  \multicolumn{1}{r|}{{\color[HTML]{F8A102} 196.35}} &
  51.70 &
  \multicolumn{1}{r|}{265.41} &
  50.50 &
  \multicolumn{1}{r|}{263.89} &
  58.89 &
  \multicolumn{1}{r|}{212.16} &
  59.43 \\ \cmidrule{2-12} 
\multicolumn{1}{|c|}{} &
  5 &
  \multicolumn{1}{c|}{{\color[HTML]{009901} \textbf{260.68}}} &
  43.62 &
  \multicolumn{1}{r|}{{\color[HTML]{F8A102} 265.73}} &
  45.07 &
  \multicolumn{1}{r|}{300.00} &
  0.00 &
  \multicolumn{1}{r|}{293.19} &
  18.14 &
  \multicolumn{1}{r|}{281.52} &
  38.36 \\ \cmidrule{2-12} 
\multicolumn{1}{|c|}{\multirow{-3}{*}{2}} &
  10 &
  \multicolumn{1}{c|}{{\color[HTML]{F8A102} 294.96}} &
  15.94 &
  \multicolumn{1}{r|}{300.00} &
  0.00 &
  \multicolumn{1}{r|}{300.00} &
  0.00 &
  \multicolumn{1}{r|}{300.00} &
  0.00 &
  \multicolumn{1}{r|}{{\color[HTML]{009901} \textbf{294.27}}} &
  13.31 \\ \hline
\multicolumn{1}{|c|}{} &
  2 &
  \multicolumn{1}{c|}{{\color[HTML]{009901} \textbf{84.52}}} &
  12.45 &
  \multicolumn{1}{r|}{{\color[HTML]{F8A102} 87.49}} &
  32.89 &
  \multicolumn{1}{r|}{233.48} &
  70.87 &
  \multicolumn{1}{r|}{180.15} &
  82.16 &
  \multicolumn{1}{r|}{{\color[HTML]{000000} 92.27}} &
  39.81 \\ \cmidrule{2-12} 
\multicolumn{1}{|c|}{} &
  5 &
  \multicolumn{1}{c|}{{\color[HTML]{F8A102} 127.98}} &
  36.55 &
  \multicolumn{1}{r|}{183.40} &
  87.40 &
  \multicolumn{1}{r|}{234.39} &
  105.07 &
  \multicolumn{1}{r|}{186.44} &
  79.90 &
  \multicolumn{1}{r|}{{\color[HTML]{009901} \textbf{127.88}}} &
  68.27 \\ \cmidrule{2-12} 
\multicolumn{1}{|c|}{\multirow{-3}{*}{4}} &
  10 &
  \multicolumn{1}{c|}{{\color[HTML]{F8A102} 232.84}} &
  51.10 &
  \multicolumn{1}{r|}{{\color[HTML]{000000} 237.32}} &
  70.40 &
  \multicolumn{1}{r|}{299.49} &
  1.96 &
  \multicolumn{1}{r|}{282.65} &
  30.90 &
  \multicolumn{1}{r|}{{\color[HTML]{009901} \textbf{198.03}}} &
  64.81 \\ \hline
\multicolumn{1}{|c|}{} &
  2 &
  \multicolumn{1}{c|}{{\color[HTML]{009901} \textbf{59.96}}} &
  9.86 &
  \multicolumn{1}{r|}{{\color[HTML]{F8A102} 81.01}} &
  28.67 &
  \multicolumn{1}{r|}{85.13} &
  44.58 &
  \multicolumn{1}{r|}{84.27} &
  40.50 &
  \multicolumn{1}{r|}{86.03} &
  33.09 \\ \cmidrule{2-12} 
\multicolumn{1}{|c|}{} &
  5 &
  \multicolumn{1}{c|}{{\color[HTML]{009901} \textbf{76.90}}} &
  8.68 &
  \multicolumn{1}{r|}{{\color[HTML]{F8A102} 79.01}} &
  28.40 &
  \multicolumn{1}{r|}{103.47} &
  47.18 &
  \multicolumn{1}{r|}{108.52} &
  76.74 &
  \multicolumn{1}{r|}{113.45} &
  43.82 \\ \cmidrule{2-12} 
\multicolumn{1}{|c|}{\multirow{-3}{*}{8}} &
  10 &
  \multicolumn{1}{c|}{{\color[HTML]{009901} \textbf{113.58}}} &
  31.62 &
  \multicolumn{1}{r|}{{\color[HTML]{F8A102} 156.77}} &
  92.01 &
  \multicolumn{1}{r|}{{\color[HTML]{000000} 164.81}} &
  74.63 &
  \multicolumn{1}{r|}{174.04} &
  66.67 &
  \multicolumn{1}{r|}{173.28} &
  86.10 \\ \hline
\multicolumn{2}{|c|}{average} &
  \multicolumn{1}{c|}{\textbf{153.94}} &
  27.55 &
  \multicolumn{1}{r|}{176.34} &
  48.51 &
  \multicolumn{1}{r|}{220.69} &
  43.86 &
  \multicolumn{1}{r|}{208.13} &
  50.43 &
  \multicolumn{1}{r|}{175.43} &
  49.67 \\ \hline
\end{tabular}%
}
\caption{Open}
\end{subtable}

\begin{subtable}{\linewidth}
\centering
\resizebox{.75\linewidth}{!}{%
\begin{tabular}{|cc|rr|rr|rr|rr|rr|}
\hline
\multicolumn{1}{|c|}{} &
   &
  \multicolumn{2}{c|}{\textbf{Ours}} &
  \multicolumn{2}{c|}{\textbf{TR, LSTM}} &
  \multicolumn{2}{c|}{\textbf{LSTM}} &
  \multicolumn{2}{c|}{\textbf{TV, LSTM}} &
  \multicolumn{2}{c|}{\textbf{TR,TV}} \\ \cmidrule{3-12} 
\multicolumn{1}{|c|}{\multirow{-2}{*}{Agent num}} &
  \multirow{-2}{*}{Target num} &
  \multicolumn{1}{c|}{mean} &
  \multicolumn{1}{c|}{std} &
  \multicolumn{1}{c|}{mean} &
  \multicolumn{1}{c|}{std} &
  \multicolumn{1}{c|}{mean} &
  \multicolumn{1}{c|}{std} &
  \multicolumn{1}{c|}{mean} &
  \multicolumn{1}{c|}{std} &
  \multicolumn{1}{c|}{mean} &
  \multicolumn{1}{c|}{std} \\ \hline
\multicolumn{1}{|c|}{} &
  2 &
  \multicolumn{1}{r|}{{\color[HTML]{009901} \textbf{208.62}}} &
  {\color[HTML]{000000} 81.88} &
  \multicolumn{1}{r|}{{\color[HTML]{F8A102} 260.07}} &
  {\color[HTML]{000000} 59.59} &
  \multicolumn{1}{r|}{{\color[HTML]{000000} 263.17}} &
  {\color[HTML]{000000} 60.78} &
  \multicolumn{1}{r|}{{\color[HTML]{000000} 284.01}} &
  {\color[HTML]{000000} 28.85} &
  \multicolumn{1}{r|}{{\color[HTML]{000000} 261.60}} &
  {\color[HTML]{000000} 54.12} \\ \cmidrule{2-12} 
\multicolumn{1}{|c|}{} &
  5 &
  \multicolumn{1}{r|}{{\color[HTML]{009901} \textbf{286.26}}} &
  {\color[HTML]{000000} 26.99} &
  \multicolumn{1}{r|}{{\color[HTML]{000000} 300.00}} &
  {\color[HTML]{000000} 0.00} &
  \multicolumn{1}{r|}{{\color[HTML]{000000} 300.00}} &
  {\color[HTML]{000000} 0.00} &
  \multicolumn{1}{r|}{{\color[HTML]{000000} 300.00}} &
  {\color[HTML]{000000} 0.00} &
  \multicolumn{1}{r|}{{\color[HTML]{000000} 300.00}} &
  {\color[HTML]{000000} 0.00} \\ \cmidrule{2-12} 
\multicolumn{1}{|c|}{\multirow{-3}{*}{2}} &
  10 &
  \multicolumn{1}{r|}{{\color[HTML]{000000} 300.00}} &
  {\color[HTML]{000000} 0.00} &
  \multicolumn{1}{r|}{{\color[HTML]{000000} 300.00}} &
  {\color[HTML]{000000} 0.00} &
  \multicolumn{1}{r|}{{\color[HTML]{000000} 300.00}} &
  {\color[HTML]{000000} 0.00} &
  \multicolumn{1}{r|}{{\color[HTML]{000000} 300.00}} &
  {\color[HTML]{000000} 0.00} &
  \multicolumn{1}{r|}{{\color[HTML]{000000} 300.00}} &
  {\color[HTML]{000000} 0.00} \\ \hline
\multicolumn{1}{|c|}{} &
  2 &
  \multicolumn{1}{r|}{{\color[HTML]{009901} \textbf{118.74}}} &
  {\color[HTML]{000000} 19.79} &
  \multicolumn{1}{r|}{{\color[HTML]{000000} 238.12}} &
  {\color[HTML]{000000} 72.87} &
  \multicolumn{1}{r|}{{\color[HTML]{000000} 263.31}} &
  {\color[HTML]{000000} 63.90} &
  \multicolumn{1}{r|}{{\color[HTML]{000000} 246.19}} &
  {\color[HTML]{000000} 86.03} &
  \multicolumn{1}{r|}{{\color[HTML]{F8A102} 222.29}} &
  {\color[HTML]{000000} 72.59} \\ \cmidrule{2-12} 
\multicolumn{1}{|c|}{} &
  5 &
  \multicolumn{1}{r|}{{\color[HTML]{009901} \textbf{137.20}}} &
  {\color[HTML]{000000} 16.98} &
  \multicolumn{1}{r|}{{\color[HTML]{000000} 266.87}} &
  {\color[HTML]{000000} 51.06} &
  \multicolumn{1}{r|}{{\color[HTML]{000000} 289.71}} &
  {\color[HTML]{000000} 39.87} &
  \multicolumn{1}{r|}{{\color[HTML]{000000} 295.76}} &
  {\color[HTML]{000000} 16.42} &
  \multicolumn{1}{r|}{{\color[HTML]{F8A102} 243.31}} &
  {\color[HTML]{000000} 68.19} \\ \cmidrule{2-12} 
\multicolumn{1}{|c|}{\multirow{-3}{*}{4}} &
  10 &
  \multicolumn{1}{r|}{{\color[HTML]{000000} 300.00}} &
  {\color[HTML]{000000} 0.00} &
  \multicolumn{1}{r|}{{\color[HTML]{000000} 300.00}} &
  {\color[HTML]{000000} 0.00} &
  \multicolumn{1}{r|}{{\color[HTML]{000000} 300.00}} &
  {\color[HTML]{000000} 0.00} &
  \multicolumn{1}{r|}{{\color[HTML]{000000} 300.00}} &
  {\color[HTML]{000000} 0.00} &
  \multicolumn{1}{r|}{{\color[HTML]{000000} 300.00}} &
  {\color[HTML]{000000} 0.00} \\ \hline
\multicolumn{1}{|c|}{} &
  2 &
  \multicolumn{1}{r|}{{\color[HTML]{009901} \textbf{102.06}}} &
  {\color[HTML]{000000} 19.13} &
  \multicolumn{1}{r|}{{\color[HTML]{000000} 151.48}} &
  {\color[HTML]{000000} 56.79} &
  \multicolumn{1}{r|}{{\color[HTML]{000000} 244.51}} &
  {\color[HTML]{000000} 70.29} &
  \multicolumn{1}{r|}{{\color[HTML]{000000} 206.32}} &
  {\color[HTML]{000000} 78.89} &
  \multicolumn{1}{r|}{{\color[HTML]{F8A102} 120.61}} &
  {\color[HTML]{000000} 29.25} \\ \cmidrule{2-12} 
\multicolumn{1}{|c|}{} &
  5 &
  \multicolumn{1}{r|}{{\color[HTML]{009901} \textbf{136.29}}} &
  {\color[HTML]{000000} 25.51} &
  \multicolumn{1}{r|}{{\color[HTML]{000000} 225.95}} &
  {\color[HTML]{000000} 66.79} &
  \multicolumn{1}{r|}{{\color[HTML]{000000} 281.71}} &
  {\color[HTML]{000000} 40.05} &
  \multicolumn{1}{r|}{{\color[HTML]{000000} 268.23}} &
  {\color[HTML]{000000} 55.12} &
  \multicolumn{1}{r|}{{\color[HTML]{F8A102} 189.76}} &
  {\color[HTML]{000000} 77.10} \\ \cmidrule{2-12} 
\multicolumn{1}{|c|}{\multirow{-3}{*}{8}} &
  10 &
  \multicolumn{1}{r|}{{\color[HTML]{333333} 299.27}} &
  {\color[HTML]{000000} 2.41} &
  \multicolumn{1}{r|}{{\color[HTML]{000000} 299.91}} &
  {\color[HTML]{000000} 0.36} &
  \multicolumn{1}{r|}{{\color[HTML]{000000} 300.00}} &
  {\color[HTML]{000000} 0.00} &
  \multicolumn{1}{r|}{{\color[HTML]{F8A102} 295.15}} &
  {\color[HTML]{000000} 12.68} &
  \multicolumn{1}{r|}{{\color[HTML]{009901} \textbf{294.81}}} &
  {\color[HTML]{000000} 10.52} \\ \hline
\multicolumn{2}{|c|}{average} &
  \multicolumn{1}{r|}{{\color[HTML]{000000} \textbf{209.83}}} &
  {\color[HTML]{000000} 21.41} &
  \multicolumn{1}{r|}{{\color[HTML]{000000} 260.27}} &
  {\color[HTML]{000000} 34.16} &
  \multicolumn{1}{r|}{{\color[HTML]{000000} 282.49}} &
  {\color[HTML]{000000} 30.54} &
  \multicolumn{1}{r|}{{\color[HTML]{000000} 277.29}} &
  {\color[HTML]{000000} 30.89} &
  \multicolumn{1}{r|}{{\color[HTML]{000000} 248.04}} &
  {\color[HTML]{000000} 34.64} \\ \hline
\end{tabular}%
}
  \caption{City1}
\end{subtable}
\caption{Ablation studies, mission time (sec), with a cut-off time at $\SI{300}{\second}$---the lower the better \revisedtext{across each row}. Best performance in {\best green}, second best in {\bestsecond orange}. \emph{City2} generally has a similar trend as \emph{City1} and is separately reported in the supplementary material.}

\label{tab:ablation}
\end{table*}

We conducted several ablations, totaling \num{1620} runs, which varied the number of agents, number of target, environments, and repetitions, on our method. This was done to systematically observe the impact of each component we introduced, namely time-varying uncertainty (TV), third-party reporting (TR), and LSTM---\tab{ablation}.

\emph{Impact of Third Party Reporting.}
\tab{ablation} shows that removing third-party reporting significantly increases the average mission completion time. This suggests that incorporating beliefs from heterogeneous sources during search can substantially enhance overall task performance. For example, except for a few cases (such as agent, target (\num{2}, \num{10}) in \emph{open} due to the small number of agents), $t\textrm{-test}$ between ours and (TV, LSTM) shows a significant difference with $p\textrm{-value} < 0.01$, e.g., agent, target (\num{4}, \num{5}) in \emph{city1}, $p\textrm{-value}=0.003$.

\emph{Impact of Time-Varying Uncertainty.} \tab{ablation} shows that removing time-varying uncertainty increases the average mission completion time. The inclusion of TV ensures that agents have the capability to revisit previously explored areas, which is crucial for effectively completing search and tracking missions of dynamic target objects. The majority of the cases showed a significant difference $p\textrm{-value} < 0.01$ by $t\textrm{-test}$ between ours and (TR, LSTM). For example, agent, target (\num{8}, \num{5}) in \emph{city1} showed $p\textrm{-value}= 0.0036$. There were a few cases where the impact of time-varying uncertainty is not directly observable, e.g., agent, target (\num{8}, \num{10}) in \emph{city1}  $p\textrm{-value}=0.38$, where the performance have been affected by a large number of target objects in a complex environment.

\emph{Impact of LSTM Use.}
Incorporating LSTM shows also improved performance on the overall task's completion time. Compared to TR and TV, the benefits of LSTM  is less prominent, e.g., agent, target (\num{4}, \num{10}) in \emph{empty} $p\textrm{-value}=0.6355$ by $t\textrm{-test}$ between ours and (TR, TV). This may be due  complicating factors in the search process and the intelligence level of coordination with HQ. These factors simultaneously affect the detection of target object(s) and subsequent tracking. 

Therefore, we conducted separate ablation studies to solely analyze the impact of the LSTM on the tracking task in a controlled setup.

\subsubsection{LSTM vs GP}
\revisedtext{We first compare LSTM with GP for the performance of long-horizon prediction. We used the test dataset in \sect{belief} for off-line testing, to analyze the applicability of the two methods. Based on the literature that uses GP for trajectory prediction \cite{gp-trajectory-nguyen-2024, gp-trajectory-IV-2018, vijay-sat-2017}, we tested different kernels of GP -- Mat\'ern, Radial Basis Function (RBF), Rational Quadratic (RQ), Exponential, and Linear -- for a covariance function of GP by using the same parameters for the best known performance, e.g., $\nu=1.5$ for Mat\'ern class \cite{vijay-sat-2017} where $\nu$ is for controlling the smoothness of the covariance function. We report the following two metrics: 
\begin{itemize}
    \item Average displacement error (ADE): mean square error between the predicted trajectories and the ground truth trajectories \cite{ade-Pellegrini_2009}.
    \item Final displacement error (FDE): $L2$ distance between the last position of the predicted trajectories and the ground truth trajectories \cite{social-lstm-2016}.
\end{itemize}
\emph{\tab{ablation-gp} shows that our LSTM better outperforms competing GP methods on both ADE and FDE.}
}

\begin{table}[]
\tablefontsize
\centering
\resizebox{\columnwidth}{!}{%
\begin{tabular}{|cc|c|ccccc|}
\hline
\multicolumn{2}{|c|}{} &
   &
  \multicolumn{5}{c|}{GP} \\ \cmidrule{4-8} 
\multicolumn{2}{|c|}{\multirow{-2}{*}{Metrics}} &
  \multirow{-2}{*}{\textbf{\begin{tabular}[c]{@{}c@{}}Ours \\ LSTM\end{tabular}}} &
  \multicolumn{1}{c|}{Mat\'ern} &
  \multicolumn{1}{c|}{RBF} &
  \multicolumn{1}{c|}{RQ} &
  \multicolumn{1}{c|}{Exponential} &
  Linear \\ \hline
\multicolumn{1}{|c|}{} &
  mean &
  {\color[HTML]{009901} \textbf{0.063}} &
  \multicolumn{1}{c|}{0.110} &
  \multicolumn{1}{c|}{0.383} &
  \multicolumn{1}{c|}{0.309} &
  \multicolumn{1}{c|}{0.403} &
  0.167 \\ \cmidrule{2-8} 
\multicolumn{1}{|c|}{\multirow{-2}{*}{ADE}} &
  std &
  0.035 &
  \multicolumn{1}{c|}{0.056} &
  \multicolumn{1}{c|}{0.064} &
  \multicolumn{1}{c|}{0.061} &
  \multicolumn{1}{c|}{0.065} &
  0.094 \\ \hline
\multicolumn{1}{|c|}{} &
  mean &
  {\color[HTML]{009901} \textbf{0.135}} &
  \multicolumn{1}{c|}{0.232} &
  \multicolumn{1}{c|}{0.782} &
  \multicolumn{1}{c|}{0.717} &
  \multicolumn{1}{c|}{0.792} &
  0.287 \\ \cmidrule{2-8} 
\multicolumn{1}{|c|}{\multirow{-2}{*}{FDE}} &
  std &
  0.077 &
  \multicolumn{1}{c|}{0.121} &
  \multicolumn{1}{c|}{0.116} &
  \multicolumn{1}{c|}{0.117} &
  \multicolumn{1}{c|}{0.117} &
  0.185 \\ \hline
\end{tabular}%
}
\caption{\revisedtext{Quantitative results of trajectory prediction methods. Best performance in {\best green}. Our LSTM method has better ADE and FDE than other GP-based method for trajectory prediction.}}
\label{tab:ablation-gp}
\end{table}

\subsubsection{LSTM vs KF}
\revisedtext{Based on the result outcome above, for a real-time and online running, we compare our LSTM-based approach with the traditional Bayesian filtering as a common, but short-term approach.} We configured the experimental setup in an open space with a single agent $\ag$ and a single target $\ob$, located within $\mi{vr}_\ag$ (\SI{6}{\meter}), while excluding the search process and the coordination with HQ, thus the agent wouldn't switch or be assigned to another task. This ensured that the LSTM module is an independent variable. We  deployed $\ag$ in different locations ($0.9 ~ \mi{vr}_\ag$ distance apart from $\ob$) and $\ob$ in a fixed location. $\ob$  exhibited different behaviors: (1) following a random waypoint; (2) adopting real-world CFD motion, as introduced in \sect{real-world-experiment}. For random waypoint following, we tested under crossing and overtaking encounters, given that $\ob$ follows nearly linear trajectories in an open space, whereas in the real-world CFD model, $\ob$ follows arbitrary non-linear motion. We ran 50 experiments of each LSTM use and non-use  for $\{$random waypoint - crossing; random waypoint - overtaking; and real-world motion$\}$, totaling 300 runs. 

\begin{figure}[t!]
    \begin{minipage}[b]{\columnwidth}
    \centering
    \tablefontsize
        \begin{tabular}{|c|c|c|}
        \hline
        \textbf{condition} & \textbf{t-statistics} & \textbf{p-value ( \textless 0.01)} \\ \hline
        crossing           & -26.72                & True                               \\ \hline
        overtaking         & -7.89                 & True                               \\ \hline
        real-world         & -30.91                & True                               \\ \hline
        \end{tabular}%
    \captionof{table}{Statistics test for LSTM use/non-use. Tracking time based on the usage of LSTM is significantly shorter than non-usage of LSTM. \label{tab:statistics-lstm-use}}
    \end{minipage}
    ~~~\begin{minipage}[b]{\columnwidth}
     \centering
     \includegraphics[clip,width=0.8\columnwidth]{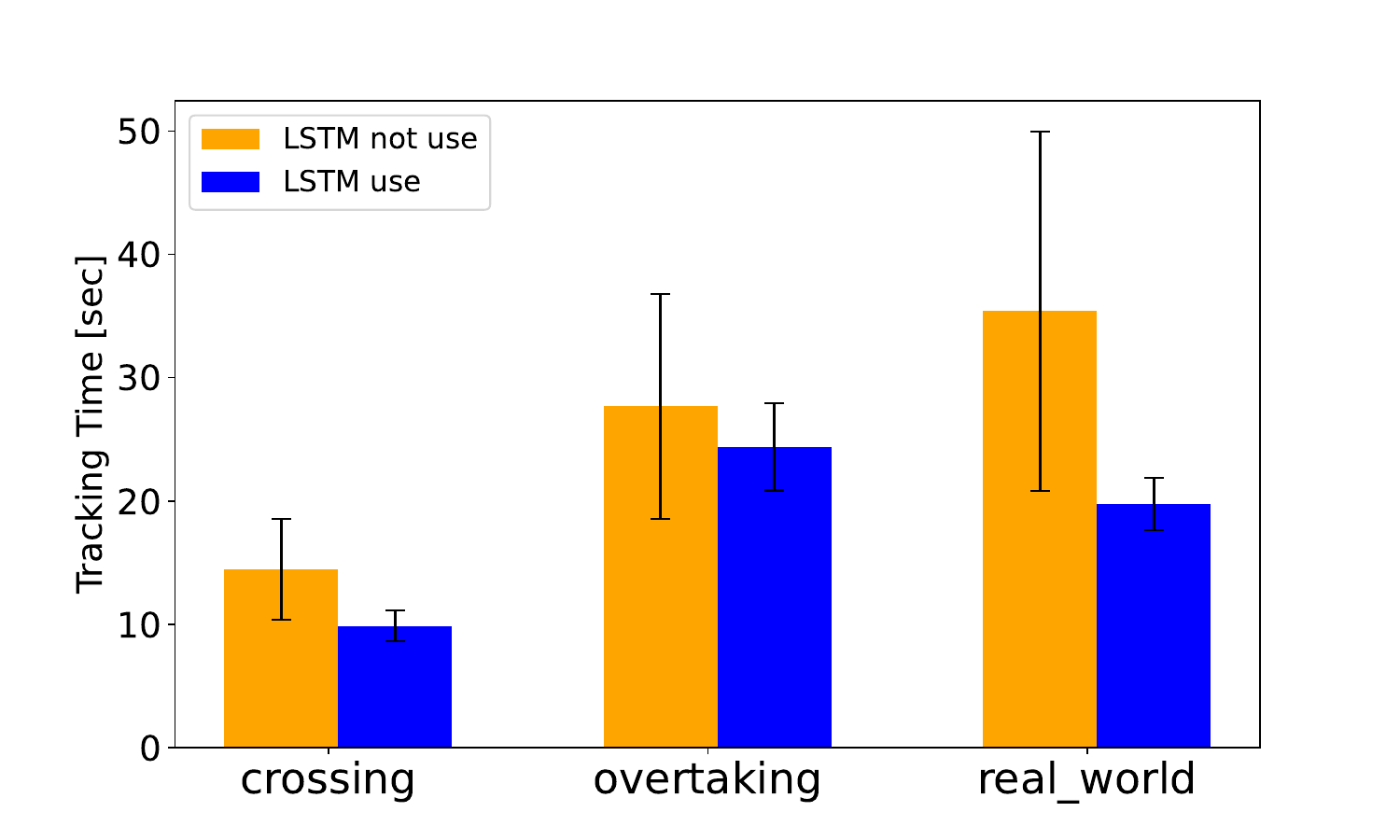}
     \captionof{figure}{Mean and standard deviation of tracking time for LSTM use/non-use---lower is better.  Using LSTM provides advantage for tracking time. \label{fig:barplot-lstm-use}}
    \end{minipage}
    \vspace{-0.2cm}
\end{figure}

\emph{\tab{statistics-lstm-use} and \fig{barplot-lstm-use} show that in all cases, our proposed approach with the LSTM module is significantly faster in terms of tracking time than when LSTM is not used (i.e., using the Kalman Filter approach).} Intuitively, the proposed approach with LSTM enables $\ag$ to take preemptive action (e.g., a trajectory that can lead to the front side of $\ob$ in a crossing situation) to reduce the approach time, while the KF approach predicts and follows $\ob$ based on the last known information.

\subsection{Robustness Test} \label{sec:robustness}
\revisedtext{Our proposed system, featuring multiple agents and external reports in an asynchronous setting, was tested for robustness against unfavorable or unreliable situations. We configured the experiments with the most challenging scenarios, where the number of targets exceeds the number of agents, i.e., by varying the agent-to-target pairs as (\(\num{2}, \num{10}\)), (\(\num{4}, \num{10}\)), and (\(\num{8}, \num{10}\)) under an individual faulty configuration. We conducted a total of \(\num{390}\) experiments (\(\num{10}\) iterations per setting) in an \emph{open} environment to monitor performance isolating it from the impact of the topology of the environment.}

\begin{figure}
    \centering
    \begin{minipage}[t]{0.8\columnwidth}
        \begin{subfigure}[b]{\textwidth}
            \includegraphics[width=\textwidth]{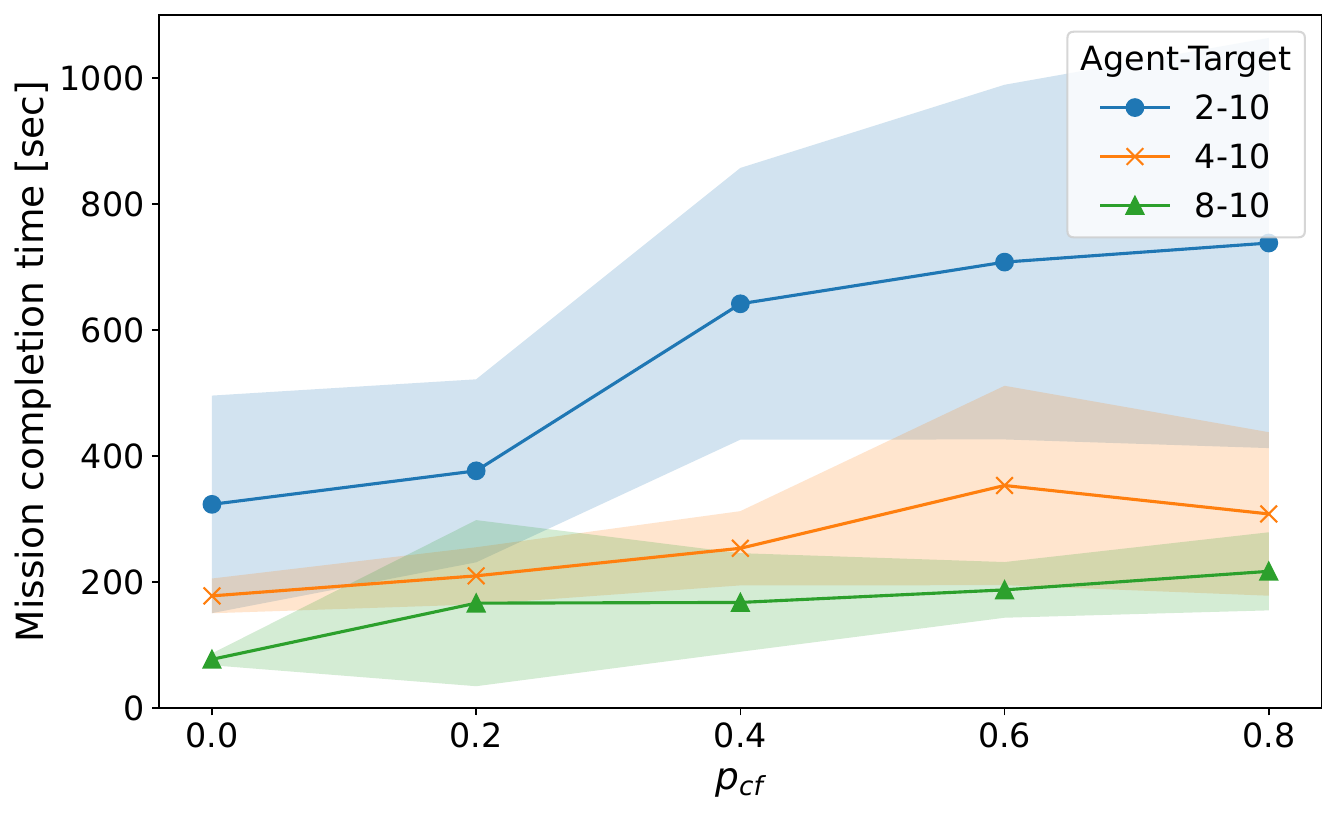}
        \vspace{-0.7cm}
        \caption{}
         \label{fig:robust-comm}
        \end{subfigure}
    \end{minipage}
    \begin{minipage}[t]{0.8\columnwidth}
        \begin{subfigure}[b]{\textwidth}
            \includegraphics[width=\textwidth]{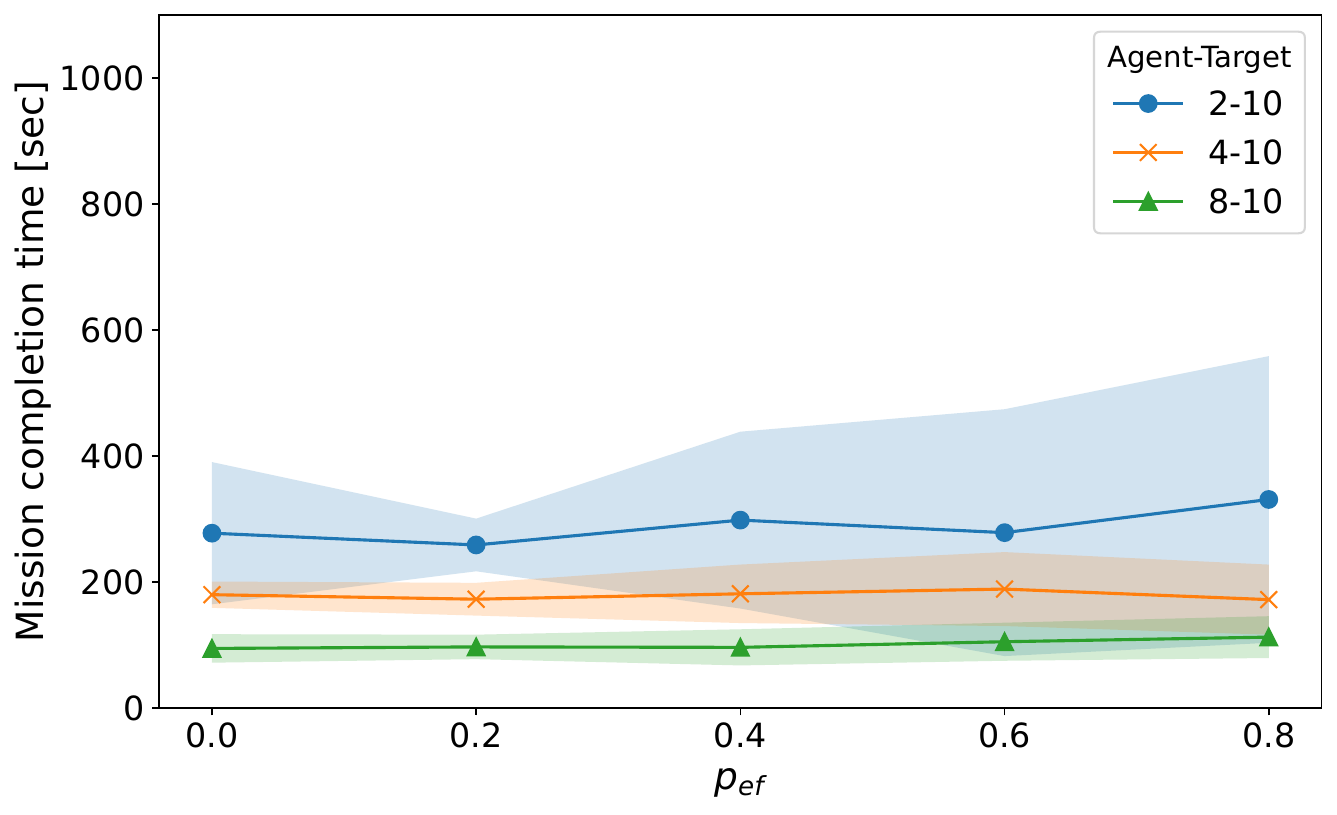}
        \vspace{-0.7cm}
        \caption{}
        \label{fig:robust-ext}
        \end{subfigure}
    \end{minipage}
    \begin{minipage}[t]{0.8\columnwidth}
        \begin{subfigure}[b]{\textwidth}
            \includegraphics[width=\textwidth]{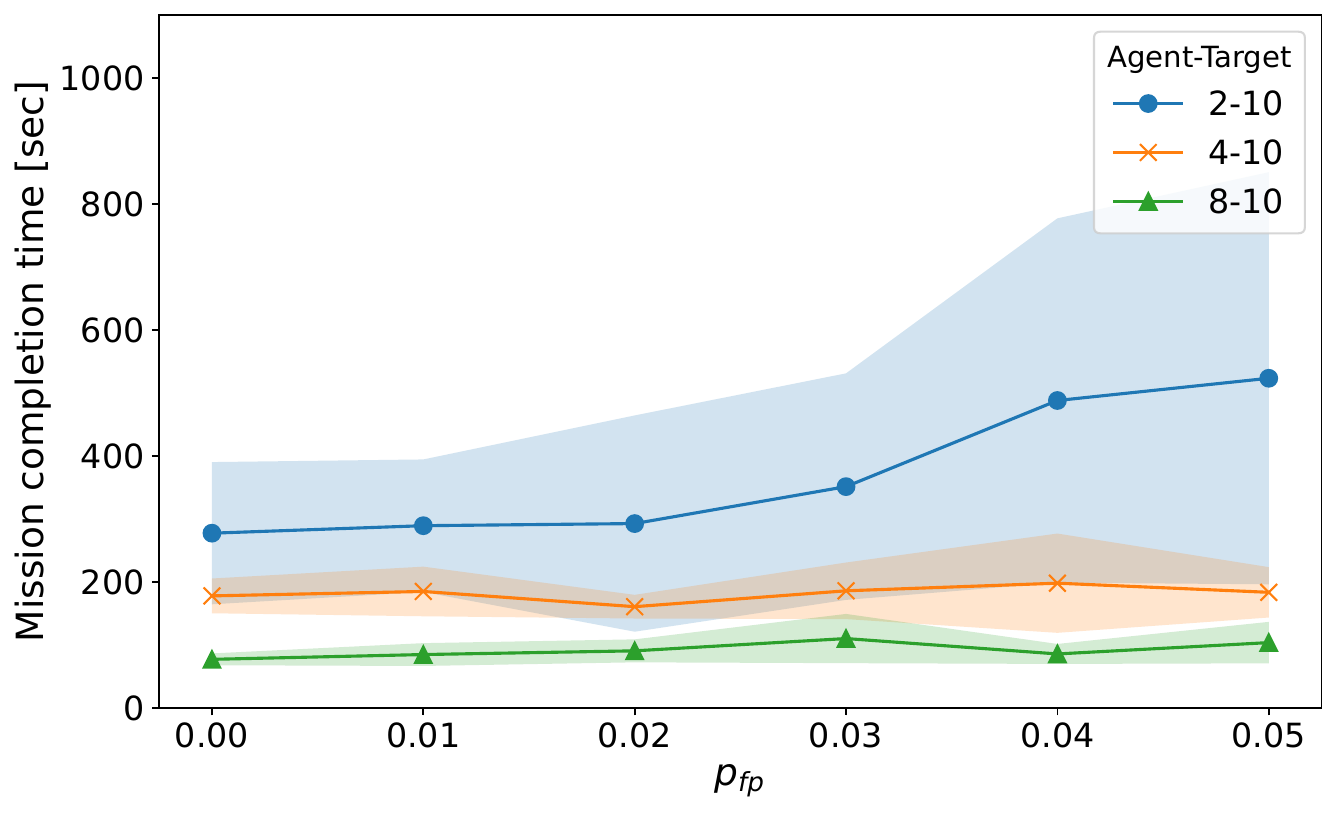}
        \vspace{-0.7cm}
        \caption{}
        \label{fig:robust-fp}
        \end{subfigure}
    \end{minipage}
    \caption{\revisedtext{Robust test of the proposed method (\emph{hybrid}) under varying failure conditions and the ratio of the agent and the target pair in \emph{open} environment. (a) communication breakdown by agents; (b) communication breakdown by external reports; and (c) perception failure.}
    }
    \label{fig:robustness-test}
\end{figure}

\revisedtext{
\textbf{Communication breakdown by agents.} We tested the robustness of our system under unreliable communication of shared information by agents, i.e., a probabilistic breakdown of agents' broadcast to HQ, inspired by real-world examples such as low battery or communication device malfunction. Specifically, we modeled the probability of communication failure ($p_{cf}$) chosen from the set $p_{cf} \in P_{cf} = \{0.2, 0.4, 0.6, 0.8\}$ for each experiment run. As shown in \fig{robust-comm}, our asynchronous and hybrid system can robustly operate even under the loss of communication between agents and HQ. We found   the loss of centralized communication can increase the overall mission time due to not selecting the optimal frontier node during the allocation of the search process; however, the system is operational because our hybrid approach enables agents to conduct search and tracking based on their own decisions, i.e., decentralized way.
}

\revisedtext{
\textbf{Communication breakdown by external reports.} We tested the robustness of our system under unreliable communication of shared information by external reports, i.e., a probabilistic breakdown of third-party reporting to HQ, inspired by real-world examples such as inconsistent feedback from third-party information sources. Specifically, we modeled the probability of communication failure ($p_{ef}$) chosen from the set $p_{ef} \in P_{ef} = \{0.2, 0.4, 0.6, 0.8\}$ for each experiment run. As shown in \fig{robust-ext}, our asynchronous and hybrid system can robustly operate even under the loss of communication between external reports and HQ. We found that the overall mission completion time is not heavily affected, because HQ still uses the most recent information available from external reports and calculates the utility based on it when allocating agents to frontiers for search.
}

\revisedtext{
\textbf{Perception error.} We tested the robustness of our system under unreliable perception performance, i.e., a probabilistic breakdown of sensor returns inspired by real-world examples such as a sensor's false positive (sensor model in \sect{prob-state}) with respect to ground truth. Specifically, we modeled the false positive rate of the sensor on agents ($p_{fp}$) chosen from the set $p_{fp} \in P_{fp} = \{0.01, 0.02, 0.03, 0.04, 0.05\}$, inspired by the state-of-the-art performance of object detection algorithms \cite{Fritsch2013ITSC,Geiger2012CVPR}. As shown in \fig{robust-fp}, our multi-agent search and track system can robustly perform the task even when the perception sensor has unreliable performance thanks to the filtering and shared information as outlined in \sect{main-approach}.
}

\revisedtext{Overall, the mission completion is mostly affected in the most challenging scenario, where the ratio of agents to targets, i.e., (\(\num{2}, \num{10}\)), and particularly when the communication with HQ encounters a breakdown. A thorough analysis of the proposed robustness test can be useful for further decision-making in practical scenarios, for example, determining how many agents are required to find moving targets in a given environment.}

\subsection{Real-world Like Experiment} \label{sec:real-world-experiment}

\begin{figure}[b!]
    \centering
        \begin{subfigure}[b]{0.65\columnwidth}            
            \includegraphics[width=\textwidth,trim={0 0 0 5.1em}, clip]{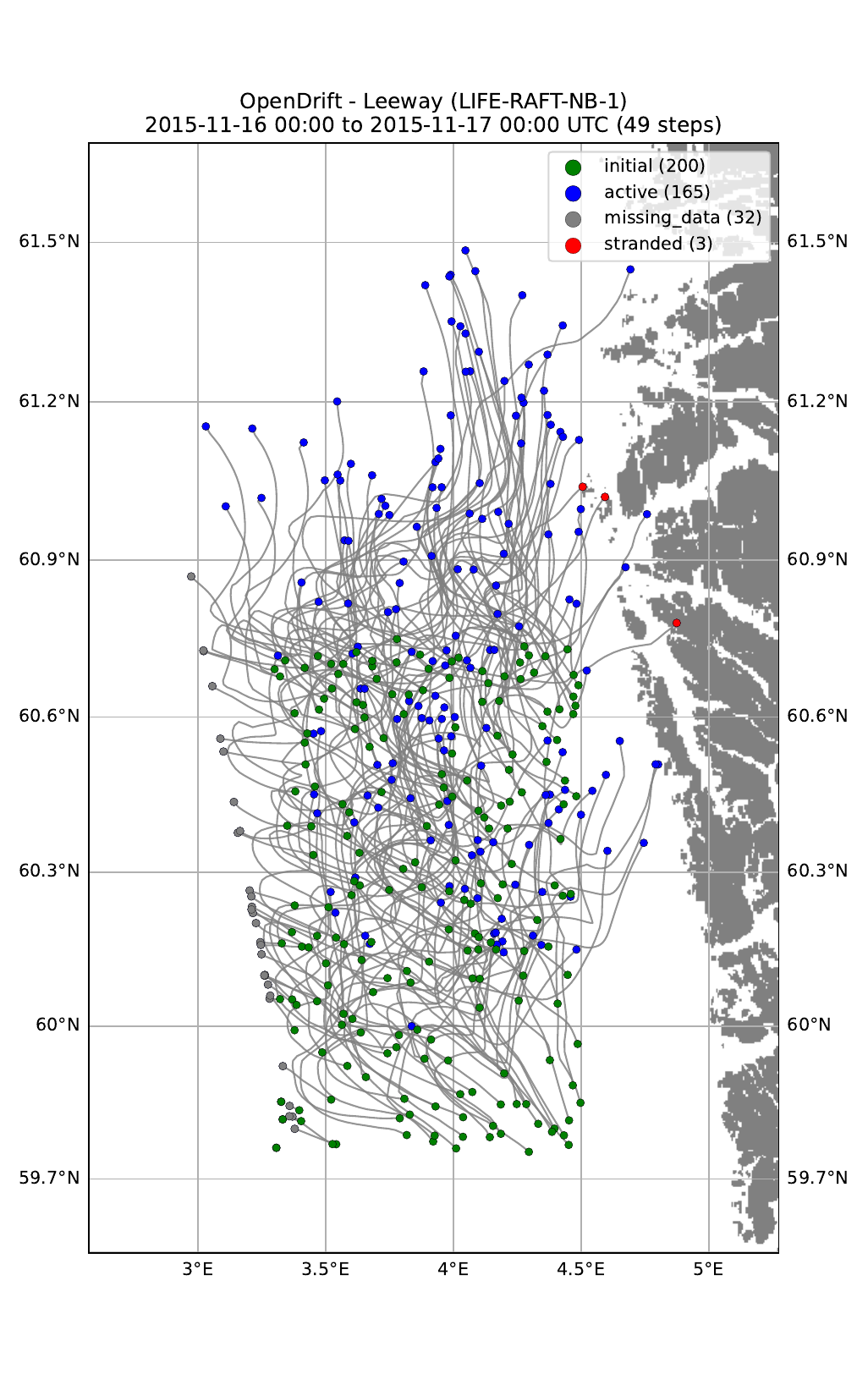}
            \vspace{-3.em}            
        \end{subfigure}
    \caption{Real-world oriented experiments -- Trajectories of floating objects driven by lee-way.}
    \label{fig:open-drift}
\end{figure}

\begin{table*}[]
\centering
\tablefontsize
\resizebox{.89\linewidth}{!}{%
\begin{tabular}{|cc|cc|cc|cc|cc|cc|cc|}
\hline
\multicolumn{1}{|c|}{{\color[HTML]{000000} }} &
  {\color[HTML]{000000} } &
  \multicolumn{2}{c|}{{\color[HTML]{000000} \textbf{Ours}}} &
  \multicolumn{2}{c|}{{\color[HTML]{000000} \textbf{Central KF}}} &
  \multicolumn{2}{c|}{{\color[HTML]{000000} \textbf{Independent}}} &
  \multicolumn{2}{c|}{{\color[HTML]{000000} \textbf{Random}}} &
  \multicolumn{2}{c|}{\revisedtext{\textbf{Exhaustive}}} &
  \multicolumn{2}{c|}{\revisedtext{\textbf{Swarm}}} \\ \cmidrule{3-14} 
\multicolumn{1}{|c|}{\multirow{-2}{*}{{\color[HTML]{000000} \textbf{\begin{tabular}[c]{@{}c@{}}Agent \\ num\end{tabular}}}}} &
  \multirow{-2}{*}{{\color[HTML]{000000} \textbf{\begin{tabular}[c]{@{}c@{}}Target \\ num\end{tabular}}}} &
  \multicolumn{1}{c|}{{\color[HTML]{000000} mean}} &
  {\color[HTML]{000000} std.} &
  \multicolumn{1}{c|}{{\color[HTML]{000000} mean}} &
  {\color[HTML]{000000} std.} &
  \multicolumn{1}{c|}{{\color[HTML]{000000} mean}} &
  {\color[HTML]{000000} std.} &
  \multicolumn{1}{c|}{{\color[HTML]{000000} mean}} &
  {\color[HTML]{000000} std.} &
  \multicolumn{1}{c|}{mean} &
  std. &
  \multicolumn{1}{c|}{mean} &
  std. \\ \hline
\multicolumn{1}{|c|}{{\color[HTML]{000000} }} &
  {\color[HTML]{000000} 2} &
  \multicolumn{1}{c|}{{\color[HTML]{009901} \textbf{90.28}}} &
  {\color[HTML]{000000} 21.03} &
  \multicolumn{1}{c|}{{\color[HTML]{000000} 174.52}} &
  {\color[HTML]{000000} 89.54} &
  \multicolumn{1}{c|}{{\color[HTML]{000000} 196.96}} &
  {\color[HTML]{000000} 103.66} &
  \multicolumn{1}{c|}{{\color[HTML]{000000} 288.04}} &
  {\color[HTML]{000000} 25.11} &
  \multicolumn{1}{c|}{{\color[HTML]{F8A102} 105.22}} &
  41.44 &
  \multicolumn{1}{c|}{300.00} &
  0.00 \\ \cmidrule{2-14} 
\multicolumn{1}{|c|}{{\color[HTML]{000000} }} &
  {\color[HTML]{000000} 5} &
  \multicolumn{1}{c|}{{\color[HTML]{009901} \textbf{105.14}}} &
  {\color[HTML]{000000} 29.91} &
  \multicolumn{1}{c|}{{\color[HTML]{F8A102} 225.62}} &
  {\color[HTML]{000000} 81.36} &
  \multicolumn{1}{c|}{{\color[HTML]{000000} 230.28}} &
  {\color[HTML]{000000} 91.45} &
  \multicolumn{1}{c|}{{\color[HTML]{000000} 298.74}} &
  {\color[HTML]{000000} 3.98} &
  \multicolumn{1}{c|}{286.16} &
  43.77 &
  \multicolumn{1}{c|}{300.00} &
  0.00 \\ \cmidrule{2-14} 
\multicolumn{1}{|c|}{\multirow{-3}{*}{{\color[HTML]{000000} 2}}} &
  {\color[HTML]{000000} 10} &
  \multicolumn{1}{c|}{{\color[HTML]{009901} \textbf{165.94}}} &
  {\color[HTML]{000000} 18.53} &
  \multicolumn{1}{c|}{{\color[HTML]{F8A102} 280.38}} &
  {\color[HTML]{000000} 37.69} &
  \multicolumn{1}{c|}{{\color[HTML]{000000} 288.76}} &
  {\color[HTML]{000000} 35.54} &
  \multicolumn{1}{c|}{{\color[HTML]{000000} 300.00}} &
  {\color[HTML]{000000} 0.00} &
  \multicolumn{1}{c|}{300.00} &
  0.00 &
  \multicolumn{1}{c|}{300.00} &
  0.00 \\ \hline
\multicolumn{1}{|c|}{{\color[HTML]{000000} }} &
  {\color[HTML]{000000} 2} &
  \multicolumn{1}{c|}{{\color[HTML]{F8A102} 69.74}} &
  {\color[HTML]{000000} 6.41} &
  \multicolumn{1}{c|}{{\color[HTML]{009901} \textbf{60.64}}} &
  {\color[HTML]{000000} 7.33} &
  \multicolumn{1}{c|}{{\color[HTML]{000000} 90.92}} &
  {\color[HTML]{000000} 39.27} &
  \multicolumn{1}{c|}{{\color[HTML]{000000} 124.56}} &
  {\color[HTML]{000000} 16.59} &
  \multicolumn{1}{c|}{119.86} &
  52.65 &
  \multicolumn{1}{c|}{280.96} &
  51.77 \\ \cmidrule{2-14} 
\multicolumn{1}{|c|}{{\color[HTML]{000000} }} &
  {\color[HTML]{000000} 5} &
  \multicolumn{1}{c|}{{\color[HTML]{009901} \textbf{82.74}}} &
  {\color[HTML]{000000} 8.08} &
  \multicolumn{1}{c|}{{\color[HTML]{F8A102} 84.06}} &
  {\color[HTML]{000000} 25.93} &
  \multicolumn{1}{c|}{{\color[HTML]{000000} 149.92}} &
  {\color[HTML]{000000} 97.30} &
  \multicolumn{1}{c|}{{\color[HTML]{000000} 166.86}} &
  {\color[HTML]{000000} 69.27} &
  \multicolumn{1}{c|}{227.40} &
  62.43 &
  \multicolumn{1}{c|}{300.00} &
  0.00 \\ \cmidrule{2-14} 
\multicolumn{1}{|c|}{\multirow{-3}{*}{{\color[HTML]{000000} 4}}} &
  {\color[HTML]{000000} 10} &
  \multicolumn{1}{c|}{{\color[HTML]{009901} \textbf{116.60}}} &
  {\color[HTML]{000000} 19.75} &
  \multicolumn{1}{c|}{{\color[HTML]{F8A102} 210.34}} &
  {\color[HTML]{000000} 73.88} &
  \multicolumn{1}{c|}{{\color[HTML]{000000} 253.16}} &
  {\color[HTML]{000000} 81.98} &
  \multicolumn{1}{c|}{{\color[HTML]{000000} 277.44}} &
  {\color[HTML]{000000} 34.99} &
  \multicolumn{1}{c|}{300.00} &
  0.00 &
  \multicolumn{1}{c|}{294.32} &
  17.96 \\ \hline
\multicolumn{1}{|c|}{{\color[HTML]{000000} }} &
  {\color[HTML]{000000} 2} &
  \multicolumn{1}{c|}{{\color[HTML]{009901} \textbf{54.50}}} &
  {\color[HTML]{000000} 7.82} &
  \multicolumn{1}{c|}{{\color[HTML]{000000} 81.68}} &
  {\color[HTML]{000000} 21.73} &
  \multicolumn{1}{c|}{{\color[HTML]{F8A102} 57.42}} &
  {\color[HTML]{000000} 3.76} &
  \multicolumn{1}{c|}{{\color[HTML]{330001} 90.46}} &
  {\color[HTML]{000000} 67.88} &
  \multicolumn{1}{c|}{72.38} &
  9.11 &
  \multicolumn{1}{c|}{278.16} &
  69.06 \\ \cmidrule{2-14} 
\multicolumn{1}{|c|}{{\color[HTML]{000000} }} &
  {\color[HTML]{000000} 5} &
  \multicolumn{1}{c|}{{\color[HTML]{F8A102} 72.54}} &
  {\color[HTML]{000000} 6.07} &
  \multicolumn{1}{c|}{{\color[HTML]{000000} 155.88}} &
  {\color[HTML]{000000} 78.66} &
  \multicolumn{1}{c|}{{\color[HTML]{009901} \textbf{65.08}}} &
  {\color[HTML]{000000} 13.27} &
  \multicolumn{1}{c|}{{\color[HTML]{000000} 89.08}} &
  {\color[HTML]{000000} 75.15} &
  \multicolumn{1}{c|}{73.90} &
  11.14 &
  \multicolumn{1}{c|}{142.60} &
  66.27 \\ \cmidrule{2-14} 
\multicolumn{1}{|c|}{\multirow{-3}{*}{{\color[HTML]{000000} 8}}} &
  {\color[HTML]{000000} 10} &
  \multicolumn{1}{c|}{{\color[HTML]{009901} \textbf{132.64}}} &
  {\color[HTML]{000000} 62.08} &
  \multicolumn{1}{c|}{{\color[HTML]{000000} 186.30}} &
  {\color[HTML]{000000} 55.37} &
  \multicolumn{1}{c|}{{\color[HTML]{F8A102} 164.96}} &
  {\color[HTML]{000000} 83.60} &
  \multicolumn{1}{c|}{{\color[HTML]{000000} 255.50}} &
  {\color[HTML]{000000} 58.36} &
  \multicolumn{1}{c|}{292.92} &
  22.39 &
  \multicolumn{1}{c|}{274.80} &
  53.59 \\ \hline
\multicolumn{2}{|c|}{{\color[HTML]{000000} average}} &
  \multicolumn{1}{c|}{{\color[HTML]{000000} \textbf{98.90}}} &
  {\color[HTML]{000000} 19.96} &
  \multicolumn{1}{c|}{{\color[HTML]{000000} 162.16}} &
  {\color[HTML]{000000} 52.39} &
  \multicolumn{1}{c|}{{\color[HTML]{000000} 166.38}} &
  {\color[HTML]{000000} 61.09} &
  \multicolumn{1}{c|}{{\color[HTML]{000000} 210.08}} &
  {\color[HTML]{000000} 39.04} &
  \multicolumn{1}{c|}{197.54} &
  26.99 &
  \multicolumn{1}{c|}{274.54} &
  28.74 \\ \hline
\end{tabular}%
}
\caption{Mission time (sec) in the real-world oriented experiments \revisedtext{with a cut-off time at $\SI{300}{\second}$---the lower the better across each row}. Best performance in {\best green}, second best in {\bestsecond orange}. Our method has better completion time than other baseline methods. %
}
\label{tab:real-world-mission-time}
\end{table*}

To further test the applicability of our proposed method in real-world environments. we validated it \revisedtext{(1) with real-world target motions in an aquatic domain; and (2) with physics-enabled 3D simulator in indoor environments. This can be generalized to tasks such as search and rescue or oil spill response.}

\revisedtext{\textbf{Aquatic domain.}} We used \texttt{OpenDrift} \cite{opendrift-2018} to randomly generate leeway and trajectories of particles based on a Computational Fluid Dynamics (CFD) model (\fig{open-drift}). We resized the scale to \qtyproduct{50 x 50}{m} considering the size of the robots, their speed, and the 5-minute time limit. We maintained the same experimental parameters and configurations for the number of agents and targets. Given the different characteristics of the motion, we separately trained our LSTM-MLP module using randomly generated trajectories based on CFD simulations \revisedtext{collected at via \texttt{OpenDrift}}. Out of a total of 200 generated floating objects, 152 were used for training with data augmentation and 38 for testing. An additional 10 objects with their corresponding trajectories, which were never used in training or testing, were used for the actual search and track experiments---a total of \revisedtext{\num{540} runs (\num{90} runs per method)}. Our proposed algorithm (hybrid) still uses third party reporting (e.g., vessels in nearby transit report to Coast Guard stations). 

\emph{The results (\tab{real-world-mission-time}) show that our method outperformed the baselines and validate that our proposed method can be generically applied to real-world applications across various domains, despite arbitrary non-linear motions of target objects.}  Many cases were showing significant difference between ours (best) and the 2nd best with $p\textrm{-value} < 0.01$. For example, agent, target (\num{4}, \num{5}) showed $p\textrm{-value}=0.001$ by $t\textrm{-test}$ between ours and central KF. On the other hand, there were two cases where other baselines showed slightly better performance, e.g., agent, target (\num{8}, \num{5}) $p\textrm{-value} = 0.1232$ and  (\num{4}, \num{2}) $p\textrm{-value} = 0.1095$, where, however, $t\textrm{-test}$ results do not show a significant difference.

\begin{figure}[t!]
    \centering
    \begin{minipage}[t]{0.42\columnwidth}
        \begin{subfigure}[b]{\textwidth}
            \includegraphics[width=\textwidth]{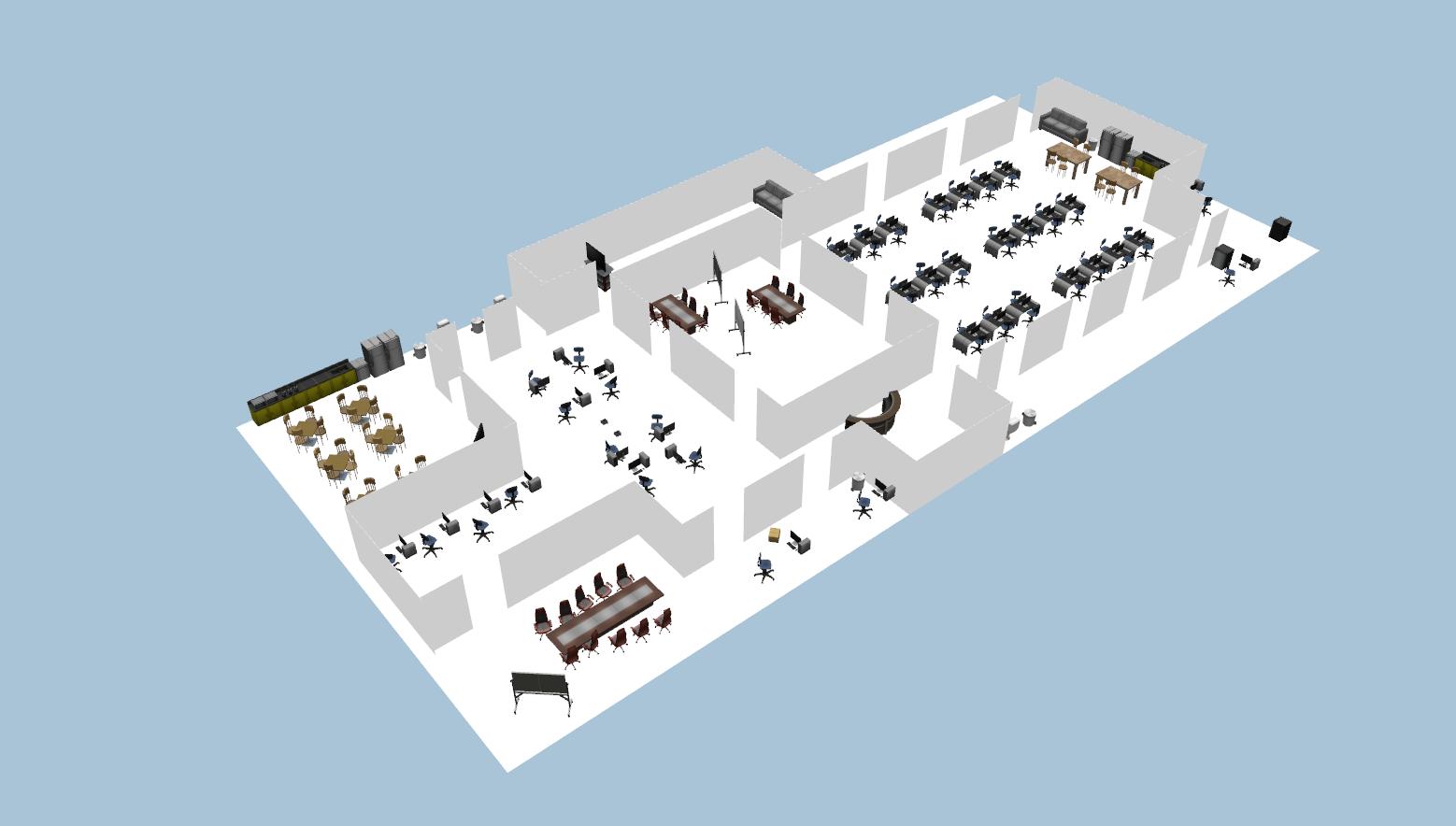}
        \end{subfigure}
    \end{minipage}
    ~
    \begin{minipage}[t]{0.47\columnwidth}
        \begin{subfigure}[b]{\textwidth}
            \includegraphics[width=\textwidth]{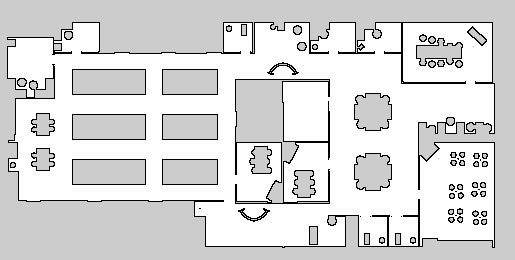}
        \end{subfigure}
    \end{minipage}
    \caption{\revisedtext{3D simulation example environment: \qtyproduct{50 x 25}{m} \emph{office1} model based on AWS office. (\textit{left}): deployed 3D model (\textit{right}): occupancy grid map that shows the cluttered characteristic}.}
    \label{fig:gazebo-env}
\end{figure}

\begin{table}[t!]
\tablefontsize
\centering
\resizebox{\columnwidth}{!}{%
\begin{tabular}{|c|c|c|c|c|c|c|}
\hline
{\color[HTML]{000000} \textbf{Env.}} &
  {\color[HTML]{000000} \textbf{Ours}} &
  {\color[HTML]{000000} \textbf{Cent. KF}} &
  {\color[HTML]{000000} \textbf{Independ.}} &
  {\color[HTML]{000000} \textbf{Rand.}} &
  \textbf{Exhaust.} &
  \textbf{Swarm} \\ \hline
{\color[HTML]{000000} Empty} &
  {\color[HTML]{009901} \textbf{577.4}} &
  {\color[HTML]{F8A102} 1141.6} &
  {\color[HTML]{000000} 1603.6} &
  {\color[HTML]{000000} 1317.2} &
  {\color[HTML]{000000} 3219.4} &
  2431.4 \\ \hline
{\color[HTML]{000000} Office1} &
  {\color[HTML]{009901} \textbf{446.2}} &
  {\color[HTML]{000000} 652.4} &
  {\color[HTML]{000000} 1068} &
  {\color[HTML]{000000} 1143.8} &
  {\color[HTML]{F8A102} 572.6} &
  - \\ \hline
{\color[HTML]{000000} Office2} &
  {\color[HTML]{009901} \textbf{537.6}} &
  {\color[HTML]{000000} 1703.6} &
  {\color[HTML]{000000} Fail (0.9)} &
  {\color[HTML]{000000} 1161.8} &
  {\color[HTML]{F8A102} 925.4} &
  - \\ \hline
\end{tabular}%
}
\caption{\revisedtext{Mission time (sec) in the 3D simulation experiments for the indoor domain as the most challenging agent, target (\num{2},\num{10}) scenario without a cut-off time---the lower the better across each row. Best performance in {\best green}, second best in {\bestsecond orange}. Our method has better completion time than other baseline methods and ensure the search and track task is completed. Note that \emph{Independent} in \emph{office2} had searched the entire region but failed to track an object and we report the track ratio instead.}}
\label{tab:gazebo-result}
\end{table}

\begin{figure}
    \centering
    \begin{minipage}[t]{0.8\columnwidth}
        \begin{subfigure}[b]{\textwidth}
            \includegraphics[width=\textwidth]{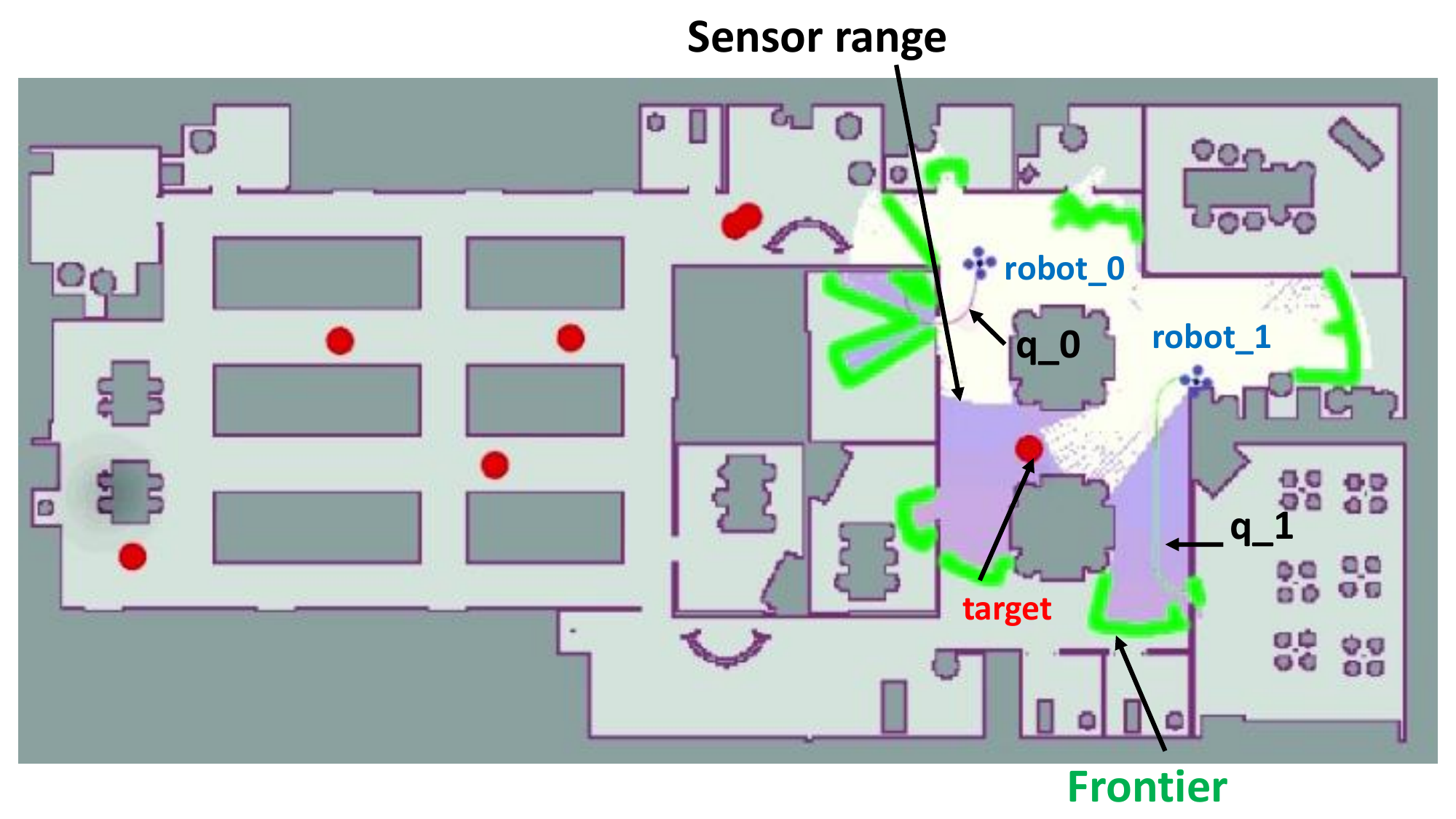}
        \vspace{-0.8cm}
        \caption{\SI{87}{\second}}
        \end{subfigure}
    \end{minipage}
    \begin{minipage}[t]{0.8\columnwidth}
        \begin{subfigure}[b]{\textwidth}
            \includegraphics[width=\textwidth]{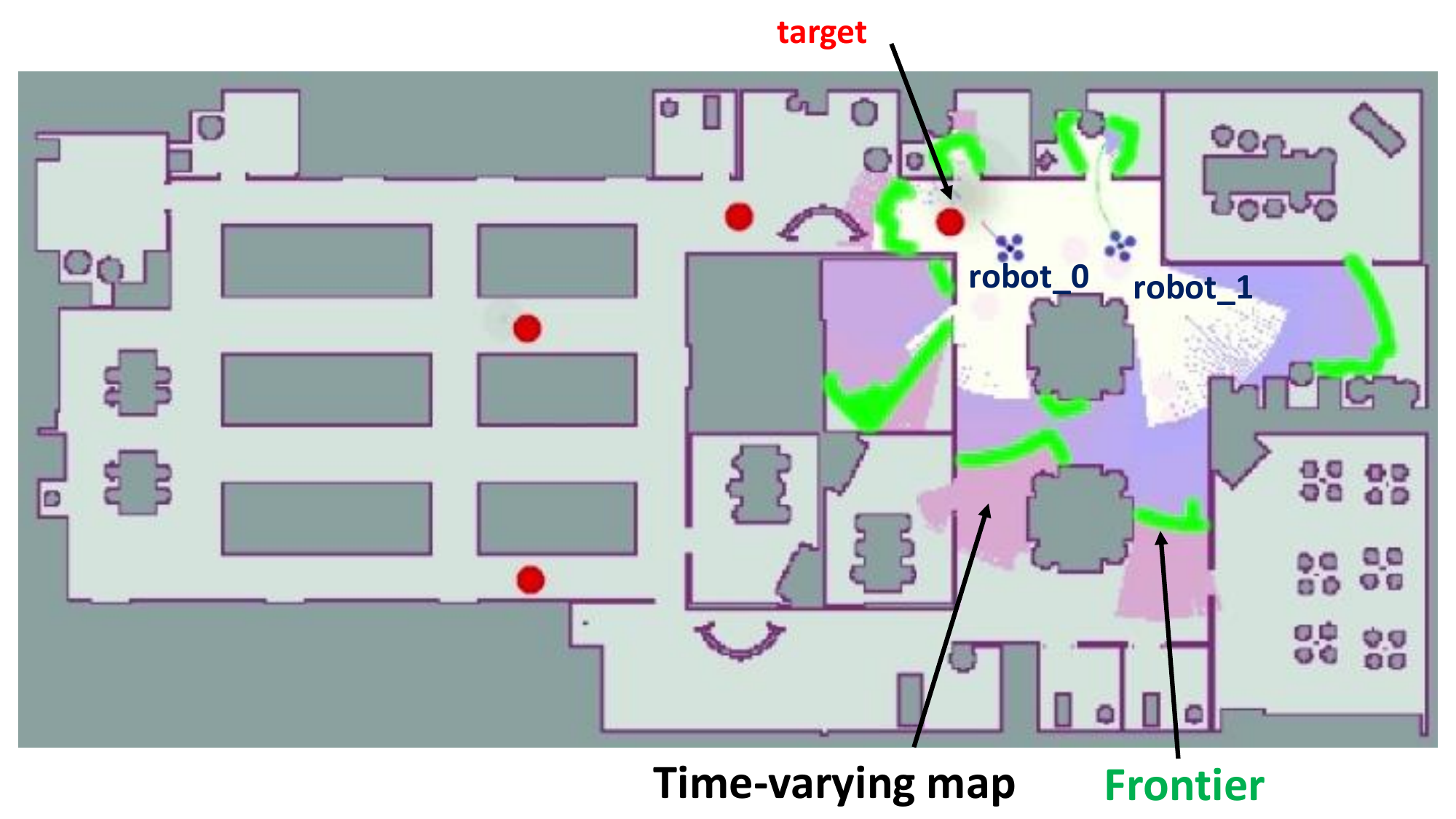}
        \vspace{-0.6cm}    
        \caption{\SI{213}{\second}}
        \end{subfigure}
    \end{minipage}
    \begin{minipage}[t]{0.8\columnwidth}
        \begin{subfigure}[b]{\textwidth}
            \includegraphics[width=\textwidth]{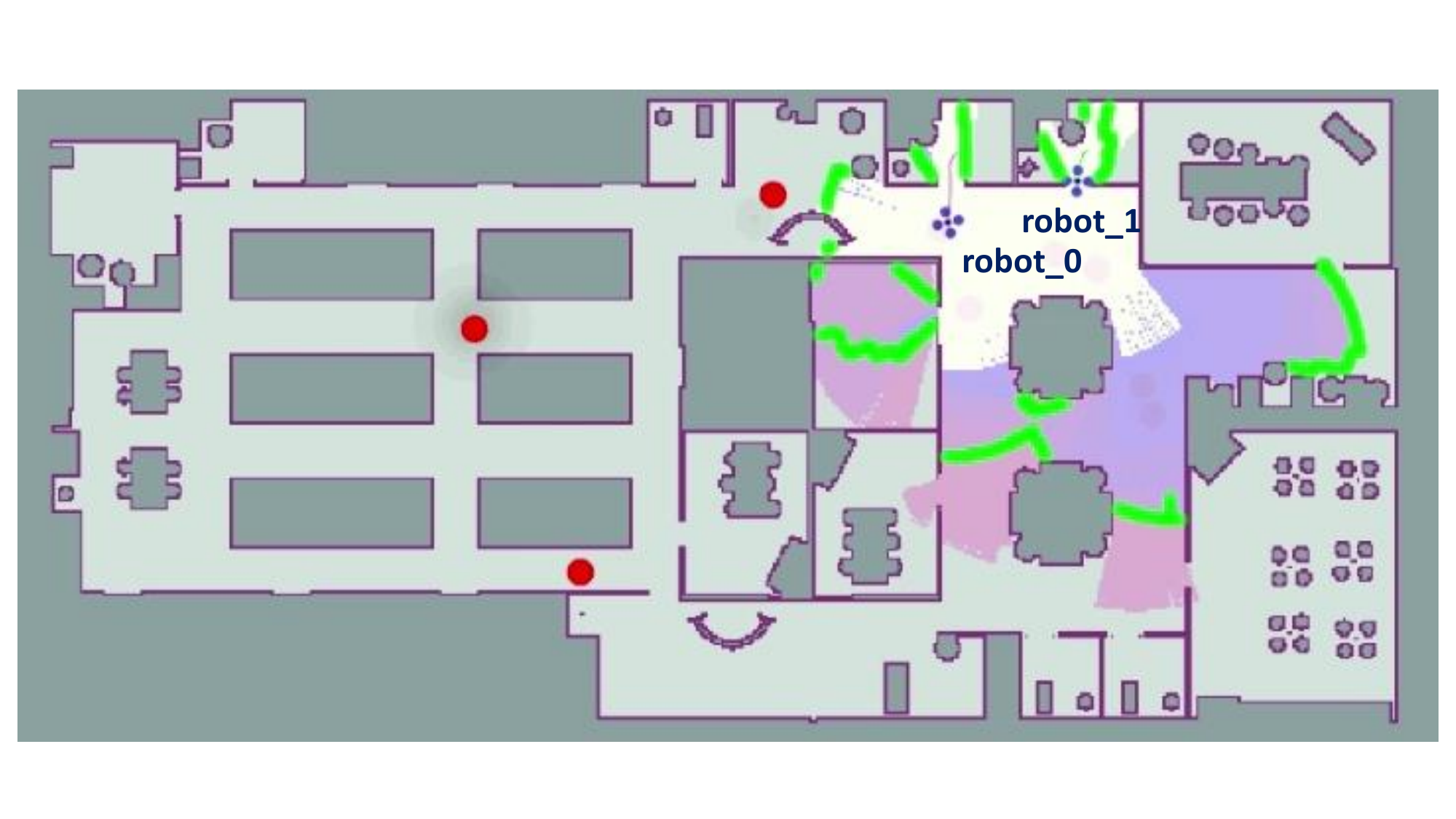}
            \vspace{-0.9cm}  
            \caption{\SI{225}{\second}}
        \end{subfigure}
    \end{minipage}
    \vspace{-0.2cm}  
    \caption{\revisedtext{Snapshots of a 3D simulator experiment using our method in \emph{office1}. (a) Robots $0$ and $1$ are in search mode, following trajectories $q_0$ and $q_1$, respectively. The target (marked) is not yet within their sensor fields of view, although third-party reports have been received about it. (b) Both robots continue their tasks after clearing the previous target. Robot $0$ detects a new target (marked) and switches to tracking mode based on its prediction, while Robot $1$ remains in search mode. The frontier formed by the shared belief map with HQ changes over time, differently from 87 seconds, as uncertainty evolves. (c) Robot $0$ clears the target, and Robot $1$ enters a room during the search. Note that the pink color indicates areas of higher uncertainty than the blue ones, while the current sensor footage is shown in white.
    }
    }
    \label{fig:gazebo-snapshot}
\end{figure}

\revisedtext{
\textbf{Indoor domain.} We used \texttt{Gazebo} \cite{gazebo_2004} to validate the proposed algorithm with realistic physical properties and high-fidelity performance in ground robots, particularly \cite{gazebo-fidelity-1, gazebo-fidelity-2}. We deployed the commonly used \texttt{Turtlebot} \cite{turtlebot-2017}, equipped with a set of sensors: laser scanner, odometry, inertial measurement unit (IMU), and camera, in the following 3D environments \qtyproduct{50 x 50}{m} (\emph{empty}, \emph{office2}) and \qtyproduct{50 x 25}{m} (\emph{office1})—see \fig{gazebo-env}. We configured the most challenging scenario with the number of agent-target pairs as (\num{2}, \num{10}); moreover, unlike in previous experiments, we set an unlimited time cap to demonstrate the feasibility of task completion and the robustness of the algorithm until all targets are found. \emph{The results in \tab{gazebo-result} show that our method outperformed the baselines and achieved completion despite the cluttered indoor environments with complex topology.} As shown in \fig{gazebo-snapshot}, our proposed method can achieve (1) revisiting of searched areas with the time-varying belief map, and (2) convergence to the target locations, despite the high ratio of the number of targets to agents in the cluttered environment.
}

\revisedtext{\subsection{Discussions}}
\revisedtext{
Our proposed method, which fully integrates the system pipeline—detect, identify, communicate, coordinate, plan, and control—including heterogeneous agents and external sources, was rigorously tested across thousands of runs. Our extensive runs yielded significantly better results than the baselines, thereby validating the integrity of our approach in diverse environments. Consequently, our fully integrated system demonstrated robustness and proved to be inherently applicable to new domains, such as indoor environments, without additional steps when tested even with physics-enabled simulations aimed at real-world applications. This contrasts with some studies in the literature (e.g., \cite{usv-pursuit-evasion-2023, upenn-balance-2015}) that provide only a few run examples, which might not sufficiently justify their generality, even when hardware experiments are conducted (e.g., \cite{increasing-autonomy-csat-2009, optimal-swarm-2020}).}

\revisedtext{Throughout the implementation and validation of the baselines, we gained the following insights: First, \emph{swarm-based} approach that utilize self-organization suffer from poor mission completion and tracking ratios (see the Appendix) when the number of agents is small; however, it remains unknown how many swarms are required for mission success, particularly being challenged when interacting with a cluttered environment. The information-driven approach that fully relies on the \emph{central} level might not necessarily enhance performance and could conflict with the objectives of individual agents, as shown in \fig{central-200}. Likewise, the \emph{independent} approach could suffer from the opposite (\fig{independent-200}). Sometimes, a straightforward \emph{exhaustive} approach demonstrates reasonable performance (ranking seconds best in several instances, as shown in \tab{mission} and \tab{gazebo-result}) but may consume unnecessary resources, traveling longer distances, without offering adaptiveness in real-world applications.}

\revisedtext{
Despite its best-in-class performance, our system currently operates based on a priori map of the deployed environment. We cannot completely exclude the possibility that the environment may change or that agents may be deployed in unknown areas. Further studies are required to understand the level of environmental knowledge needed and to guarantee task completion.
}

\section{Conclusions and Future Work}
We presented a complete approach for search and tracking of unknown target objects with a team of autonomous agents. Our approach \revisedtext{contributes to: (1) a time-varying weighted belief representation based on uncertainty and trust, ensuring the successful completion of search and track tasks with heterogeneous agents, third-party reports, and dynamic targets; (2) LSTM-based trajectory prediction that enhances detection probabilities in time-configuration space and long-horizon decision-making that adopts the prediction in real-time; (3) a fully integrated pipeline for multiple agents, merging information-theoretic optimization with mode switching to provide comprehensive search and track capabilities; and (4) extensive Monte Carlo simulations, including ablation studies, which offer insights into the search and track problem for both our method and other state-of-the-art approaches, along with real-world applications using Computational Fluid Dynamic datasets and a 3D physics-based engine robotic simulator, validating the generalization of our proposed method across various domains.} Experiments showed the effectiveness of our proposed approach, \revisedtext{demonstrating that it is $1.3$ to $3.2$ times faster at finding all targets, even under the most challenging scenarios where the ratio of agents to targets is $1:5$.} \emph{We will release our implementation of our method and baselines when the paper is published.}

Interesting future work will involve theoretical analysis about the search phase, exploiting visibility constraint intuitions to keep the area cleared~\cite{quattrini2018search,wang2022marine} or to include explicit multi-agent coordination during the track phase \revisedtext{that can be further applicable to interaction-awareness with deployed environments. We plan to enhance the time-varying belief map with topology of complex environment and the long-horizon decision-making capabilities with an attention-based mechanism to focus on intentions.} We are also looking at implementing the proposed approach on \revisedtext{real robots} and 
extending it for adversarial scenarios.

Ultimately, our proposed method, which considers real scenarios, has the potential to benefit decision-makings and many high-impact applications, such as search and rescue.

\section*{Declarations}

\paragraph{Author contribution}
\input{author-contributions.txt}

\paragraph{Funding Information}
 The work of Mingi Jeong and
Alberto Quattrini Li was partially supported by the National Science Foundation (NSF) under Grant 1923004 and Grant 2144624. The work of Cristian Molinaro and Andrea Pugliese was supported by Project SERICS funded through the MUR National Recovery and Resilience Plan by the European Union—NextGenerationEU under Grant PE00000014. The work of Eugene Santos Jr. was supported by AFOSR under Grant FA9550-20-1-0032. The work of V. S. Subrahmanian was supported in part by ARO under Grant W911NF2320240 and in part by ONR under Grant N00014-20-1-2407. The work of Youzhi Zhang was supported by the InnoHK Fund.

\paragraph{Ethical Statements}
This research did not use any sensitive data. This research can have positive broader impacts in several applications, such as search and rescue, animal tracking, environmental monitoring. Some applications, such as surveillance, need careful thought in the deployment of the proposed method so that it respects human rights, privacy, etc., requiring corresponding policies, which are not the focus of this paper.

\paragraph{Data Availability}
Code will be made available opensource.

\appendix
\input{appendix}

\bibliography{IEEEabrv,references}

\vspace{0.5em}
    \setlength\intextsep{0pt} %
    \begin{wrapfigure}{l}{0.13\textwidth}
        \centering
        \includegraphics[width=0.15\textwidth]{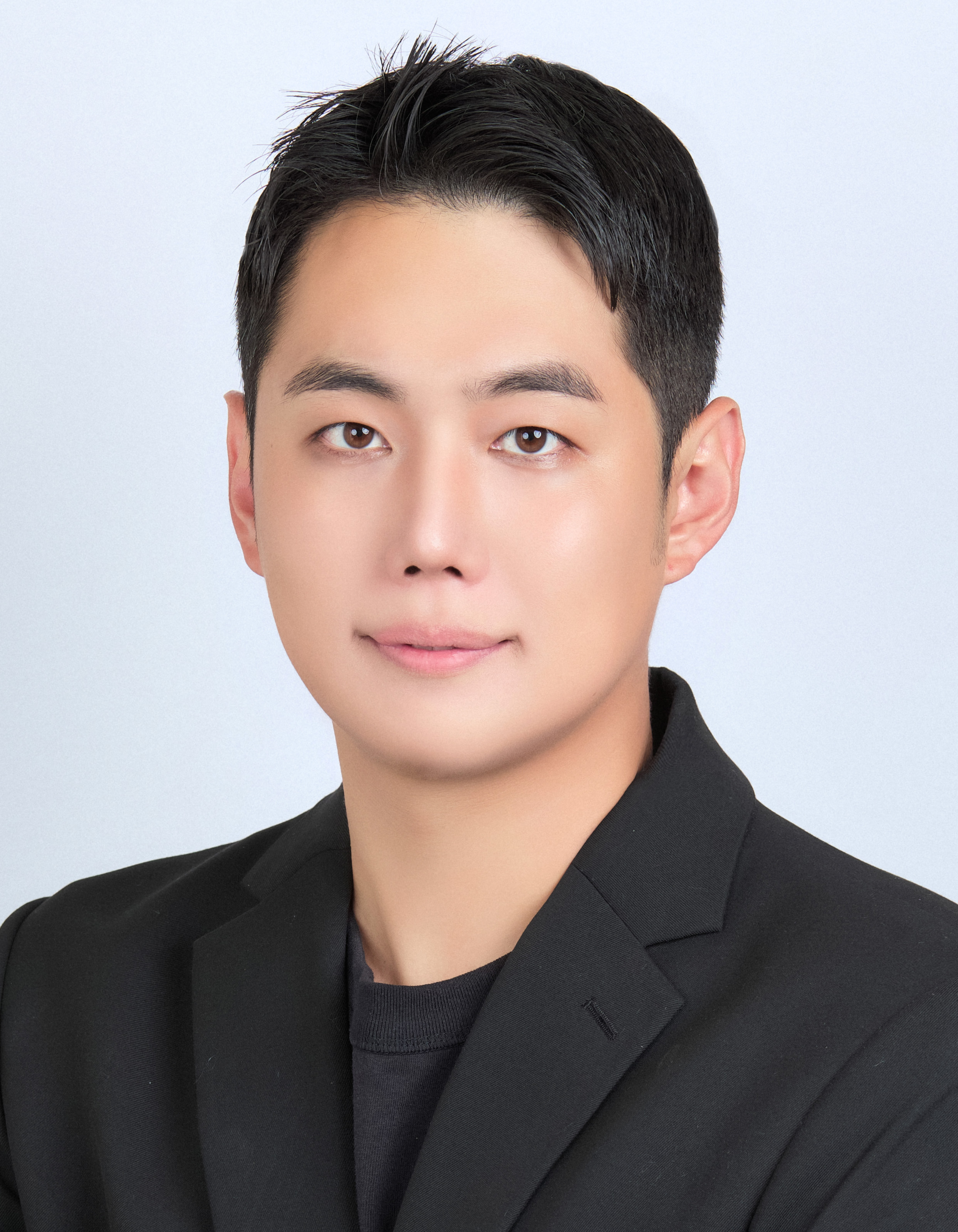}
    \end{wrapfigure}
    \noindent \textbf{Mingi Jeong} is currently pursuing the Ph.D. degree of Reality
and Robotics Lab with the Department of Computer
Science, Dartmouth College, Hanover, NH, USA.
His current interests are autonomous navigation,
multirobot system, and maritime collision avoidance
decision making.
\vspace{0.5em}

    \setlength\intextsep{0pt} %
    \begin{wrapfigure}{l}{0.13\textwidth}
        \centering
        \includegraphics[width=0.15\textwidth]{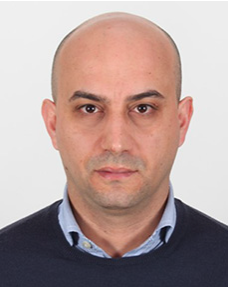}
    \end{wrapfigure}
    \noindent \textbf{Cristian Molinaro} received the Ph.D. degree
in computer engineering from the University of
Calabria, Arcavacata, Italy.
He is an Associate Professor with the DIMES
Department, University of Calabria. His main
research interests include knowledge representation
and reasoning and explainable AI.
\vspace{0.5em}

    \setlength\intextsep{0pt} %
    \begin{wrapfigure}{l}{0.13\textwidth}
        \centering
        \includegraphics[width=0.15\textwidth]{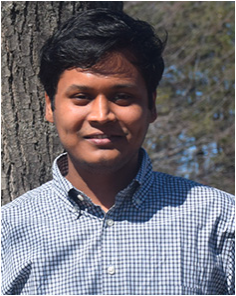}
    \end{wrapfigure}
    \noindent \textbf{Tonmoay Deb} is currently pursuing the Ph.D.
degree with the Department of Computer Science,
Northwestern University, Evanston, IL, USA.
He is doing research under the supervision of
Professor V. S. Subrahmanian with Northwestern
Security and AI Lab. His research interests are
in the intersection of machine learning, computer
vision, natural language processing, and multiagent
systems.
\vspace{0.5em}

    \setlength\intextsep{0pt} %
    \begin{wrapfigure}{l}{0.13\textwidth}
        \centering
        \includegraphics[width=0.15\textwidth]{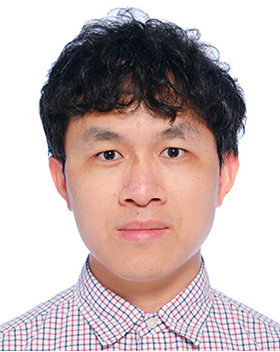}
    \end{wrapfigure}
    \noindent \textbf{Youzhi Zhang} received the Ph.D. degree in computer science from Nanyang Technological University, Singapore.
He is an Assistant Professor with the Centre
for Artificial Intelligence and Robotics, Hong
Kong Institute of Science and Innovation, Chinese
Academy of Sciences, Hong Kong. His research
interests include AI, multiagent systems, and computational
game theory.
\vspace{0.5em}

\vfill\eject

    \setlength\intextsep{0pt} %
    \begin{wrapfigure}{l}{0.13\textwidth}
        \centering
        \includegraphics[width=0.15\textwidth]{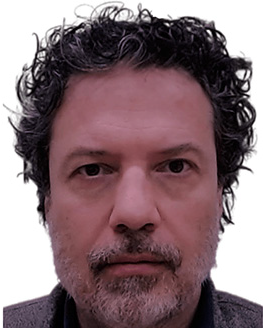}
    \end{wrapfigure}
    \noindent \textbf{Andrea Pugliese} received the
Ph.D. degree in computer engineering from the
University of Calabria, Arcavacata, Italy.
He is an Associate Professor with the DIMES
Department, University of Calabria. His main
research interests include computer security and
graph data management.
\vspace{0.5em}

    \setlength\intextsep{0pt} %
    \begin{wrapfigure}{l}{0.13\textwidth}
        \centering
        \includegraphics[width=0.15\textwidth]{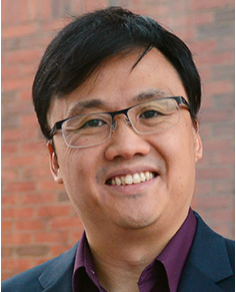}
    \end{wrapfigure}
    \noindent \textbf{Eugene Santos Jr.} received the
Ph.D. degree in computer science from Brown
University, Providence, RI, USA.
He is the Sydney E. Junkins 1887 Professor of
Engineering with the Thayer School of Engineering
and a Adjunct Professor of Computer Science with
Dartmouth College, Hanover, NH, USA. His current
focus is on computational intent, dynamic human
behavior, and decision-making with an emphasis
on learning nonlinear and emergent behaviors and
explainable AI.
\vspace{0.5em}

    \setlength\intextsep{0pt} %
    \begin{wrapfigure}{l}{0.13\textwidth}
        \centering
        \includegraphics[width=0.15\textwidth]{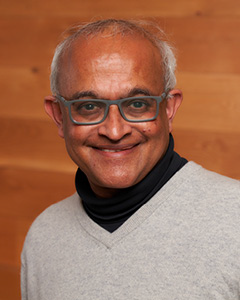}
    \end{wrapfigure}
    \noindent \textbf{V.\ S.\ Subrahmanian} received the M.Sc. (Tech)
degree in computer science from the Birla Institute
of Technology and Science, Pilani, India, and the
M.S. and Ph.D. degrees in computer science from
Syracuse University, Syracuse, NY, USA.
He is the Walter P. Murphy Professor of Computer
Science and a Fellow with the Buffett Institute for
Global Affairs, Northwestern University, Evanston,
IL, USA. He works at the intersection of AI and
security issues.
\vspace{0.5em}

    \setlength\intextsep{0pt} %
    \begin{wrapfigure}{l}{0.13\textwidth}
        \centering
        \includegraphics[width=0.15\textwidth]{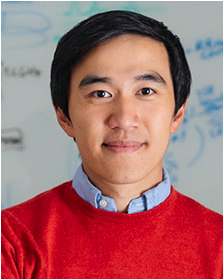}
    \end{wrapfigure}
    \noindent \textbf{Alberto Quattrini Li} received the Ph.D. degree in
computer science and engineering from Polytechnic
of Milan, Milan, Italy.
He is an Associate Professor with the Computer
Science Department, Dartmouth College, Hanover,
NH, USA, and a Co-Director of the Reality
and Robotics Lab. His research interests include
autonomous mobile robotics, artificial intelligence,
and agents and multiagent systems.
\vspace{0.5em}

\end{document}

%% file: author-contributions.txt
MJ, AQL, ESJ, VSS contributed to the study conception and design. MJ, AQL performed implementation, data collection, and analysis. The first draft of the manuscript was written by MJ and AQL, with major input from CM, ESJ, and VSS. All authors commented on previous versions of the manuscript. All authors contributed intellectually to regular discussions towards this manuscript and approved the final manuscript.

%% file: appendix.tex
\section*{Appendix}
This appendix contains additional implementation/experimental details, mathematical and experimental details.

We also included a document with additional results and a video showing the behavior of our proposed method in some experiments and the code \revisedtext{of the proposed method, as well as of all the baselines we implemented}. %

\section{LSTM-based prediction details}
\begin{figure}[b!]
  \centering
  \includegraphics[width=\columnwidth]{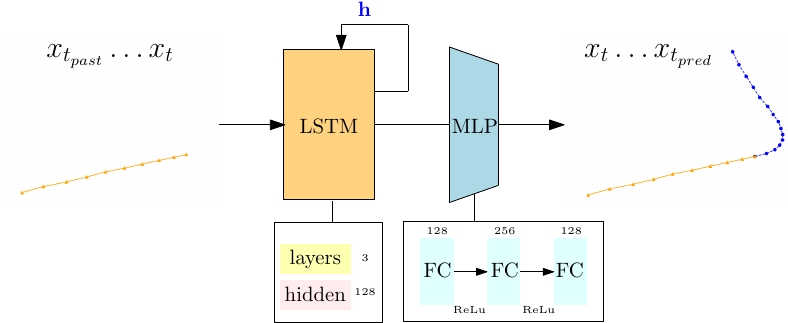}
  \caption{LSTM-MLP module architecture. The {\inputtraj input trajectory} has coordinates from $x_{t_{\textit{past}}}$ to $x_t$ and the {\outputtraj output trajectory} has coordinates from $x_t$ to $x_{t_{\textit{pred}}}$.} 
  \label{fig:lstm-arch}
\end{figure}

Our proposed model for predicting trajectories includes a Long Short-Term Memory (LSTM) layer, designed to understand the temporal relationships within the trajectory data, and a three-layer multi-layer perceptron (MLP) that outputs the predicted trajectory (see \fig{lstm-arch}). 

\begin{figure}[t!]
    \centering
    \begin{minipage}[t]{0.49\columnwidth}
        \begin{subfigure}[b]{\textwidth}            
            \includegraphics[width=\textwidth]{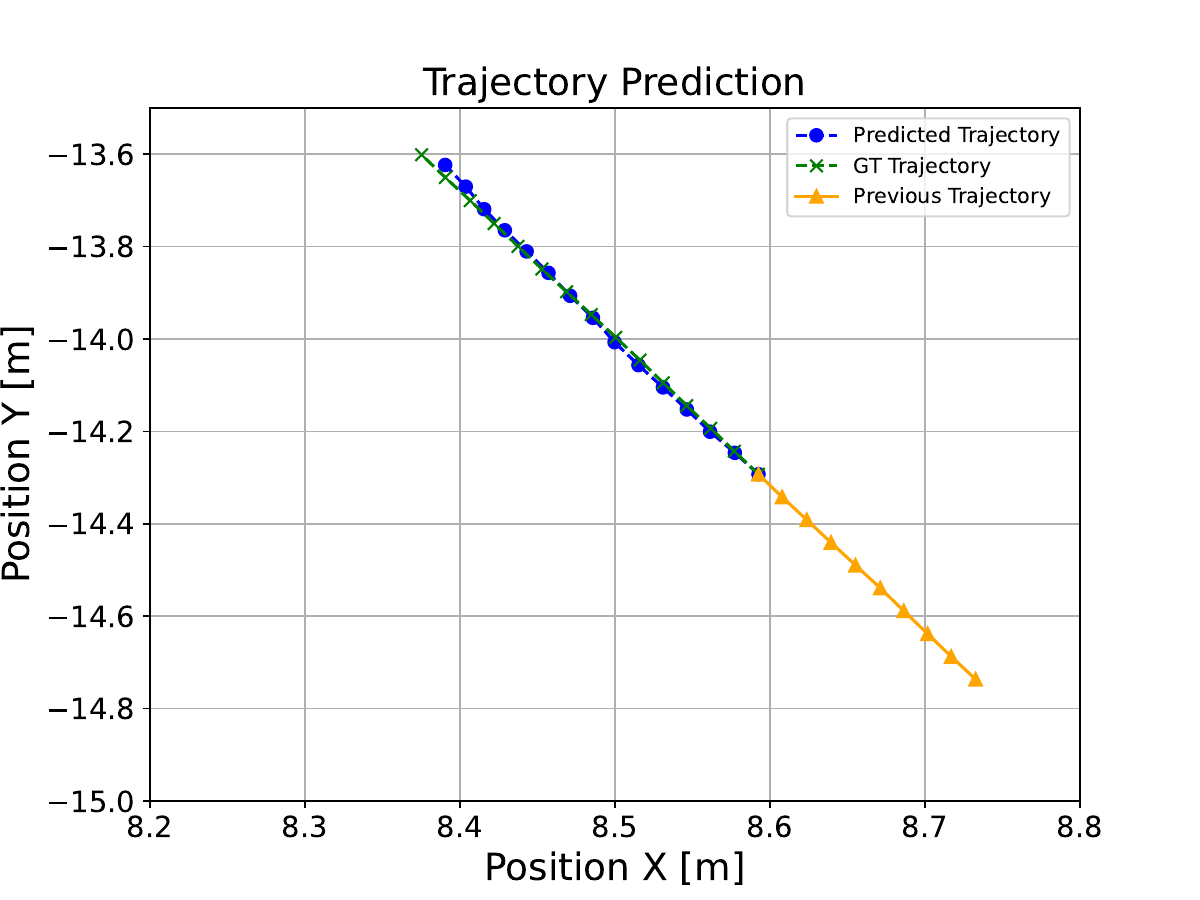}
        \end{subfigure}
    \end{minipage}
    \begin{minipage}[t]{0.49\columnwidth}
        \begin{subfigure}[b]{\textwidth}
            \includegraphics[width=\textwidth]{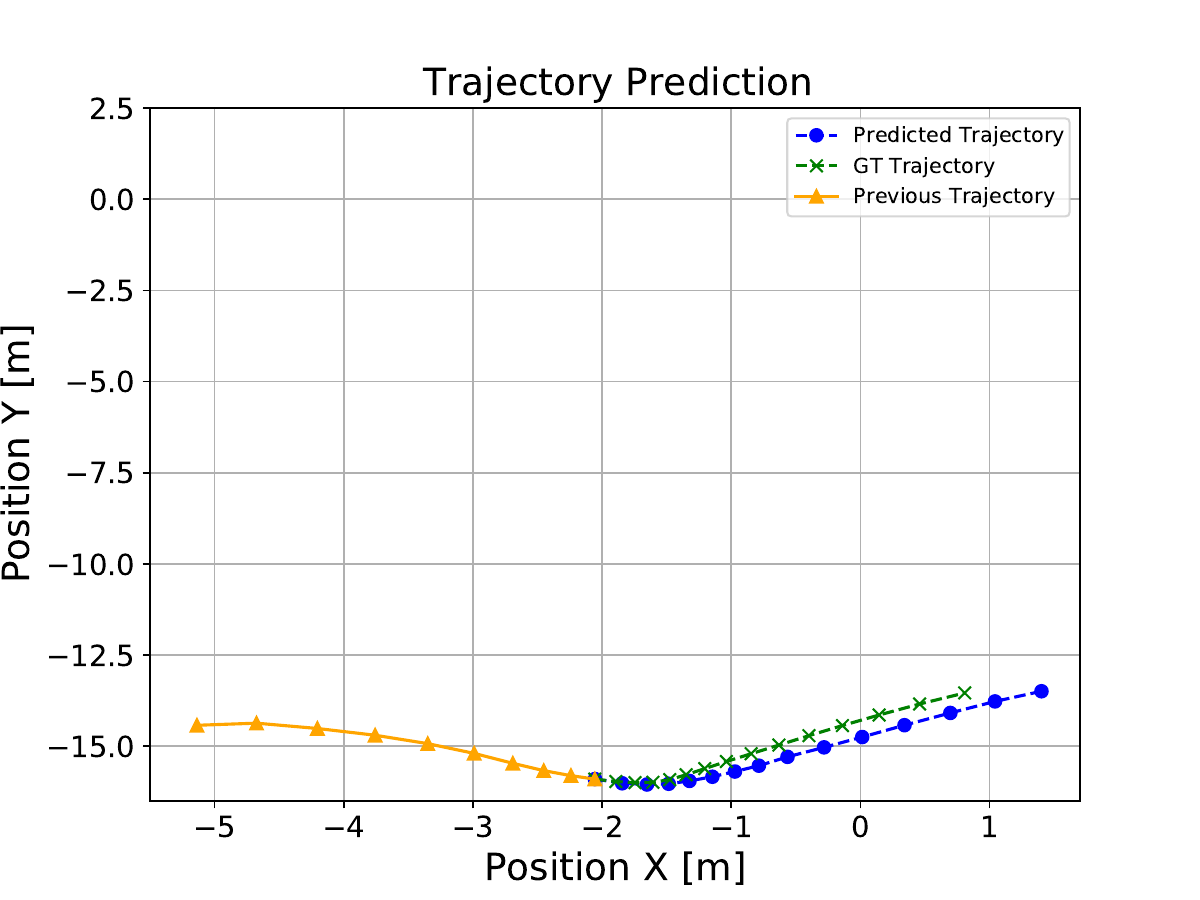}
        \end{subfigure}
    \end{minipage}
    \caption{LSTM-based trajectory prediction results. X,Y coordinate is based on meters. \textit{(left)}: simpler scenario with linear motion of a target object; \textit{(right)}: complex scenario with non-linear motion of a target object.}
    \label{fig:lstm-result}
\end{figure}

This LSTM-MLP framework processes an input of $10$ previous coordinates and forecasts a future trajectory consisting of $15$ coordinates, matching the length used in our trajectory tracking evaluation. However, the past and future trajectory length is not strictly constrained to $10, 15$, respectively. Users can adaptively choose the parameters based on the size of the environments, robots, and frequency of sensors. After normalization of an input trajectory, we used data augmentation by rotating the trajectory into every $\SI{45}{\degree}$ angle, i.e., total $8$ trajectories per one input trajectory. During training, we used $32$ batch size, $300$ epoch, drop out $0.4$, learning rate $0.0001$, Adam optimizer, and MSE loss function. The example results of the trajectory prediction are shown in \fig{lstm-result}. 

\begin{figure}[t!]
    \centering
    \begin{minipage}[t]{0.49\columnwidth}
        \begin{subfigure}[b]{\textwidth}            
            \includegraphics[width=\textwidth]{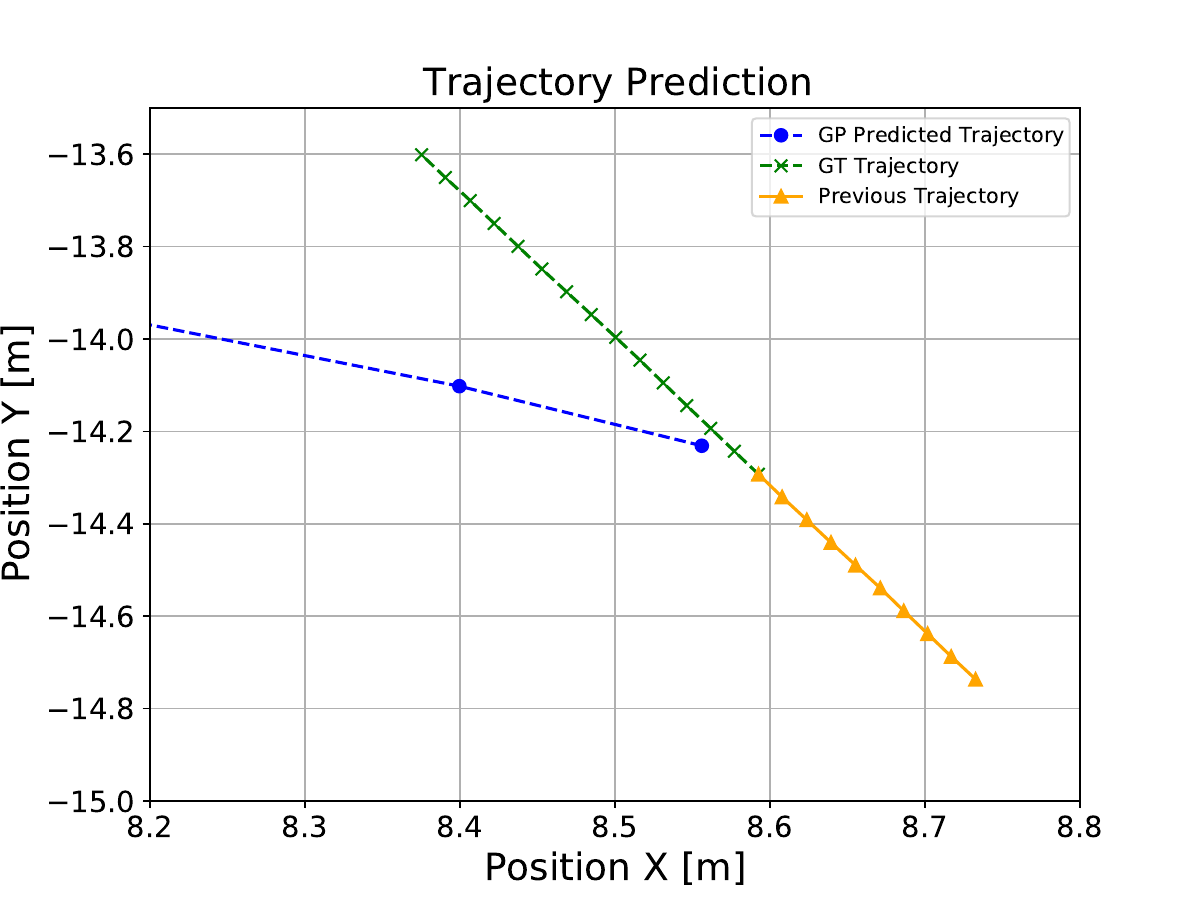}
        \end{subfigure}
    \end{minipage}
    \begin{minipage}[t]{0.49\columnwidth}
        \begin{subfigure}[b]{\textwidth}
            \includegraphics[width=\textwidth]{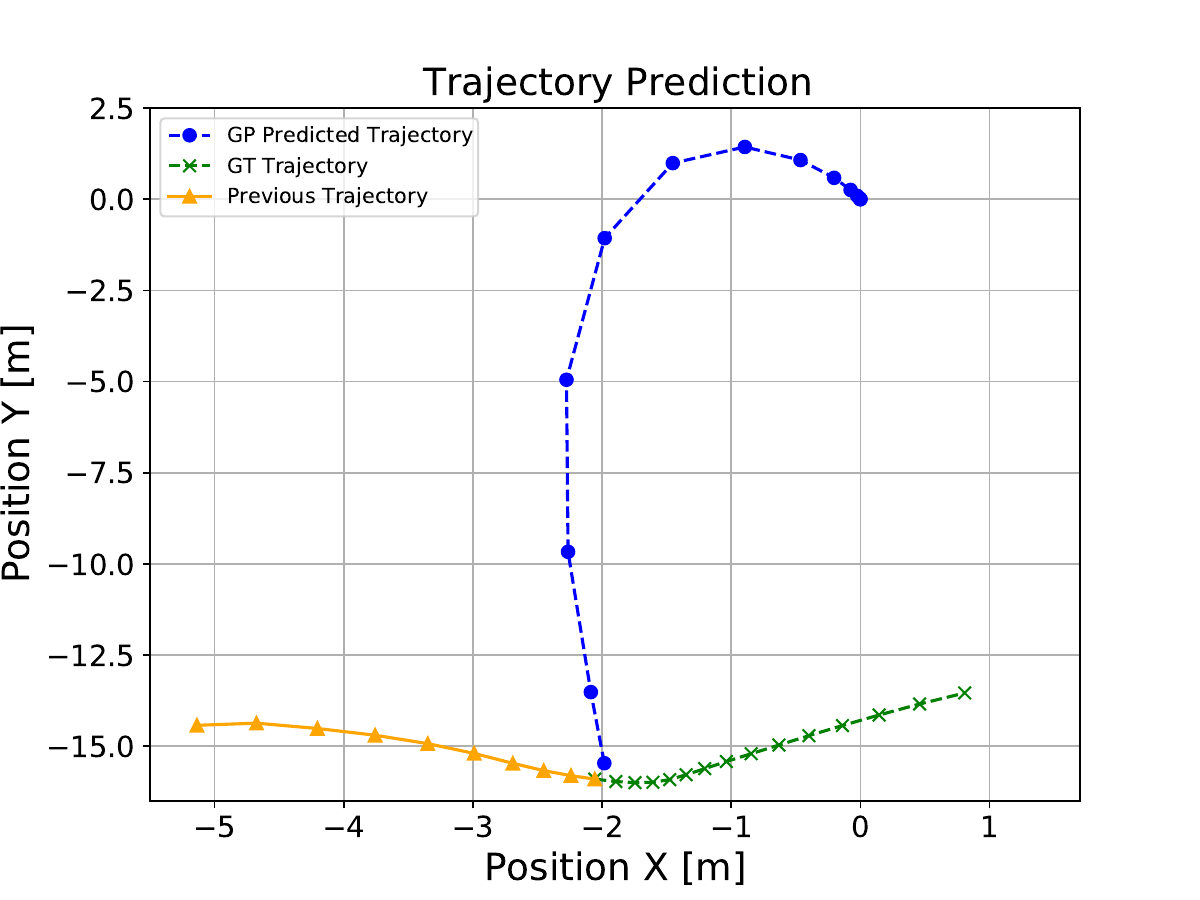}
        \end{subfigure}
    \end{minipage}
    \caption{\revisedtext{GP-based trajectory prediction results. X,Y coordinate is based on meters. \textit{(left)}: simpler scenario with linear motion of a target object; \textit{(right)}: complex scenario with non-linear motion of a target object.}}
    \label{fig:gp-result}
\end{figure}

\revisedtext{We also demonstrate the trajectory prediction results using the Gaussian Process (GP) with the Mat\'ern kernel (\fig{gp-result}). Despite being tested under the same conditions as the LSTM, the GP exhibited low performance. The limitations of the GP--(1) the need for an explicit choice of kernel and its high computational complexity; (2) the requirement for continuous training with new input data and newly detected target objects within the sensor range; and (3) the assumption of independence between $x$ and $y$ coordinates--make it not an ideal choice for real-time trajectory prediction during the search and tracking of multiple targets.}

\section{\revisedtext{Action Algorithm Details}}
\revisedtext{We include the algorithmic detail about how the HQ places bids for search to agents and assign the search task with the best utility.}

\revisedtext{\textbf{Search} (\alg{alg:search-task-assignment}): the headquarters (HQ) identifies best frontiers by \texttt{getBestFrontiers} (line 2) based on the highest expected utility (calculated by $\jexplore$ and $\jexploit$) and then starts an auction for the frontiers in order at each round (line 3--4). The HQ communicates to \textit{non-active} agents, i.e., not conducting a \emph{tracking} task at that time, by \texttt{findNonActiveAgents} (line 5). The HQ places bids among unassigned agents and collect the submitted bid result (lines 9--11). Finally, the HQ selects $\ag_{\mi{opt}}$ with the best utility by \texttt{getBestBid} (line 12) and updates the auction result (lines 13--14).
}

\begin{algorithm}[t]
\caption{\revisedtext{Search Task Assignment (HQ).}}
\label{alg:search-task-assignment}
\begin{algorithmic}[1]
\renewcommand{\algorithmicrequire}{\textbf{Input:}}
\renewcommand{\algorithmicensure}{\textbf{Output:}}
    \scriptsize
    \REQUIRE \,
    \vspace{-1em}
    \begin{itemize}
        \itemsep-0.2em
        \item Agents set $\A$, active agent set $\A_{act}$
        \item Frontiers set $F$
    \end{itemize} 
    \vspace{-0.5em}
    \ENSURE hash table $Q$ for optimal trajectory $\tau^*$ and corresponding $\ag_{\mi{opt}}$ assigned to $\tau^*$
    \IF{$F \neq \emptyset $}
        \STATE $F_{\search}$ $\gets$ $\texttt{getBestFrontiers}(F)$ \comment{priority queue}
        \WHILE{$F_{\search} \neq \emptyset$}
            \STATE $fr_{\search}$ $\gets$ $F_{\search}$\textrm{.pop()}
            \STATE $\A_{\neg \mi{act}}$ $\gets$ $\texttt{findNonActiveAgents}(\A \setminus \A_{\mi{act}})$
            \IF{$\A_{\neg \mi{act}} == \emptyset$}
                \BREAK
            \ENDIF
            \STATE $Q_{\mi{search}}$ $\gets$ \textrm{empty} \comment{hash table for searching}
            \FOR{$\ag_{\neg \mi{act}} \in \A_{\neg act}$}
                \STATE $\tau_{\ag_{\neg \mi{act}}}$ $\gets$ \texttt{requestBid}($\ag_{\neg \mi{act}}$)
                \comment{$\ag_{\neg \mi{act}}$ submits a bid for $fr_{\mi{search}}$}
                \STATE $Q_{\search} \gets \textrm{update}(Q_{\search}, \tau_{\ag_{\neg \mi{act}}}, \ag_{\neg \mi{act}})$
            \ENDFOR
            \STATE $\tau^*_{fr_{\mi{search}}}, \ag_{\mi{opt}}$ $\gets$ $\texttt{getBestBid}(Q_{\search})$
            \STATE $Q$ $\gets$ $\textrm{update}(\tau^*_{fr_{\mi{search}}}, \ag_{\mi{opt}})$
            \STATE $\A_{\mi{act}}$ $\gets$ $\A_{\mi{act}} \cup \{\ag_{\mi{opt}}\}$
        \ENDWHILE
    \ENDIF
    \RETURN $Q$ 
\end{algorithmic} 
\end{algorithm}

\section{\revisedtext{Bayesian Filtering Details}}
\revisedtext{We include the math derivation for the prediction and update steps as a Bayesian filtering on the belief of object's positions measured by an agent.}

\revisedtext{
\textbf{Prediction step:}
\begin{equation}
    \obelieftrack{\ag}{\ob}{t} = \sum_{q_\ag^{t}} \, p(\ob^t|\ob^{t-1}, q_\ag^{t}) \cdot \belieftrack{\ag}{\ob}{t-1}
    \label{eq:predict}
\end{equation}
To derive this model, we use the \textit{total probability} theorem:
\begin{align}
    & \obelieftrack{\ag}{\ob}{t} \nonumber \\
    & = p(\ob^t|\zlocag^{0:t-1}, q_\ag^{0:t}) \nonumber \\
    & = \sum_{q_\ag^{t}} p(\ob^t | \ob^{t-1}, \zlocag^{0:t-1}, q_\ag^{0:t}) \cdot p(\ob^{t-1} | \zlocag^{0:t-1}, q_\ag^{0:t}) \nonumber
\end{align}  
Given the Markov assumption, we get $p(\ob^t | \ob^{t-1}, \zlocag^{0:t-1}, q_\ag^{0:t}) =  p(\ob^t|\ob^{t-1}, q_\ag^{t})$. Also, the second term can be omitted $q_\ag^{t}$ as it does not affect $\ob^{t-1}$ directly. Hence, 
\begin{align}
    & \sum_{q_\ag^{t}} p(\ob^t | \ob^{t-1}, \zlocag^{0:t-1}, q_\ag^{0:t}) \cdot p(\ob^{t-1} | \zlocag^{0:t-1}, q_\ag^{0:t-1}) \nonumber  \\
    & = \sum_{q_\ag^{t}} \, p(\ob^t|\ob^{t-1}, q_\ag^{t}) \cdot \belieftrack{\ag}{\ob}{t-1} \nonumber
\end{align}
}

\revisedtext{
\textbf{Update step:} By using \textit{Bayes rule}, we can get the belief state updated:
\begin{equation}
\belieftrack{\ag}{\ob}{t} = \eta \,\, p(\zlocag^{t}|\ob^t) \cdot \obelieftrack{\ag}{\ob}{t}
\label{eq:update}
\end{equation} where $\belieftrack{\ag}{\ob}{t}$ is a corrected state, $\eta$ is a normalizer, $\zlocag^{t}$ is a sensor measurement of the target $\ob$ at time $t$, and $\obelieftrack{\ag}{\ob}{t}$ is a predicted state. The derivation is as follows:
\begin{align}
    & \belieftrack{\ag}{\ob}{t} \nonumber 
    = p(\ob^t|\zlocag^{0:t}, q_\ag^{0:t}) \nonumber \\
    & = \frac{p(\zlocag^{t}|\ob^t, \zlocag^{0:t-1}, q_\ag^{0:t}) \cdot p(\ob^t | \zlocag^{0:t-1}, q_\ag^{0:t})}
    {p(\zlocag^{t} | \zlocag^{0:t-1}, q_\ag^{0:t})} \nonumber \\
    & = \eta \,\, p(\zlocag^{t}|\ob^t) \cdot \obelieftrack{\ag}{\ob}{t} \nonumber
\end{align}
Using \textit{conditional independence}, $p(\zlocag^{t}|\ob^t, \zlocag^{0:t-1}, q_\ag^{0:t}) = p(\zlocag^{t}|\ob^t)$. Also, $p(\ob^t | \zlocag^{0:t-1}, q_\ag^{0:t}) = \obelieftrack{\ag}{\ob}{t}$ according to the \textit{definition of the predicted belief}.
}

\section{Experimental setup details}
\begin{figure}[t]
    \centering
    \begin{minipage}[t]{0.40\columnwidth}
        \begin{subfigure}[b]{\textwidth}
            \includegraphics[width=\textwidth]{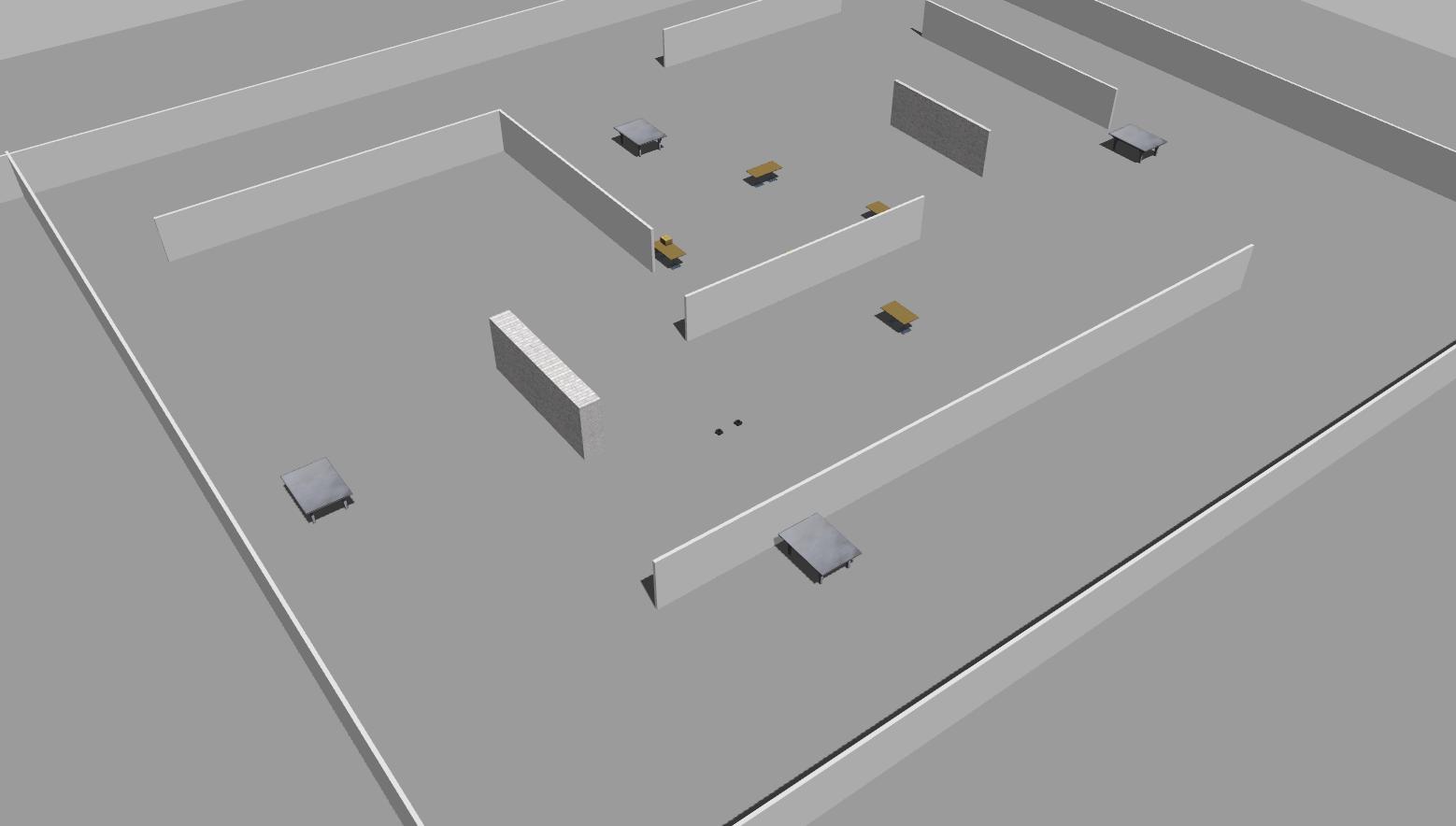}
        \end{subfigure}
    \end{minipage}
    \begin{minipage}[t]{0.42\columnwidth}
        \begin{subfigure}[b]{\textwidth}
            \includegraphics[width=\textwidth]{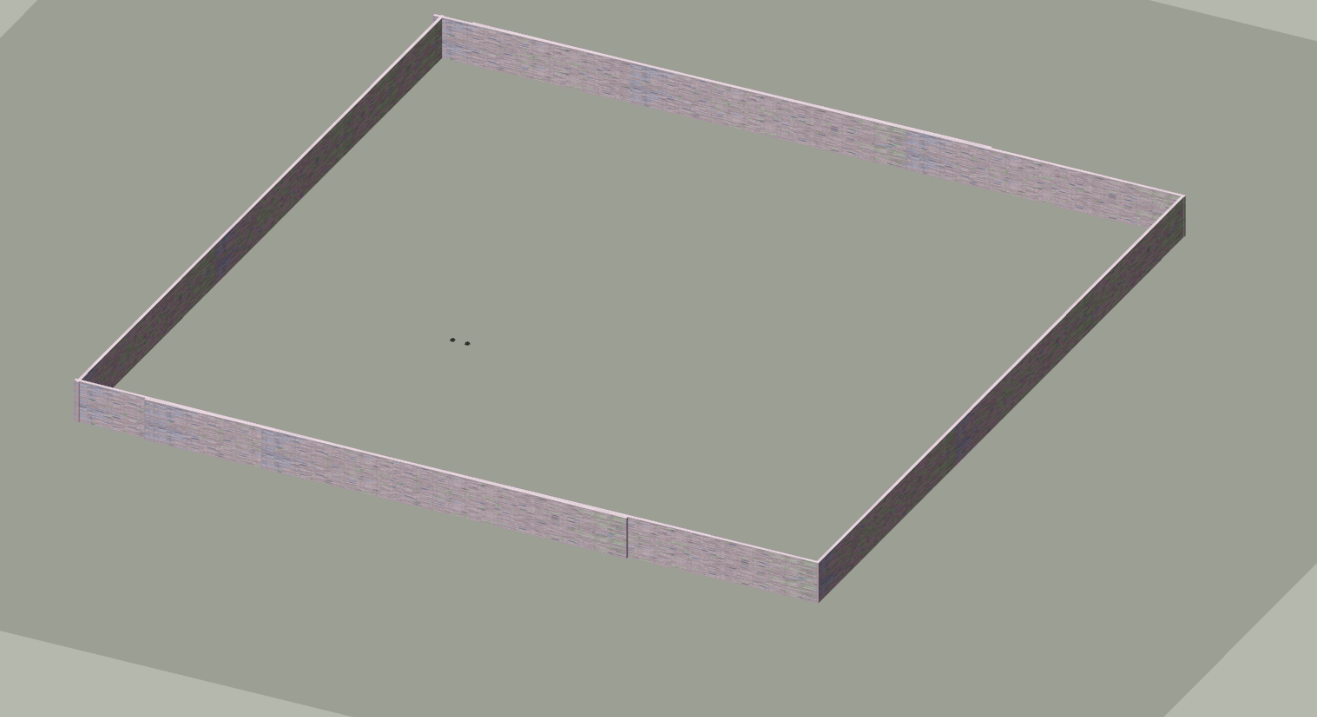}
        \end{subfigure}
    \end{minipage}
    \caption{\revisedtext{3D simulation \qtyproduct{50 x 50}{m} environments: (\textit{left}): \emph{office2} (\textit{right}): \emph{empty}}.}
    \label{fig:gazebo-env-app}
\end{figure}

The following parameters in \tab{experimental_parameters} were what we have used for the experimental configurations as discussed in the paper. Unlike the literature, for heterogeneous capabilities, we modeled varying parameters  $vr_a, \Sigma_o$ for agents, and $\Sigma_e$ for external reports. \revisedtext{For 3D simulations, we tested our method and the baselines in diverse environments (see \fig{gazebo-env-app}) in addition to \emph{office1} in the paper.}
\begin{table}[t!]
    \centering
    \begin{tabular}{|c|c|}
        \hline
        \textbf{Parameter} & \textbf{Values}  \\ \hline \hline
        \multicolumn{2}{|c|}{for simulation} \\ \hline
        $s_o$ & $\SI{0.2}{\meter/\second}$ \\
        $s_a$ & $\SI{0.4}{\meter/\second}$\\
        $vr_a$ & $\SIrange{5}{8}{\meter}$\\
        $\alpha$ & $\SIrange{0.08}{0.2}{}$\\
        $\beta$ & $0$ \\
        $\Sigma_o$ & $\SIrange{0.7}{1.3}{\meter}$\\
        $\Sigma_e$ & $\SIrange{0.5}{1.0}{\meter}$\\
        \hline
        \multicolumn{2}{|c|}{for method} \\ \hline
        $r$ & $1/90$ \\
        $w$ & $0.3$ (search), $0.2$ (track) \\
        $D_\textit{thre}$ & $\SI{3.5}{\meter}$ \\
        $\trace(\Sigma_{\mi{thre}})$  & $2.0$ \\ \hline
    \end{tabular}
    \caption{Parameters used in the experiments.}
    \label{tab:experimental_parameters}
\end{table}

\section{Supplementary material}
We included the following supplementary material:
\begin{itemize}
 \item a document with additional results,
 \item a video describing the proposed method and highlighting experiments and results.
\end{itemize}

%% file: journal-root.bbl

\begin{thebibliography}{80}
\ifx \bisbn   \undefined \def \bisbn  #1{ISBN #1}\fi
\ifx \binits  \undefined \def \binits#1{#1}\fi
\ifx \bauthor  \undefined \def \bauthor#1{#1}\fi
\ifx \batitle  \undefined \def \batitle#1{#1}\fi
\ifx \bjtitle  \undefined \def \bjtitle#1{#1}\fi
\ifx \bvolume  \undefined \def \bvolume#1{\textbf{#1}}\fi
\ifx \byear  \undefined \def \byear#1{#1}\fi
\ifx \bissue  \undefined \def \bissue#1{#1}\fi
\ifx \bfpage  \undefined \def \bfpage#1{#1}\fi
\ifx \blpage  \undefined \def \blpage #1{#1}\fi
\ifx \burl  \undefined \def \burl#1{\textsf{#1}}\fi
\ifx \doiurl  \undefined \def \doiurl#1{\url{https://doi.org/#1}}\fi
\ifx \betal  \undefined \def \betal{\textit{et al.}}\fi
\ifx \binstitute  \undefined \def \binstitute#1{#1}\fi
\ifx \binstitutionaled  \undefined \def \binstitutionaled#1{#1}\fi
\ifx \bctitle  \undefined \def \bctitle#1{#1}\fi
\ifx \beditor  \undefined \def \beditor#1{#1}\fi
\ifx \bpublisher  \undefined \def \bpublisher#1{#1}\fi
\ifx \bbtitle  \undefined \def \bbtitle#1{#1}\fi
\ifx \bedition  \undefined \def \bedition#1{#1}\fi
\ifx \bseriesno  \undefined \def \bseriesno#1{#1}\fi
\ifx \blocation  \undefined \def \blocation#1{#1}\fi
\ifx \bsertitle  \undefined \def \bsertitle#1{#1}\fi
\ifx \bsnm \undefined \def \bsnm#1{#1}\fi
\ifx \bsuffix \undefined \def \bsuffix#1{#1}\fi
\ifx \bparticle \undefined \def \bparticle#1{#1}\fi
\ifx \barticle \undefined \def \barticle#1{#1}\fi
\bibcommenthead
\ifx \bconfdate \undefined \def \bconfdate #1{#1}\fi
\ifx \botherref \undefined \def \botherref #1{#1}\fi
\ifx \url \undefined \def \url#1{\textsf{#1}}\fi
\ifx \bchapter \undefined \def \bchapter#1{#1}\fi
\ifx \bbook \undefined \def \bbook#1{#1}\fi
\ifx \bcomment \undefined \def \bcomment#1{#1}\fi
\ifx \oauthor \undefined \def \oauthor#1{#1}\fi
\ifx \citeauthoryear \undefined \def \citeauthoryear#1{#1}\fi
\ifx \endbibitem  \undefined \def \endbibitem {}\fi
\ifx \bconflocation  \undefined \def \bconflocation#1{#1}\fi
\ifx \arxivurl  \undefined \def \arxivurl#1{\textsf{#1}}\fi
\csname PreBibitemsHook\endcsname

\bibitem[\protect\citeauthoryear{Khan et~al.}{2018}]{review-csat2018}
\begin{barticle}
\bauthor{\bsnm{Khan}, \binits{A.}},
\bauthor{\bsnm{Rinner}, \binits{B.}},
\bauthor{\bsnm{Cavallaro}, \binits{A.}}:
\batitle{Cooperative robots to observe moving targets: Review}.
\bjtitle{IEEE Transactions on Cybernetics}
\bvolume{48}(\bissue{1}),
\bfpage{187}--\blpage{198}
(\byear{2018})
\end{barticle}
\endbibitem

\bibitem[\protect\citeauthoryear{Robin and
  Lacroix}{2016}]{taxonomy-review-2016}
\begin{barticle}
\bauthor{\bsnm{Robin}, \binits{C.}},
\bauthor{\bsnm{Lacroix}, \binits{S.}}:
\batitle{Multi-robot target detection and tracking: taxonomy and survey}.
\bjtitle{Autonomous Robots}
\bvolume{40}(\bissue{4}),
\bfpage{729}--\blpage{760}
(\byear{2016})
\end{barticle}
\endbibitem

\bibitem[\protect\citeauthoryear{Charrow et~al.}{2015}]{upenn-balance-2015}
\begin{bchapter}
\bauthor{\bsnm{Charrow}, \binits{B.}},
\bauthor{\bsnm{Michael}, \binits{N.}},
\bauthor{\bsnm{Kumar}, \binits{V.}}:
\bctitle{Active control strategies for discovering and localizing devices with
  range-only sensors}.
In: \beditor{\bsnm{Akin}, \binits{H.L.}},
\beditor{\bsnm{Amato}, \binits{N.M.}},
\beditor{\bsnm{Isler}, \binits{V.}},
\beditor{\bsnm{Van Der~Stappen}, \binits{A.F.}} (eds.)
\bbtitle{Algorithmic Foundations of Robotics XI (WAFR)}
vol. \bseriesno{107},
pp. \bfpage{55}--\blpage{71}
(\byear{2015}).
\burl{https://link.springer.com/chapter/10.1007/978-3-319-16595-0_4}
\end{bchapter}
\endbibitem

\bibitem[\protect\citeauthoryear{Frew and
  Elston}{2008}]{target-assign-csat-2008}
\begin{bchapter}
\bauthor{\bsnm{Frew}, \binits{E.W.}},
\bauthor{\bsnm{Elston}, \binits{J.}}:
\bctitle{Target assignment for integrated search and tracking by active robot
  networks}.
In: \bbtitle{IEEE International Conference on Robotics and Automation (ICRA)},
pp. \bfpage{2354}--\blpage{2359}
(\byear{2008}).
\burl{https://ieeexplore.ieee.org/abstract/document/4543565/}
\end{bchapter}
\endbibitem

\bibitem[\protect\citeauthoryear{Dames et~al.}{2017}]{vijay-sat-2017}
\begin{barticle}
\bauthor{\bsnm{Dames}, \binits{P.}},
\bauthor{\bsnm{Tokekar}, \binits{P.}},
\bauthor{\bsnm{Kumar}, \binits{V.}}:
\batitle{Detecting, localizing, and tracking an unknown number of moving
  targets using a team of mobile robots}.
\bjtitle{The International Journal of Robotics Research}
\bvolume{36}(\bissue{13–14}),
\bfpage{1540}--\blpage{1553}
(\byear{2017})
\end{barticle}
\endbibitem

\bibitem[\protect\citeauthoryear{How
  et~al.}{2009}]{increasing-autonomy-csat-2009}
\begin{barticle}
\bauthor{\bsnm{How}, \binits{J.P.}},
\bauthor{\bsnm{Fraser}, \binits{C.}},
\bauthor{\bsnm{Kulling}, \binits{K.C.}},
\bauthor{\bsnm{Bertuccelli}, \binits{L.F.}},
\bauthor{\bsnm{Toupet}, \binits{O.}},
\bauthor{\bsnm{Brunet}, \binits{L.}},
\bauthor{\bsnm{Bachrach}, \binits{A.}},
\bauthor{\bsnm{Roy}, \binits{N.}}:
\batitle{Increasing autonomy of {UAVs}}.
\bjtitle{IEEE Robotics \& Automation Magazine}
\bvolume{16}(\bissue{2}),
\bfpage{43}--\blpage{51}
(\byear{2009})
\end{barticle}
\endbibitem

\bibitem[\protect\citeauthoryear{Furukawa
  et~al.}{2006}]{recursive-bayesian-csat-2006}
\begin{bchapter}
\bauthor{\bsnm{Furukawa}, \binits{T.}},
\bauthor{\bsnm{Bourgault}, \binits{F.}},
\bauthor{\bsnm{Lavis}, \binits{B.}},
\bauthor{\bsnm{Durrant-Whyte}, \binits{H.F.}}:
\bctitle{Recursive bayesian search-and-tracking using coordinated {UAVs} for
  lost targets}.
In: \bbtitle{IEEE International Conference on Robotics and Automation (ICRA)},
pp. \bfpage{2521}--\blpage{2526}
(\byear{2006}).
\burl{https://ieeexplore.ieee.org/abstract/document/1642081/}
\end{bchapter}
\endbibitem

\bibitem[\protect\citeauthoryear{Kwa et~al.}{2020}]{optimal-swarm-2020}
\begin{bchapter}
\bauthor{\bsnm{Kwa}, \binits{H.L.}},
\bauthor{\bsnm{Kit}, \binits{J.L.}},
\bauthor{\bsnm{Bouffanais}, \binits{R.}}:
\bctitle{Optimal swarm strategy for dynamic target search and tracking}.
In: \bbtitle{International Conference on Autonomous Agents and MultiAgent
  Systems (AAMAS)},
pp. \bfpage{672}--\blpage{680}
(\byear{2020}).
\burl{https://www.ifaamas.org/Proceedings/aamas2020/pdfs/p672.pdf}
\end{bchapter}
\endbibitem

\bibitem[\protect\citeauthoryear{OV et~al.}{2020}]{ov2020impact}
\begin{bchapter}
\bauthor{\bsnm{OV}, \binits{S.S.}},
\bauthor{\bsnm{Parasuraman}, \binits{R.}},
\bauthor{\bsnm{Pidaparti}, \binits{R.}}:
\bctitle{Impact of heterogeneity in multi-robot systems on collective behaviors
  studied using a search and rescue problem}.
In: \bbtitle{IEEE International Symposium on Safety, Security, and Rescue
  Robotics (SSRR)},
pp. \bfpage{290}--\blpage{297}
(\byear{2020}).
\bcomment{IEEE}
\end{bchapter}
\endbibitem

\bibitem[\protect\citeauthoryear{Papaioannou
  et~al.}{2021}]{search-track-cyprus-2021-journal}
\begin{barticle}
\bauthor{\bsnm{Papaioannou}, \binits{S.}},
\bauthor{\bsnm{Kolios}, \binits{P.}},
\bauthor{\bsnm{Theocharides}, \binits{T.}},
\bauthor{\bsnm{Panayiotou}, \binits{C.G.}},
\bauthor{\bsnm{Polycarpou}, \binits{M.M.}}:
\batitle{A cooperative multiagent probabilistic framework for search and track
  missions}.
\bjtitle{IEEE Transactions on Control of Network Systems}
\bvolume{8}(\bissue{2}),
\bfpage{847}--\blpage{858}
(\byear{2021})
\end{barticle}
\endbibitem

\bibitem[\protect\citeauthoryear{Zhang et~al.}{2021}]{harvard-source-seek-2021}
\begin{bchapter}
\bauthor{\bsnm{Zhang}, \binits{T.}},
\bauthor{\bsnm{Qin}, \binits{V.}},
\bauthor{\bsnm{Tang}, \binits{Y.}},
\bauthor{\bsnm{Li}, \binits{N.}}:
\bctitle{Source seeking by dynamic source location estimation}.
In: \bbtitle{IEEE/RSJ International Conference on Intelligent Robots and
  Systems (IROS)},
pp. \bfpage{2598}--\blpage{2605}
(\byear{2021}).
\burl{https://ieeexplore.ieee.org/abstract/document/9636841/}
\end{bchapter}
\endbibitem

\bibitem[\protect\citeauthoryear{Meghjani et~al.}{2016}]{malika-search-2016}
\begin{bchapter}
\bauthor{\bsnm{Meghjani}, \binits{M.}},
\bauthor{\bsnm{Manjanna}, \binits{S.}},
\bauthor{\bsnm{Dudek}, \binits{G.}}:
\bctitle{Multi-target search strategies}.
In: \bbtitle{IEEE International Symposium on Safety, Security, and Rescue
  Robotics (SSRR)},
\bconflocation{Lausanne, Switzerland},
pp. \bfpage{328}--\blpage{333}
(\byear{2016}).
\burl{https://ieeexplore.ieee.org/abstract/document/7784323/}
\end{bchapter}
\endbibitem

\bibitem[\protect\citeauthoryear{Zhu et~al.}{2019}]{gas-mapping-2019}
\begin{botherref}
\oauthor{\bsnm{Zhu}, \binits{P.}},
\oauthor{\bsnm{Ferrari}, \binits{S.}},
\oauthor{\bsnm{Morelli}, \binits{J.}},
\oauthor{\bsnm{Linares}, \binits{R.}},
\oauthor{\bsnm{Doerr}, \binits{B.}}:
Scalable gas sensing, mapping, and path planning via decentralized hilbert
  maps.
Sensors
\textbf{19}(7)
(2019)
\end{botherref}
\endbibitem

\bibitem[\protect\citeauthoryear{LeGrand et~al.}{2023}]{cell-mb-2023}
\begin{barticle}
\bauthor{\bsnm{LeGrand}, \binits{K.A.}},
\bauthor{\bsnm{Zhu}, \binits{P.}},
\bauthor{\bsnm{Ferrari}, \binits{S.}}:
\batitle{Cell multi-bernoulli (cell-mb) sensor control for multi-object
  search-while-tracking (swt)}.
\bjtitle{IEEE Transactions on Pattern Analysis and Machine Intelligence}
\bvolume{45}(\bissue{6}),
\bfpage{7195}--\blpage{7207}
(\byear{2023})
\end{barticle}
\endbibitem

\bibitem[\protect\citeauthoryear{Wei et~al.}{2019}]{pan-tilt-camera-2019}
\begin{barticle}
\bauthor{\bsnm{Wei}, \binits{H.}},
\bauthor{\bsnm{Zhu}, \binits{P.}},
\bauthor{\bsnm{Liu}, \binits{M.}},
\bauthor{\bsnm{How}, \binits{J.P.}},
\bauthor{\bsnm{Ferrari}, \binits{S.}}:
\batitle{Automatic pan–tilt camera control for learning dirichlet process
  gaussian process (dpgp) mixture models of multiple moving targets}.
\bjtitle{IEEE Transactions on Automatic Control}
\bvolume{64}(\bissue{1}),
\bfpage{159}--\blpage{173}
(\byear{2019})
\end{barticle}
\endbibitem

\bibitem[\protect\citeauthoryear{Foderaro
  et~al.}{2018}]{distributed-control-silvia-2018}
\begin{barticle}
\bauthor{\bsnm{Foderaro}, \binits{G.}},
\bauthor{\bsnm{Zhu}, \binits{P.}},
\bauthor{\bsnm{Wei}, \binits{H.}},
\bauthor{\bsnm{Wettergren}, \binits{T.A.}},
\bauthor{\bsnm{Ferrari}, \binits{S.}}:
\batitle{Distributed optimal control of sensor networks for dynamic target
  tracking}.
\bjtitle{IEEE Transactions on Control of Network Systems}
\bvolume{5}(\bissue{1}),
\bfpage{142}--\blpage{153}
(\byear{2018})
\end{barticle}
\endbibitem

\bibitem[\protect\citeauthoryear{Sun et~al.}{2023}]{usv-pursuit-evasion-2023}
\begin{barticle}
\bauthor{\bsnm{Sun}, \binits{Z.}},
\bauthor{\bsnm{Sun}, \binits{H.}},
\bauthor{\bsnm{Li}, \binits{P.}},
\bauthor{\bsnm{Zou}, \binits{J.}}:
\batitle{Cooperative strategy for pursuit-evasion problem in the presence of
  static and dynamic obstacles}.
\bjtitle{Ocean Engineering}
\bvolume{279},
\bfpage{114476}
(\byear{2023})
\end{barticle}
\endbibitem

\bibitem[\protect\citeauthoryear{Galceran and
  Carreras}{2013}]{galceran2013survey}
\begin{barticle}
\bauthor{\bsnm{Galceran}, \binits{E.}},
\bauthor{\bsnm{Carreras}, \binits{M.}}:
\batitle{A survey on coverage path planning for robotics}.
\bjtitle{Robotics and Autonomous Systems}
\bvolume{61}(\bissue{12}),
\bfpage{1258}--\blpage{1276}
(\byear{2013})
\end{barticle}
\endbibitem

\bibitem[\protect\citeauthoryear{Quattrini~Li}{2020}]{quattrini2020exploration}
\begin{barticle}
\bauthor{\bsnm{Quattrini~Li}, \binits{A.}}:
\batitle{Exploration and mapping with groups of robots: Recent trends}.
\bjtitle{Current Robotics Reports}
\bvolume{1},
\bfpage{227}--\blpage{237}
(\byear{2020})
\end{barticle}
\endbibitem

\bibitem[\protect\citeauthoryear{Drew}{2021}]{drew2021multi}
\begin{barticle}
\bauthor{\bsnm{Drew}, \binits{D.S.}}:
\batitle{Multi-agent systems for search and rescue applications}.
\bjtitle{Current Robotics Reports}
\bvolume{2},
\bfpage{189}--\blpage{200}
(\byear{2021})
\end{barticle}
\endbibitem

\bibitem[\protect\citeauthoryear{Bai et~al.}{2021}]{bai2021information}
\begin{barticle}
\bauthor{\bsnm{Bai}, \binits{S.}},
\bauthor{\bsnm{Shan}, \binits{T.}},
\bauthor{\bsnm{Chen}, \binits{F.}},
\bauthor{\bsnm{Liu}, \binits{L.}},
\bauthor{\bsnm{Englot}, \binits{B.}}:
\batitle{Information-driven path planning}.
\bjtitle{Current Robotics Reports}
\bvolume{2}(\bissue{2}),
\bfpage{177}--\blpage{188}
(\byear{2021})
\end{barticle}
\endbibitem

\bibitem[\protect\citeauthoryear{Hajieghrary
  et~al.}{2017}]{hajieghrary2017information}
\begin{barticle}
\bauthor{\bsnm{Hajieghrary}, \binits{H.}},
\bauthor{\bsnm{Mox}, \binits{D.}},
\bauthor{\bsnm{Hsieh}, \binits{M.A.}}:
\batitle{Information theoretic source seeking strategies for multiagent plume
  tracking in turbulent fields}.
\bjtitle{Journal of Marine Science and Engineering}
\bvolume{5}(\bissue{1}),
\bfpage{3}
(\byear{2017})
\end{barticle}
\endbibitem

\bibitem[\protect\citeauthoryear{Marjovi et~al.}{2010}]{marjovi2010multi}
\begin{bchapter}
\bauthor{\bsnm{Marjovi}, \binits{A.}},
\bauthor{\bsnm{Nunes}, \binits{J.G.}},
\bauthor{\bsnm{Marques}, \binits{L.}},
\bauthor{\bsnm{Almeida}, \binits{A.}}:
\bctitle{Multi-robot fire searching in unknown environment}.
In: \bbtitle{Field and Service Robotics: Results of the 7th International
  Conference},
pp. \bfpage{341}--\blpage{351}
(\byear{2010}).
\bcomment{Springer}
\end{bchapter}
\endbibitem

\bibitem[\protect\citeauthoryear{Rou{\v{c}}ek et~al.}{2020}]{rouvcek2020darpa}
\begin{bchapter}
\bauthor{\bsnm{Rou{\v{c}}ek}, \binits{T.}},
\bauthor{\bsnm{Pecka}, \binits{M.}},
\bauthor{\bsnm{{\v{C}}{\'\i}{\v{z}}ek}, \binits{P.}},
\bauthor{\bsnm{Pet{\v{r}}{\'\i}{\v{c}}ek}, \binits{T.}},
\bauthor{\bsnm{Bayer}, \binits{J.}},
\bauthor{\bsnm{{\v{S}}alansk{\`y}}, \binits{V.}},
\bauthor{\bsnm{He{\v{r}}t}, \binits{D.}},
\bauthor{\bsnm{Petrl{\'\i}k}, \binits{M.}},
\bauthor{\bsnm{B{\'a}{\v{c}}a}, \binits{T.}},
\bauthor{\bsnm{Spurn{\`y}}, \binits{V.}}, \betal:
\bctitle{Darpa subterranean challenge: Multi-robotic exploration of underground
  environments}.
In: \bbtitle{Modelling and Simulation for Autonomous Systems: 6th International
  Conference, MESAS 2019, Palermo, Italy, October 29--31, 2019, Revised
  Selected Papers 6},
pp. \bfpage{274}--\blpage{290}
(\byear{2020}).
\bcomment{Springer}
\end{bchapter}
\endbibitem

\bibitem[\protect\citeauthoryear{Yamauchi}{1998}]{yamauchi1998frontier}
\begin{bchapter}
\bauthor{\bsnm{Yamauchi}, \binits{B.}}:
\bctitle{Frontier-based exploration using multiple robots}.
In: \bbtitle{International Conference on Autonomous Agents},
pp. \bfpage{47}--\blpage{53}
(\byear{1998}).
\burl{https://dl.acm.org/doi/pdf/10.1145/280765.280773}
\end{bchapter}
\endbibitem

\bibitem[\protect\citeauthoryear{Burgard et~al.}{2005}]{burgard2005coordinated}
\begin{barticle}
\bauthor{\bsnm{Burgard}, \binits{W.}},
\bauthor{\bsnm{Moors}, \binits{M.}},
\bauthor{\bsnm{Stachniss}, \binits{C.}},
\bauthor{\bsnm{Schneider}, \binits{F.E.}}:
\batitle{Coordinated multi-robot exploration}.
\bjtitle{IEEE Transactions on robotics}
\bvolume{21}(\bissue{3}),
\bfpage{376}--\blpage{386}
(\byear{2005})
\end{barticle}
\endbibitem

\bibitem[\protect\citeauthoryear{Basilico and
  Amigoni}{2011}]{basilico2011exploration}
\begin{barticle}
\bauthor{\bsnm{Basilico}, \binits{N.}},
\bauthor{\bsnm{Amigoni}, \binits{F.}}:
\batitle{Exploration strategies based on multi-criteria decision making for
  searching environments in rescue operations}.
\bjtitle{Autonomous Robots}
\bvolume{31},
\bfpage{401}--\blpage{417}
(\byear{2011})
\end{barticle}
\endbibitem

\bibitem[\protect\citeauthoryear{Otte et~al.}{2018}]{otte2018competitive}
\begin{barticle}
\bauthor{\bsnm{Otte}, \binits{M.}},
\bauthor{\bsnm{Kuhlman}, \binits{M.}},
\bauthor{\bsnm{Sofge}, \binits{D.}}:
\batitle{Competitive target search with multi-agent teams: symmetric and
  asymmetric communication constraints}.
\bjtitle{Autonomous Robots}
\bvolume{42},
\bfpage{1207}--\blpage{1230}
(\byear{2018})
\end{barticle}
\endbibitem

\bibitem[\protect\citeauthoryear{Chung et~al.}{2011}]{chung2011search}
\begin{barticle}
\bauthor{\bsnm{Chung}, \binits{T.H.}},
\bauthor{\bsnm{Hollinger}, \binits{G.A.}},
\bauthor{\bsnm{Isler}, \binits{V.}}:
\batitle{Search and pursuit-evasion in mobile robotics: A survey}.
\bjtitle{Autonomous robots}
\bvolume{31},
\bfpage{299}--\blpage{316}
(\byear{2011})
\end{barticle}
\endbibitem

\bibitem[\protect\citeauthoryear{Isler et~al.}{2006}]{isler2006randomized}
\begin{barticle}
\bauthor{\bsnm{Isler}, \binits{V.}},
\bauthor{\bsnm{Kannan}, \binits{S.}},
\bauthor{\bsnm{Khanna}, \binits{S.}}:
\batitle{Randomized pursuit-evasion with local visibility}.
\bjtitle{SIAM Journal on Discrete Mathematics}
\bvolume{20}(\bissue{1}),
\bfpage{26}--\blpage{41}
(\byear{2006})
\end{barticle}
\endbibitem

\bibitem[\protect\citeauthoryear{Borie et~al.}{2011}]{borie2011algorithms}
\begin{barticle}
\bauthor{\bsnm{Borie}, \binits{R.}},
\bauthor{\bsnm{Tovey}, \binits{C.}},
\bauthor{\bsnm{Koenig}, \binits{S.}}:
\batitle{Algorithms and complexity results for graph-based pursuit evasion}.
\bjtitle{Autonomous Robots}
\bvolume{31},
\bfpage{317}--\blpage{332}
(\byear{2011})
\end{barticle}
\endbibitem

\bibitem[\protect\citeauthoryear{Kehagias et~al.}{2009}]{kehagias2009graph}
\begin{barticle}
\bauthor{\bsnm{Kehagias}, \binits{A.}},
\bauthor{\bsnm{Hollinger}, \binits{G.}},
\bauthor{\bsnm{Singh}, \binits{S.}}:
\batitle{A graph search algorithm for indoor pursuit/evasion}.
\bjtitle{Mathematical and Computer Modelling}
\bvolume{50}(\bissue{9-10}),
\bfpage{1305}--\blpage{1317}
(\byear{2009})
\end{barticle}
\endbibitem

\bibitem[\protect\citeauthoryear{Isler et~al.}{2005}]{isler2005randomized}
\begin{barticle}
\bauthor{\bsnm{Isler}, \binits{V.}},
\bauthor{\bsnm{Kannan}, \binits{S.}},
\bauthor{\bsnm{Khanna}, \binits{S.}}:
\batitle{Randomized pursuit-evasion in a polygonal environment}.
\bjtitle{IEEE Transactions on Robotics}
\bvolume{21}(\bissue{5}),
\bfpage{875}--\blpage{884}
(\byear{2005})
\end{barticle}
\endbibitem

\bibitem[\protect\citeauthoryear{Guibas et~al.}{1999}]{guibas1999visibility}
\begin{barticle}
\bauthor{\bsnm{Guibas}, \binits{L.J.}},
\bauthor{\bsnm{Latombe}, \binits{J.-C.}},
\bauthor{\bsnm{LaValle}, \binits{S.M.}},
\bauthor{\bsnm{Lin}, \binits{D.}},
\bauthor{\bsnm{Motwani}, \binits{R.}}:
\batitle{A visibility-based pursuit-evasion problem}.
\bjtitle{International Journal of Computational Geometry \& Applications}
\bvolume{9}(\bissue{04n05}),
\bfpage{471}--\blpage{493}
(\byear{1999})
\end{barticle}
\endbibitem

\bibitem[\protect\citeauthoryear{Quattrini~Li
  et~al.}{2018}]{quattrini2018search}
\begin{bchapter}
\bauthor{\bsnm{Quattrini~Li}, \binits{A.}},
\bauthor{\bsnm{Fioratto}, \binits{R.}},
\bauthor{\bsnm{Amigoni}, \binits{F.}},
\bauthor{\bsnm{Isler}, \binits{V.}}:
\bctitle{A search-based approach to solve pursuit-evasion games with limited
  visibility in polygonal environments}.
In: \bbtitle{International Conference on Autonomous Agents and MultiAgent
  Systems (AAMAS)},
pp. \bfpage{1693}--\blpage{1701}
(\byear{2018}).
\burl{https://ifaamas.org/Proceedings/aamas2018/pdfs/p1693.pdf}
\end{bchapter}
\endbibitem

\bibitem[\protect\citeauthoryear{Stiffler and
  O’Kane}{2017}]{stiffler2017complete}
\begin{barticle}
\bauthor{\bsnm{Stiffler}, \binits{N.M.}},
\bauthor{\bsnm{O’Kane}, \binits{J.M.}}:
\batitle{Complete and optimal visibility-based pursuit-evasion}.
\bjtitle{The International Journal of Robotics Research}
\bvolume{36}(\bissue{8}),
\bfpage{923}--\blpage{946}
(\byear{2017})
\end{barticle}
\endbibitem

\bibitem[\protect\citeauthoryear{Li et~al.}{2021}]{li2021bayesian}
\begin{barticle}
\bauthor{\bsnm{Li}, \binits{Y.}},
\bauthor{\bsnm{Liu}, \binits{T.}},
\bauthor{\bsnm{Zhou}, \binits{E.}},
\bauthor{\bsnm{Zhang}, \binits{F.}}:
\batitle{Bayesian learning model predictive control for process-aware source
  seeking}.
\bjtitle{IEEE Control Systems Letters}
\bvolume{6},
\bfpage{692}--\blpage{697}
(\byear{2021})
\end{barticle}
\endbibitem

\bibitem[\protect\citeauthoryear{Chung and Burdick}{2011}]{chung2011analysis}
\begin{barticle}
\bauthor{\bsnm{Chung}, \binits{T.H.}},
\bauthor{\bsnm{Burdick}, \binits{J.W.}}:
\batitle{Analysis of search decision making using probabilistic search
  strategies}.
\bjtitle{IEEE Transactions on Robotics}
\bvolume{28}(\bissue{1}),
\bfpage{132}--\blpage{144}
(\byear{2011})
\end{barticle}
\endbibitem

\bibitem[\protect\citeauthoryear{Lau et~al.}{2006}]{lau2006probabilistic}
\begin{bchapter}
\bauthor{\bsnm{Lau}, \binits{H.}},
\bauthor{\bsnm{Huang}, \binits{S.}},
\bauthor{\bsnm{Dissanayake}, \binits{G.}}:
\bctitle{Probabilistic search for a moving target in an indoor environment}.
In: \bbtitle{IEEE/RSJ International Conference on Intelligent Robots and
  Systems (IROS)},
pp. \bfpage{3393}--\blpage{3398}
(\byear{2006}).
\burl{https://ieeexplore.ieee.org/abstract/document/4058925/}
\end{bchapter}
\endbibitem

\bibitem[\protect\citeauthoryear{Matzliach
  et~al.}{2020}]{matzliach2020cooperative}
\begin{barticle}
\bauthor{\bsnm{Matzliach}, \binits{B.}},
\bauthor{\bsnm{Ben-Gal}, \binits{I.}},
\bauthor{\bsnm{Kagan}, \binits{E.}}:
\batitle{Cooperative detection of multiple targets by the group of mobile
  agents}.
\bjtitle{Entropy}
\bvolume{22}(\bissue{5}),
\bfpage{512}
(\byear{2020})
\end{barticle}
\endbibitem

\bibitem[\protect\citeauthoryear{Basilico}{2022}]{basilico2022recent}
\begin{barticle}
\bauthor{\bsnm{Basilico}, \binits{N.}}:
\batitle{Recent trends in robotic patrolling}.
\bjtitle{Current Robotics Reports}
\bvolume{3}(\bissue{2}),
\bfpage{65}--\blpage{76}
(\byear{2022})
\end{barticle}
\endbibitem

\bibitem[\protect\citeauthoryear{Senanayake
  et~al.}{2016}]{sat-swarm-review-2016}
\begin{barticle}
\bauthor{\bsnm{Senanayake}, \binits{M.}},
\bauthor{\bsnm{Senthooran}, \binits{I.}},
\bauthor{\bsnm{Barca}, \binits{J.C.}},
\bauthor{\bsnm{Chung}, \binits{H.}},
\bauthor{\bsnm{Kamruzzaman}, \binits{J.}},
\bauthor{\bsnm{Murshed}, \binits{M.}}:
\batitle{Search and tracking algorithms for swarms of robots: A survey}.
\bjtitle{Robotics and Autonomous Systems}
\bvolume{75},
\bfpage{422}--\blpage{434}
(\byear{2016})
\end{barticle}
\endbibitem

\bibitem[\protect\citeauthoryear{Goldhoorn
  et~al.}{2018}]{goldhoorn2018searching}
\begin{barticle}
\bauthor{\bsnm{Goldhoorn}, \binits{A.}},
\bauthor{\bsnm{Garrell}, \binits{A.}},
\bauthor{\bsnm{Alqu{\'e}zar}, \binits{R.}},
\bauthor{\bsnm{Sanfeliu}, \binits{A.}}:
\batitle{Searching and tracking people with cooperative mobile robots}.
\bjtitle{Autonomous robots}
\bvolume{42}(\bissue{4}),
\bfpage{739}--\blpage{759}
(\byear{2018})
\end{barticle}
\endbibitem

\bibitem[\protect\citeauthoryear{Pitre
  et~al.}{2012}]{uav-route-csat-balance-2012}
\begin{barticle}
\bauthor{\bsnm{Pitre}, \binits{R.R.}},
\bauthor{\bsnm{Li}, \binits{X.R.}},
\bauthor{\bsnm{Delbalzo}, \binits{R.}}:
\batitle{Uav route planning for joint search and track missions---an
  information-value approach}.
\bjtitle{IEEE Transactions on Aerospace and Electronic Systems}
\bvolume{48}(\bissue{3}),
\bfpage{2551}--\blpage{2565}
(\byear{2012})
\end{barticle}
\endbibitem

\bibitem[\protect\citeauthoryear{Blum and Gro{\ss}}{2015}]{blum2015swarm}
\begin{botherref}
\oauthor{\bsnm{Blum}, \binits{C.}},
\oauthor{\bsnm{Gro{\ss}}, \binits{R.}}:
Swarm intelligence in optimization and robotics.
Springer handbook of computational intelligence,
1291--1309
(2015)
\end{botherref}
\endbibitem

\bibitem[\protect\citeauthoryear{Chen
  et~al.}{2022a}]{intention-vehicle-prediction-2022}
\begin{barticle}
\bauthor{\bsnm{Chen}, \binits{X.}},
\bauthor{\bsnm{Zhang}, \binits{H.}},
\bauthor{\bsnm{Zhao}, \binits{F.}},
\bauthor{\bsnm{Hu}, \binits{Y.}},
\bauthor{\bsnm{Tan}, \binits{C.}},
\bauthor{\bsnm{Yang}, \binits{J.}}:
\batitle{Intention-aware vehicle trajectory prediction based on
  spatial-temporal dynamic attention network for internet of vehicles}.
\bjtitle{IEEE Transactions on Intelligent Transportation Systems}
\bvolume{23}(\bissue{10}),
\bfpage{19471}--\blpage{19483}
(\byear{2022})
\doiurl{10.1109/TITS.2022.3170551}
\end{barticle}
\endbibitem

\bibitem[\protect\citeauthoryear{Chen et~al.}{2022b}]{phd-filter-limit-2022}
\begin{barticle}
\bauthor{\bsnm{Chen}, \binits{J.}},
\bauthor{\bsnm{Xie}, \binits{Z.}},
\bauthor{\bsnm{Dames}, \binits{P.}}:
\batitle{The semantic phd filter for multi-class target tracking: From theory
  to practice}.
\bjtitle{Robotics and Autonomous Systems}
\bvolume{149},
\bfpage{103947}
(\byear{2022})
\doiurl{10.1016/j.robot.2021.103947}
\end{barticle}
\endbibitem

\bibitem[\protect\citeauthoryear{Hart et~al.}{1968}]{hart1968formal}
\begin{barticle}
\bauthor{\bsnm{Hart}, \binits{P.E.}},
\bauthor{\bsnm{Nilsson}, \binits{N.J.}},
\bauthor{\bsnm{Raphael}, \binits{B.}}:
\batitle{A formal basis for the heuristic determination of minimum cost paths}.
\bjtitle{IEEE transactions on Systems Science and Cybernetics}
\bvolume{4}(\bissue{2}),
\bfpage{100}--\blpage{107}
(\byear{1968})
\end{barticle}
\endbibitem

\bibitem[\protect\citeauthoryear{R{\"o}smann
  et~al.}{2017}]{rosmann2017integrated}
\begin{barticle}
\bauthor{\bsnm{R{\"o}smann}, \binits{C.}},
\bauthor{\bsnm{Hoffmann}, \binits{F.}},
\bauthor{\bsnm{Bertram}, \binits{T.}}:
\batitle{Integrated online trajectory planning and optimization in distinctive
  topologies}.
\bjtitle{Robotics and Autonomous Systems}
\bvolume{88},
\bfpage{142}--\blpage{153}
(\byear{2017})
\end{barticle}
\endbibitem

\bibitem[\protect\citeauthoryear{Brännström
  et~al.}{2010}]{model-based-prediction-2010}
\begin{barticle}
\bauthor{\bsnm{Brännström}, \binits{M.}},
\bauthor{\bsnm{Coelingh}, \binits{E.}},
\bauthor{\bsnm{Sjöberg}, \binits{J.}}:
\batitle{Model-based threat assessment for avoiding arbitrary vehicle
  collisions}.
\bjtitle{IEEE Transactions on Intelligent Transportation Systems}
\bvolume{11}(\bissue{3}),
\bfpage{658}--\blpage{669}
(\byear{2010})
\end{barticle}
\endbibitem

\bibitem[\protect\citeauthoryear{Hillenbrand
  et~al.}{2006}]{model-based-prediction2-2006}
\begin{barticle}
\bauthor{\bsnm{Hillenbrand}, \binits{J.}},
\bauthor{\bsnm{Spieker}, \binits{A.M.}},
\bauthor{\bsnm{Kroschel}, \binits{K.}}:
\batitle{A multilevel collision mitigation approach—its situation assessment,
  decision making, and performance tradeoffs}.
\bjtitle{IEEE Transactions on Intelligent Transportation Systems}
\bvolume{7}(\bissue{4}),
\bfpage{528}--\blpage{540}
(\byear{2006})
\end{barticle}
\endbibitem

\bibitem[\protect\citeauthoryear{Kalman}{1960}]{kalman-filter}
\begin{barticle}
\bauthor{\bsnm{Kalman}, \binits{R.E.}}:
\batitle{{A New Approach to Linear Filtering and Prediction Problems}}.
\bjtitle{Journal of Basic Engineering}
\bvolume{82}(\bissue{1}),
\bfpage{35}--\blpage{45}
(\byear{1960})
\end{barticle}
\endbibitem

\bibitem[\protect\citeauthoryear{Rasmussen and Williams}{2006}]{GPR-2006}
\begin{bbook}
\bauthor{\bsnm{Rasmussen}, \binits{C.E.}},
\bauthor{\bsnm{Williams}, \binits{C.K.I.}}:
\bbtitle{Gaussian Processes for Machine Learning.}
\bsertitle{Adaptive computation and machine learning},
pp. --\blpage{1248}.
\bpublisher{MIT Press}, \blocation{???}
(\byear{2006})
\end{bbook}
\endbibitem

\bibitem[\protect\citeauthoryear{Nguyen
  et~al.}{2024}]{gp-trajectory-nguyen-2024}
\begin{bbook}
\bauthor{\bsnm{Nguyen}, \binits{K.}},
\bauthor{\bsnm{Krumm}, \binits{J.}},
\bauthor{\bsnm{Shahabi}, \binits{C.}}:
\bbtitle{Gaussian Process for Trajectories},
\bedition{1}st edn.,
pp. \bfpage{37}--\blpage{48}.
\bpublisher{Association for Computing Machinery},
\blocation{New York, NY, USA}
(\byear{2024}).
\burl{https://doi.org/10.1145/3617291.3617296}
\end{bbook}
\endbibitem

\bibitem[\protect\citeauthoryear{Goli et~al.}{2018}]{gp-trajectory-IV-2018}
\begin{bchapter}
\bauthor{\bsnm{Goli}, \binits{S.A.}},
\bauthor{\bsnm{Far}, \binits{B.H.}},
\bauthor{\bsnm{Fapojuwo}, \binits{A.O.}}:
\bctitle{Vehicle trajectory prediction with gaussian process regression in
  connected vehicle environment$\star$}.
In: \bbtitle{2018 IEEE Intelligent Vehicles Symposium (IV)},
pp. \bfpage{550}--\blpage{555}
(\byear{2018}).
\burl{https://ieeexplore.ieee.org/document/8500614}
\end{bchapter}
\endbibitem

\bibitem[\protect\citeauthoryear{Park and
  Choi}{2020}]{gp-trajectory-iros-park-2020}
\begin{barticle}
\bauthor{\bsnm{Park}, \binits{J.}},
\bauthor{\bsnm{Choi}, \binits{J.}}:
\batitle{Gaussian process online learning with a sparse data stream}.
\bjtitle{IEEE Robotics and Automation Letters}
\bvolume{5}(\bissue{4}),
\bfpage{5977}--\blpage{5984}
(\byear{2020})
\end{barticle}
\endbibitem

\bibitem[\protect\citeauthoryear{Alahi et~al.}{2016}]{social-lstm-2016}
\begin{bchapter}
\bauthor{\bsnm{Alahi}, \binits{A.}},
\bauthor{\bsnm{Goel}, \binits{K.}},
\bauthor{\bsnm{Ramanathan}, \binits{V.}},
\bauthor{\bsnm{Robicquet}, \binits{A.}},
\bauthor{\bsnm{Fei-Fei}, \binits{L.}},
\bauthor{\bsnm{Savarese}, \binits{S.}}:
\bctitle{Social {LSTM}: Human trajectory prediction in crowded spaces}.
In: \bbtitle{IEEE Conference on Computer Vision and Pattern Recognition
  (CVPR)},
pp. \bfpage{961}--\blpage{971}
(\byear{2016}).
\burl{http://openaccess.thecvf.com/content_cvpr_2016/papers/Alahi_Social_LSTM_Human_CVPR_2016_paper.pdf}
\end{bchapter}
\endbibitem

\bibitem[\protect\citeauthoryear{Zamboni
  et~al.}{2022}]{LSTM-normalization-pedestrain-2022}
\begin{barticle}
\bauthor{\bsnm{Zamboni}, \binits{S.}},
\bauthor{\bsnm{Kefato}, \binits{Z.T.}},
\bauthor{\bsnm{Girdzijauskas}, \binits{S.}},
\bauthor{\bsnm{Norén}, \binits{C.}},
\bauthor{\bsnm{Dal~Col}, \binits{L.}}:
\batitle{Pedestrian trajectory prediction with convolutional neural networks}.
\bjtitle{Pattern Recognition}
\bvolume{121},
\bfpage{108252}
(\byear{2022})
\doiurl{10.1016/j.patcog.2021.108252}
\end{barticle}
\endbibitem

\bibitem[\protect\citeauthoryear{Zhao
  et~al.}{2019}]{multi-agent-tensor-fusion-2019}
\begin{bchapter}
\bauthor{\bsnm{Zhao}, \binits{T.}},
\bauthor{\bsnm{Xu}, \binits{Y.}},
\bauthor{\bsnm{Monfort}, \binits{M.}},
\bauthor{\bsnm{Choi}, \binits{W.}},
\bauthor{\bsnm{Baker}, \binits{C.}},
\bauthor{\bsnm{Zhao}, \binits{Y.}},
\bauthor{\bsnm{Wang}, \binits{Y.}},
\bauthor{\bsnm{Wu}, \binits{Y.N.}}:
\bctitle{Multi-agent tensor fusion for contextual trajectory prediction}.
In: \bbtitle{IEEE/CVF Conference on Computer Vision and Pattern Recognition
  (CVPR)},
pp. \bfpage{12126}--\blpage{12134}
(\byear{2019}).
\burl{http://openaccess.thecvf.com/content_CVPR_2019/html/Zhao_Multi-Agent_Tensor_Fusion_for_Contextual_Trajectory_Prediction_CVPR_2019_paper.html}
\end{bchapter}
\endbibitem

\bibitem[\protect\citeauthoryear{Carrillo
  et~al.}{2015}]{Carrillo-shanon-renyi-entropy-2015}
\begin{bchapter}
\bauthor{\bsnm{Carrillo}, \binits{H.}},
\bauthor{\bsnm{Dames}, \binits{P.}},
\bauthor{\bsnm{Kumar}, \binits{V.}},
\bauthor{\bsnm{Castellanos}, \binits{J.A.}}:
\bctitle{Autonomous robotic exploration using occupancy grid maps and graph
  slam based on shannon and rényi entropy}.
In: \bbtitle{IEEE International Conference on Robotics and Automation (ICRA)},
pp. \bfpage{487}--\blpage{494}
(\byear{2015}).
\burl{https://ieeexplore.ieee.org/abstract/document/7139224}
\end{bchapter}
\endbibitem

\bibitem[\protect\citeauthoryear{Hollinger}{2015}]{sampling-long-horizon-hollinger-2015}
\begin{bchapter}
\bauthor{\bsnm{Hollinger}, \binits{G.}}:
\bctitle{Long-horizon robotic search and classification using sampling-based
  motion planning}.
In: \bbtitle{Robotics: Science and Systems (RSS)}
(\byear{2015}).
\burl{https://www.roboticsproceedings.org/rss11/p10.pdf}
\end{bchapter}
\endbibitem

\bibitem[\protect\citeauthoryear{Hausman et~al.}{2015}]{Hausman-et-al-2015}
\begin{barticle}
\bauthor{\bsnm{Hausman}, \binits{K.}},
\bauthor{\bsnm{Müller}, \binits{J.}},
\bauthor{\bsnm{Hariharan}, \binits{A.}},
\bauthor{\bsnm{Ayanian}, \binits{N.}},
\bauthor{\bsnm{Sukhatme}, \binits{G.S.}}:
\batitle{Cooperative multi-robot control for target tracking with onboard
  sensing}.
\bjtitle{The International Journal of Robotics Research}
\bvolume{34}(\bissue{13}),
\bfpage{1660}--\blpage{1677}
(\byear{2015})
\end{barticle}
\endbibitem

\bibitem[\protect\citeauthoryear{Dias
  et~al.}{2006}]{market-approach-survey-2006}
\begin{barticle}
\bauthor{\bsnm{Dias}, \binits{M.B.}},
\bauthor{\bsnm{Zlot}, \binits{R.}},
\bauthor{\bsnm{Kalra}, \binits{N.}},
\bauthor{\bsnm{Stentz}, \binits{A.}}:
\batitle{Market-based multirobot coordination: A survey and analysis}.
\bjtitle{Proceedings of the IEEE}
\bvolume{94}(\bissue{7}),
\bfpage{1257}--\blpage{1270}
(\byear{2006})
\doiurl{10.1109/JPROC.2006.876939}
\end{barticle}
\endbibitem

\bibitem[\protect\citeauthoryear{Kuhn}{1955}]{Kuhn1955Hungarian}
\begin{barticle}
\bauthor{\bsnm{Kuhn}, \binits{H.W.}}:
\batitle{{The Hungarian Method for the Assignment Problem}}.
\bjtitle{Naval Research Logistics Quarterly}
\bvolume{2}(\bissue{1--2}),
\bfpage{83}--\blpage{97}
(\byear{1955})
\end{barticle}
\endbibitem

\bibitem[\protect\citeauthoryear{Vaughan}{2008}]{vaughan2008massively}
\begin{barticle}
\bauthor{\bsnm{Vaughan}, \binits{R.}}:
\batitle{Massively multi-robot simulation in stage}.
\bjtitle{Swarm intelligence}
\bvolume{2},
\bfpage{189}--\blpage{208}
(\byear{2008})
\end{barticle}
\endbibitem

\bibitem[\protect\citeauthoryear{Pellegrini
  et~al.}{2009}]{eth-dataset-iccv-2009}
\begin{bchapter}
\bauthor{\bsnm{Pellegrini}, \binits{S.}},
\bauthor{\bsnm{Ess}, \binits{A.}},
\bauthor{\bsnm{Schindler}, \binits{K.}},
\bauthor{\bsnm{Van~Gool}, \binits{L.}}:
\bctitle{You'll never walk alone: Modeling social behavior for multi-target
  tracking}.
In: \bbtitle{IEEE International Conference on Computer Vision (ICCV)},
pp. \bfpage{261}--\blpage{268}
(\byear{2009}).
\burl{https://ieeexplore.ieee.org/abstract/document/5459260/}
\end{bchapter}
\endbibitem

\bibitem[\protect\citeauthoryear{Sturtevant}{2024}]{pathfinding-benchmark}
\begin{botherref}
\oauthor{\bsnm{Sturtevant}, \binits{N.}}:
Pathfinding Benchmarks.
\url{https://movingai.com/benchmarks/index.html}, Accessed on Jan 1, 2024
(2024).
\url{https://movingai.com/benchmarks/index.html}
\end{botherref}
\endbibitem

\bibitem[\protect\citeauthoryear{Pestman}{1998}]{pestman1998mathematical}
\begin{bbook}
\bauthor{\bsnm{Pestman}, \binits{W.R.}}:
\bbtitle{Mathematical Statistics: an Introduction}.
\bpublisher{W. de Gruyter}, \blocation{???}
(\byear{1998}).
\burl{https://www.degruyter.com/document/doi/10.1515/9783110809350/html?lang=en}
\end{bbook}
\endbibitem

\bibitem[\protect\citeauthoryear{Aurenhammer}{1991}]{vornoi-1991}
\begin{barticle}
\bauthor{\bsnm{Aurenhammer}, \binits{F.}}:
\batitle{Voronoi diagrams—a survey of a fundamental geometric data
  structure}.
\bjtitle{ACM Computing Surveys}
\bvolume{23}(\bissue{3}),
\bfpage{345}--\blpage{405}
(\byear{1991})
\doiurl{10.1145/116873.116880}
\end{barticle}
\endbibitem

\bibitem[\protect\citeauthoryear{Figueiredo et~al.}{2018}]{vornoi-mcpp-2018}
\begin{bchapter}
\bauthor{\bsnm{Figueiredo}, \binits{L.C.}},
\bauthor{\bsnm{Lelis De~Carvalho}},
\bauthor{\bsnm{Pimenta}, \binits{L.C.D.A.}}:
\bctitle{Voronoi multi-robot coverage control in non-convex environments with
  human interaction in virtual reality}.
In: \bbtitle{XXII Congresso Brasileiro de Automática}
(\byear{2018}).
\burl{http://www.swge.inf.br/proceedings/paper/?P=CBA2018-0563}
\end{bchapter}
\endbibitem

\bibitem[\protect\citeauthoryear{Mier
  et~al.}{2023}]{Mier_Fields2Cover_An_open-source_2023}
\begin{barticle}
\bauthor{\bsnm{Mier}, \binits{G.}},
\bauthor{\bsnm{Valente}, \binits{J.}},
\bauthor{\bsnm{Bruin}, \binits{S.}}:
\batitle{Fields2cover: An open-source coverage path planning library for
  unmanned agricultural vehicles}.
\bjtitle{IEEE Robotics and Automation Letters}
\bvolume{8}(\bissue{4}),
\bfpage{2166}--\blpage{2172}
(\byear{2023})
\doiurl{10.1109/LRA.2023.3248439}
\end{barticle}
\endbibitem

\bibitem[\protect\citeauthoryear{Pellegrini et~al.}{2009}]{ade-Pellegrini_2009}
\begin{bchapter}
\bauthor{\bsnm{Pellegrini}, \binits{S.}},
\bauthor{\bsnm{Ess}, \binits{A.}},
\bauthor{\bsnm{Schindler}, \binits{K.}},
\bauthor{\bsnm{Van~Gool}, \binits{L.}}:
\bctitle{You’ll never walk alone: Modeling social behavior for multi-target
  tracking}.
In: \bbtitle{2009 IEEE 12th International Conference on Computer Vision},
pp. \bfpage{261}--\blpage{268}.
\bpublisher{IEEE}, \blocation{???}
(\byear{2009}).
\burl{http://ieeexplore.ieee.org/document/5459260/}
\end{bchapter}
\endbibitem

\bibitem[\protect\citeauthoryear{Fritsch et~al.}{2013}]{Fritsch2013ITSC}
\begin{bchapter}
\bauthor{\bsnm{Fritsch}, \binits{J.}},
\bauthor{\bsnm{Kuehnl}, \binits{T.}},
\bauthor{\bsnm{Geiger}, \binits{A.}}:
\bctitle{A new performance measure and evaluation benchmark for road detection
  algorithms}.
In: \bbtitle{International Conference on Intelligent Transportation Systems
  (ITSC)}
(\byear{2013})
\end{bchapter}
\endbibitem

\bibitem[\protect\citeauthoryear{Geiger et~al.}{2012}]{Geiger2012CVPR}
\begin{bchapter}
\bauthor{\bsnm{Geiger}, \binits{A.}},
\bauthor{\bsnm{Lenz}, \binits{P.}},
\bauthor{\bsnm{Urtasun}, \binits{R.}}:
\bctitle{Are we ready for autonomous driving? the kitti vision benchmark
  suite}.
In: \bbtitle{Conference on Computer Vision and Pattern Recognition (CVPR)}
(\byear{2012})
\end{bchapter}
\endbibitem

\bibitem[\protect\citeauthoryear{Dagestad et~al.}{2018}]{opendrift-2018}
\begin{barticle}
\bauthor{\bsnm{Dagestad}, \binits{K.-F.}},
\bauthor{\bsnm{R{\"o}hrs}, \binits{J.}},
\bauthor{\bsnm{Breivik}, \binits{{\O}.}},
\bauthor{\bsnm{\r{A}dlandsvik}, \binits{B.}}:
\batitle{Opendrift v1.0: a generic framework for trajectory modelling}.
\bjtitle{Geoscientific Model Development}
\bvolume{11}(\bissue{4}),
\bfpage{1405}--\blpage{1420}
(\byear{2018})
\end{barticle}
\endbibitem

\bibitem[\protect\citeauthoryear{Koenig and Howard}{2004}]{gazebo_2004}
\begin{bchapter}
\bauthor{\bsnm{Koenig}, \binits{N.}},
\bauthor{\bsnm{Howard}, \binits{A.}}:
\bctitle{Design and use paradigms for gazebo, an open-source multi-robot
  simulator}.
In: \bbtitle{IEEE/RSJ International Conference on Intelligent Robots and
  Systems (IROS)},
vol. \bseriesno{3},
pp. \bfpage{2149}--\blpage{2154}
(\byear{2004})
\end{bchapter}
\endbibitem

\bibitem[\protect\citeauthoryear{Wang et~al.}{2022}]{gazebo-fidelity-1}
\begin{barticle}
\bauthor{\bsnm{Wang}, \binits{S.}},
\bauthor{\bsnm{Han}, \binits{R.}},
\bauthor{\bsnm{Hong}, \binits{Y.}},
\bauthor{\bsnm{Hao}, \binits{Q.}},
\bauthor{\bsnm{Wen}, \binits{M.}},
\bauthor{\bsnm{Musavian}, \binits{L.}},
\bauthor{\bsnm{Mumtaz}, \binits{S.}},
\bauthor{\bsnm{Kwan~Ng}, \binits{D.W.}}:
\batitle{Robotic wireless energy transfer in dynamic environments: System
  design and experimental validation}.
\bjtitle{IEEE Communications Magazine}
\bvolume{60}(\bissue{3}),
\bfpage{40}--\blpage{46}
(\byear{2022})
\doiurl{10.1109/MCOM.001.2100738}
\end{barticle}
\endbibitem

\bibitem[\protect\citeauthoryear{Giubilato et~al.}{2020}]{gazebo-fidelity-2}
\begin{bchapter}
\bauthor{\bsnm{Giubilato}, \binits{R.}},
\bauthor{\bsnm{Masili}, \binits{A.}},
\bauthor{\bsnm{Chiodini}, \binits{S.}},
\bauthor{\bsnm{Pertile}, \binits{M.}},
\bauthor{\bsnm{Debei}, \binits{S.}}:
\bctitle{Simulation framework for mobile robots in planetary-like
  environments}.
In: \bbtitle{IEEE 7th International Workshop on Metrology for AeroSpace
  (MetroAeroSpace)},
pp. \bfpage{594}--\blpage{599}
(\byear{2020}).
\doiurl{10.1109/MetroAeroSpace48742.2020.9160154}
\end{bchapter}
\endbibitem

\bibitem[\protect\citeauthoryear{Guizzo and Ackerman}{2017}]{turtlebot-2017}
\begin{barticle}
\bauthor{\bsnm{Guizzo}, \binits{E.}},
\bauthor{\bsnm{Ackerman}, \binits{E.}}:
\batitle{The turtlebot3 teacher}.
\bjtitle{IEEE Spectrum}
\bvolume{54}(\bissue{8}),
\bfpage{19}--\blpage{20}
(\byear{2017})
\doiurl{10.1109/MSPEC.2017.8000281}
\end{barticle}
\endbibitem

\bibitem[\protect\citeauthoryear{Wang et~al.}{2022}]{wang2022marine}
\begin{bchapter}
\bauthor{\bsnm{Wang}, \binits{P.}},
\bauthor{\bsnm{Meghjani}, \binits{M.}},
\bauthor{\bsnm{Chen}, \binits{G.}}:
\bctitle{Marine trash collection: A multi-agent, multi-target search}.
In: \bbtitle{OCEANS 2022, Hampton Roads},
pp. \bfpage{1}--\blpage{7}
(\byear{2022}).
\bcomment{IEEE}
\end{bchapter}
\endbibitem

\end{thebibliography}
